\documentclass[11pt]{article}

\usepackage[final]{acl}

\usepackage{times}
\usepackage{latexsym}

\usepackage{makecell}
\usepackage{adjustbox}
\usepackage[T1]{fontenc}
\usepackage[utf8]{inputenc}

\usepackage{microtype}

\usepackage{inconsolata}

\usepackage{graphicx}

\usepackage{pifont}   % provides \ding{}

\newcommand{\xmark}{\ding{55}}

\newcommand{\cmark}{\ding{51}}

\usepackage{wrapfig}

\usepackage{xcolor}
\usepackage{comment}
\usepackage{tikz}
\definecolor{darkgreen}{rgb}{0.0, 0.5, 0.0}
\usepackage{xcolor}
\usepackage{colortbl}
\usepackage{tabularx}  % Add to preamble

\definecolor{customblue}{rgb}{0.168, 0.364, 0.557}
\usepackage[most]{tcolorbox}

\usepackage[edges]{forest}
\definecolor{hidden-draw}{RGB}{0,0,0}

\title{It's High Time \raisebox{-0.35\height}
{\includegraphics[width=1.1cm]{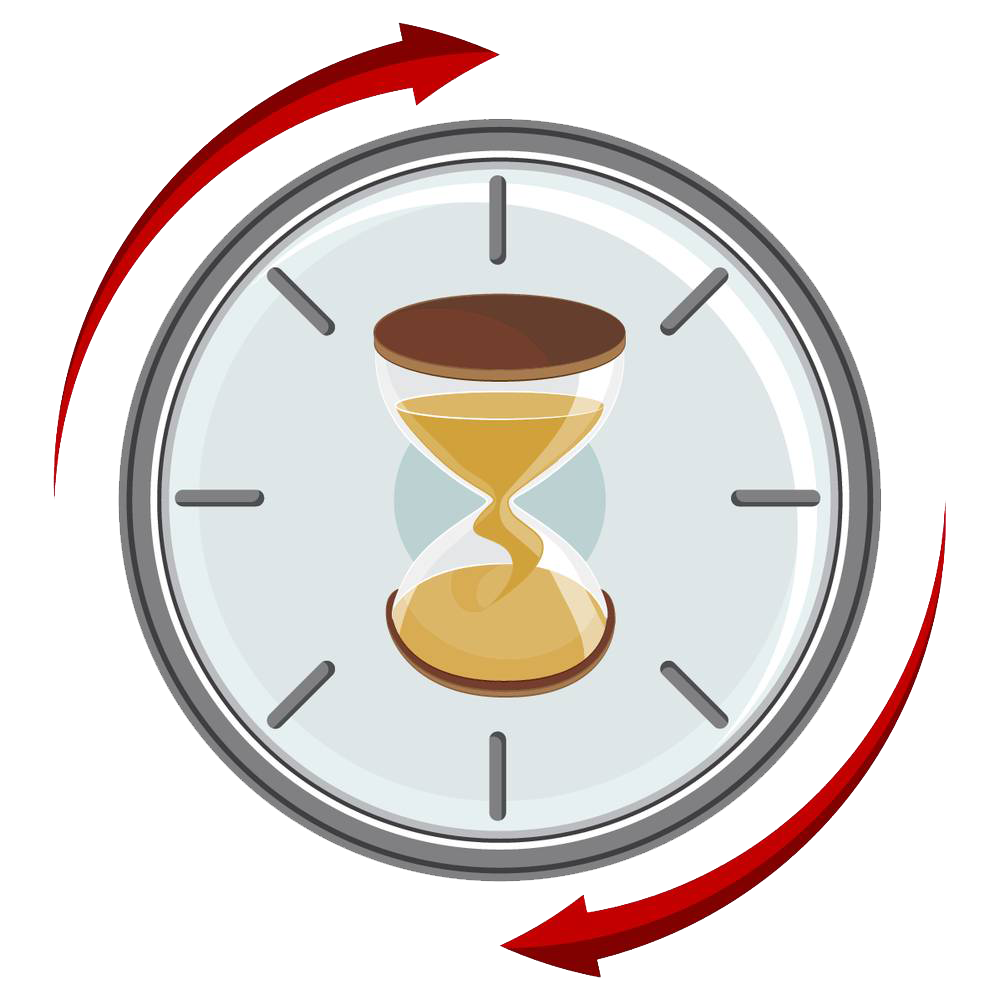}}: A Survey of Temporal Question Answering}

\author{
  Bhawna Piryani\thanks{Corresponding Author.}$^{\spadesuit}$, Abdelrahman Abdallah$^{\spadesuit}$, Jamshid Mozafari$^{\spadesuit}$, \\ {\bf Avishek Anand$^{\heartsuit}$}, {\bf Adam Jatowt$^{\spadesuit}$} \\
  $^{\spadesuit}$University of Innsbruck, $^{\heartsuit}$Delft University of Technology \\
  \texttt{\{bhawna.piryani, abdelrahman.abdallah, jamshid.mozafari, adam.jatowt\}@uibk.ac.at} \\
  \texttt{avishek.anand@tudelft.nl}
}

\begin{document}
\maketitle
\begin{abstract}

Time plays a critical role in how information is generated, retrieved, and interpreted. In this survey, we provide a comprehensive overview of \textit{Temporal Question Answering} (TQA), a research area that focuses on answering questions involving temporal constraints or context. As time-stamped content from sources like news articles, web archives, and knowledge bases continues to grow, TQA systems must address challenges such as detecting temporal intent, normalizing time expressions, ordering events, and reasoning over evolving or ambiguous facts. We organize existing work through a unified perspective that captures the interaction between corpus temporality, question temporality, and model capabilities, enabling a systematic comparison of datasets, tasks, and approaches.  We review recent advances in TQA enabled by neural architectures, especially transformer-based models and Large Language Models (LLMs), highlighting progress in temporal language modeling, retrieval-augmented generation (RAG), and temporal reasoning. We also discuss benchmark datasets and evaluation strategies designed to test temporal robustness, recency awareness, and generalization.
\end{abstract}

\noindent
\begin{wrapfigure}{l}{0.03\textwidth}
    \centering
    \hypertarget{github-link}{}
    \href{https://github.com/DataScienceUIBK/TemporalQA-Survey}{%
    \includegraphics[width=0.05\textwidth]{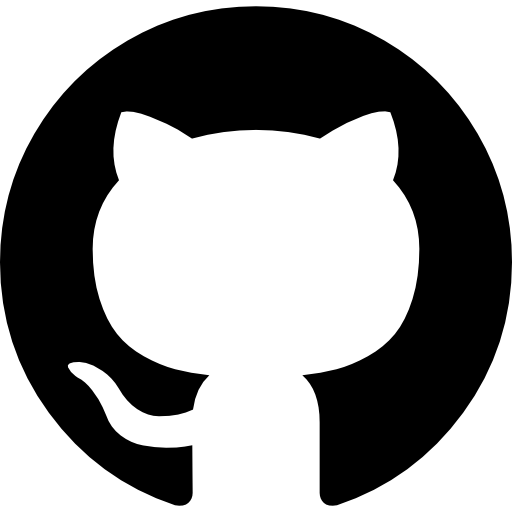}
    }
\vspace{-2em}
\end{wrapfigure}

\noindent
{\fontsize{10}{10}\selectfont\url{https://github.com/DataScienceUIBK/TemporalQA-Survey}}

\begin{figure*}[t]
    \centering
    \includegraphics[width=0.9\textwidth]{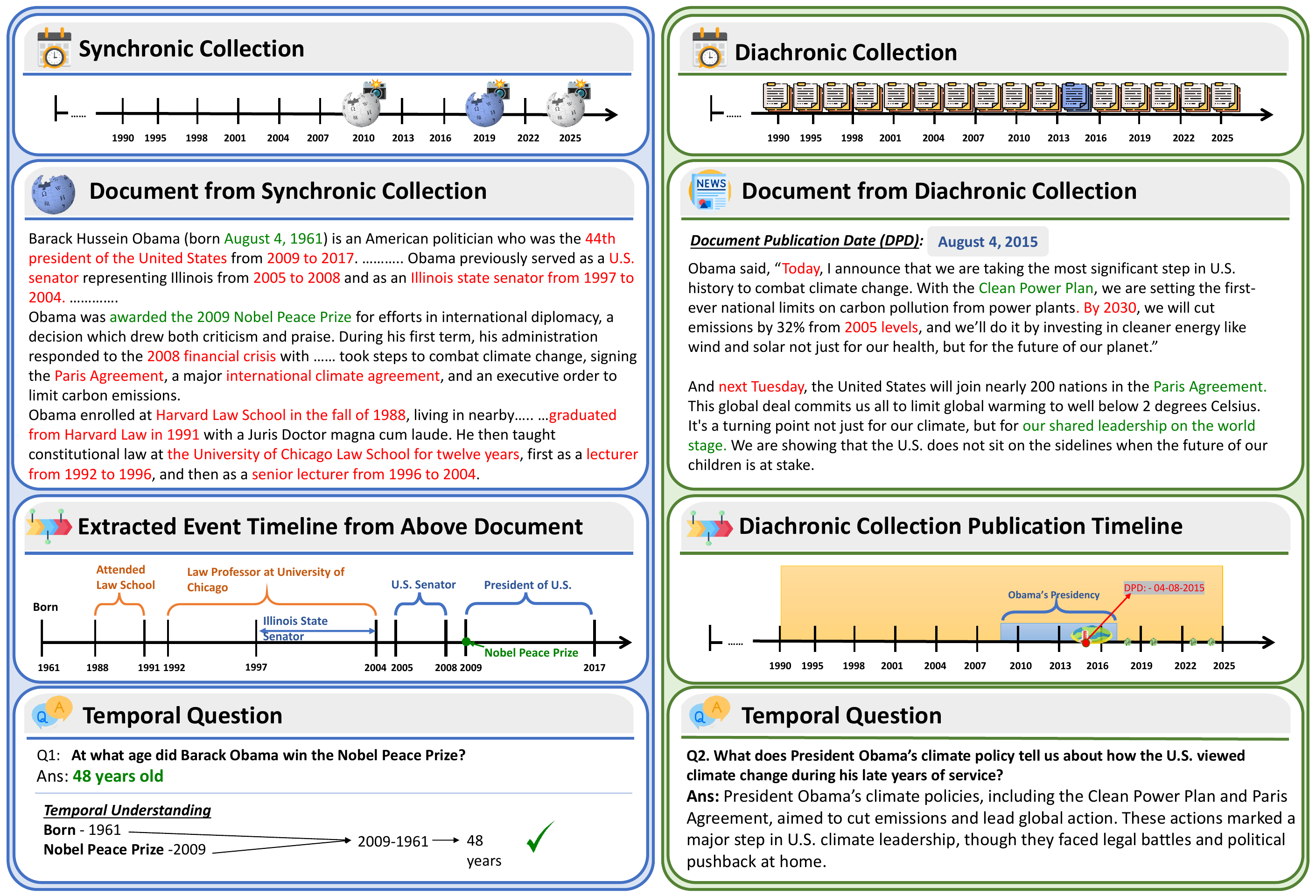}
    \caption{ The figure illustrates an example of answering a temporal question using documents from a synchronic (left) and diachronic (right) collection. The Synchronic and Diachronic Collection panel highlights the specific document used to answer the question in each case in \textcolor{blue} {blue}. \textcolor{red}{Red} marks temporal signals within the text, while \textcolor{darkgreen}{Green} highlights the answer-supporting spans. On the left, an event timeline is inferred from the internal structure of an undated document in the synchronic collection. On the right, the collection timeline displays the distribution of documents over time in the diachronic collection; \textcolor{red}{Red} dots indicate documents that contain an answer, while \textcolor{green}{green} dots denote documents related to the event described in Question Q2. }
    \label{fig:example_figure}
\end{figure*}

\section{Introduction}
\label{sec:Introduction}

Time fundamentally shapes how information is written, retrieved, and interpreted. As digital content continues to expand across time-stamped sources such as news archives, social media, and knowledge bases, the ability to reason about when events occur and how information evolves has become essential \cite{alonso2007value}. Temporal Question Answering (TQA) focuses on questions whose interpretation or answers depend on temporal context. Unlike standard Question Answering (QA), TQA requires systems to detect temporal intent, ground temporal references, infer event order, and reason over dynamic or ambiguous facts, rather than relying solely on surface-level retrieval.

% TQA faces distinct challenges that set it apart from standard QA tasks. 
These requirements introduce challenges that distinguish TQA from conventional QA tasks. One key challenge is temporal ambiguity resolution, where vague expressions such as \textit{"recently"} or \textit{"after the war"} must be interpreted relative to context. Another is cross-temporal reasoning, which involves understanding causal and sequential relationships across events. In addition, knowledge volatility refers to the evolution of facts over time, which renders static corpora and pre-trained models inadequate for answering time-sensitive queries. Temporal intent may be expressed explicitly %(e.g., "post–World War II," "in 1998") 
or implicitly, requiring systems to infer appropriate timeframes. Addressing these challenges requires temporally aware reasoning, context-sensitive retrieval, and mechanisms for adapting to evolving knowledge \citep{berberich2010language}.

These challenges are illustrated in Figure~\ref{fig:example_figure}, which presents two example questions. \textit{Q1: "At what age did Obama win the Nobel Peace Prize?"} requires identifying and grounding two temporal anchors-Obama's birth year (1961) and the year he received the Nobel Peace Prize (2009)-and computing the interval between them to derive the answer: \textit{48 years old}. %This question exemplifies explicit temporal grounding and temporal arithmetic.

\textit{Q2: "What does President Obama's climate policy tell us about how the U.S. viewed climate change during his late years of service?"} requires a more nuanced understanding of context. The phrase \textit{"late years of service"} must be anchored to Obama's presidency timeline (2009-2017). Accurate answering requires access to contemporaneous policy documents from that specific period, rather than relying on retrospective analyses. This highlights that temporal context determines which knowledge sources are relevant. 
As shown in the figure, the model must also interpret relative expressions such as \textit{"today"} or \textit{"next week"} within document content by anchoring them to the \textit{document’s publication date (August 4, 2015)}, thereby contextualizing Obama's climate policy statement within it's correct historical context.

These examples illustrate a broader point: effective TQA requires aligning three interacting dimensions: (i) the temporality of the underlying corpus (diachronic vs. synchronic), (ii) the temporal structure of the question (explicit vs. implicit intent, temporal orientation, and reasoning complexity), and (iii) the temporal capabilities of current models (temporal language modeling, temporally aware retrieval, and temporal reasoning). We adopt this three-dimensional perspective as the organizing framework of this survey, enabling us to synthesize prior work not just as datasets and systems, but as a landscape shaped by how different approaches handle mismatches across these dimensions.

Research in TQA has evolved substantially, moving from rule-based pipelines \cite{harabagiu2005question} and statistical models \cite{berberich2010language,wang2020answering} to neural systems capable of richer temporal reasoning capabilities \cite{dhingra-etal-2022-time, wang2023bitimebert}. Early work focused on handcrafted normalization and event ordering rules, which struggled with scale and coverage. Pre-trained language models further shifted the field by supporting more robust temporal reasoning \cite{jain-etal-2023-language-models}, event sequencing \cite{lin-etal-2021-conditional}, and continual temporal adaptation \cite{han-etal-2021-econet}, enabling multi-hop temporal inference and retrieval-augmented updates.

\begin{figure*}[t!]
    \centering
    \tikzstyle{my-box}=[
        rectangle,
        rounded corners,
        minimum height=0.1em,
        minimum width=0.1em,
        inner sep=2pt,
        align=left,
        fill opacity=.5,
        text opacity=1,
    ]
    \tikzstyle{leaf}=[my-box]

    \resizebox{0.9\textwidth}{!}{
        \begin{forest}
            forked edges,
            for tree={
                grow=east,
                reversed=true,
                anchor=base west,
                parent anchor=east,
                child anchor=west,
                base=left,
                font=\fontsize{6}{6}\selectfont,
                rectangle,
                rounded corners,
                align=left,
                minimum width=0.1em,
                edge+={darkgray, line width=0.8pt},
                s sep=2pt,
                inner xsep=2pt,
                inner ysep=2pt,
                draw=hidden-draw,
                ver/.style={rotate=90, child anchor=north, parent anchor=south, anchor=center},
            },
            where level=1{text width=6em,font=\fontsize{6.5}{6.5}\selectfont,}{},
            where level=2{text width=7em,font=\fontsize{5.5}{5.5}\selectfont,}{},
            where level=3{text width=7em,font=\fontsize{5.5}{5.5}\selectfont,}{},
            where level=4{text width=11.3em,font=\fontsize{5.5}{5.5}\selectfont,}{},
            where level=5{text width=6.8em,font=\fontsize{5.5}{5.5}\selectfont,}{},
            [
                Temporal QA, draw=black, fill=gray!10, thick, text=black, ver
                [
                    Dataset \& Evaluation \\  Benchmarks (~\S\ref{sec:dataset_evaluation_becnhmarks}),
                    color=orange!90, fill=orange!20, thick, text=black
                    [
                        Temporal Document\\ Collections (~\S\ref{sec:temporal_collection}),  
                        color=orange!90, fill=orange!20, thick, text=black
                        [
                            Diachronic Collection,
                            , color=orange!90, fill=orange!20, thick, text=black, 
                            [
                            \citet{sandhaus2008new}{,} \citet{NEURIPS2024_58f52a01_newswire}{,} \\
                            \citet{trischler-etal-2017-newsqa}
                            ,color=orange!90, fill=orange!20, thick, text=black, leaf
                            ]
                        ]
                        [
                            Synchronic Collection,
                            color=orange!90, fill=orange!20, thick, text=black
                             [
                            \citet{vrandevcic2014wikidata}
                            ,color=orange!90, fill=orange!20, thick, text=black, leaf
                            ]
                        ]
                    ]
                    [
                        TQA Datasets (~\S\ref{sec:tqa_dataset}),
                        color=orange!90, fill=orange!20, thick, text=black
                        [
                            Simple QA  
                           , color=orange!90, fill=orange!20, thick, text=black,
                             [
                                 ArchivalQA \cite{wang2022archivalqa}{,} \\ TimeQA \cite{chen2dataset}{,} \\
                                 ChroniclingAmericaQA \cite{piryani2024chroniclingamericaqa}{,}\\
                                 SituatedQA \cite{zhang-choi-2021-situatedqa}
                                ,color=orange!90, fill=orange!20, thick, text=black, leaf
                             ]
                        ]
                        [
                            Complex QA  
                           , color=orange!90, fill=orange!20, thick, text=black,
                             [
                                 ComplexTempQA \cite{gruber2024complextempqa}{,} \\
                                 Complex-TR \cite{tan-etal-2024-towards}{,}\\
                                 TempLAMA \cite{dhingra-etal-2022-time}{,} \\ MenatQA \cite{wei-etal-2023-menatqa}{,}\\
                                 TempReason \cite{tan-etal-2023-towards}{,}\\
                                 TempTabQA \cite{gupta-etal-2023-temptabqa}
                                , color=orange!90, fill=orange!20, thick, text=black, leaf
                             ]
                        ]
                    ]
                ]
                [
                    Temporal Tasks (~\S\ref{sec:temporal_prediction_tasks}),
                    color=cyan!100, fill=cyan!15, thick, text=black
                        [
                            Document Dating ~\S\ref{appendix:document_dating_task},
                            color=cyan!100, fill=cyan!15, thick, text=black,
                             [
                                 \citet{dalli-2006-temporal}{,} \citet{jatowt2007detecting}{,}  \citet{kumar2012dating}{,} \\ \citet{salaheldeen2013carbon}{,} \citet{kanhabua2008improving}{,} \citet{jong2005temporal}{,} \\ \citet{niculae-etal-2014-temporal}{,} \citet{vashishth-etal-2018-dating},
                                , color=cyan!100, fill=cyan!15, thick, text=black, text width=20em, leaf 
                             ]
                        ]
                        [
                            Document Focus Time \\ Estimation ~\S\ref{appendix:document_focus_time_estiamtion_task},
                            color=cyan!100, fill=cyan!15, thick, text=black,
                             [
                                 \citet{jatowt2013estimating}{,}  \citet{shrivastava2017concept}{,} \citet{jatowt2015generic},
                                , color=cyan!100, fill=cyan!15, thick, text=black, text width=20em, leaf 
                             ]
                        ]
                        [
                            Temporal Query Profiling ~\S\ref{appendix:temporal_query_profiling_task},
                            color=cyan!100, fill=cyan!15, thick, text=black,
                             [
                                 \citet{kanhabua2010determining}{,} \citet{dakka2008answering}{,}  \citet{jones2007temporal}{,} \\ \citet{kanhabua2011comparison}{,} \citet{gupta2014identifying}
                                , color=cyan!100, fill=cyan!15, thick, text=black, text width=20em, leaf 
                             ]
                        ]
                        [
                            Event Occurrence Time \\ Estimation ~\S\ref{appendix:document_dating_task},
                            color=cyan!100, fill=cyan!15, thick, text=black,
                             [
                                 \citet{das2017estimating}{,} \citet{wang2021event} {,} \citet{morbidoni2018leveraging}{,} \citet{honovich-etal-2020-machine}
                                , color=cyan!100, fill=cyan!15, thick, text=black, text width=20em, leaf 
                             ]
                        ]
                ]
                [
                    Approaches (~\S\ref{sec:temporal_qa_approaches}),
                    color=red!80, fill=red!15, thick, text=black
                    [
                        Rule-Based \& \\Statistical Models (~\S\ref{sec:temporal_rulebased}),
                        color=red!80, fill=red!15, thick, text=black
                        [
                            \citet{Li-Croft}{,} \citet{alonso2007value}{,} \citet{alonso2009clustering}{,}  \citet{arikan2009time}{,} \\ \citet{metzler2009improving}{,}
                            \citet{kanhabua2012learning}{,} \citet{jatowt2005temporal}{,} \\ \citet{holzmann2016tempas}{,} \citet{kulkarni2011understanding} {,}  \citet{harabagiu2005question}{,} \\ \citet{saquete2004splitting}{,} \citet{saquete2009enhancing}{,}  \citet{setty2017modeling}{,} \citet{anand2011temporal}{,} \\ \citet{anand2012index}{,}
                            \citet{saquete2003terseo}{,}  \citet{rizzo2022ranking}{,} \citet{kulkarni2011understanding},
                            color=red!80, fill=red!15, thick, text=black, leaf, text width=20em
                        ]
                    ]
                    [
                        Temporal Language \\ Models (~\S\ref{sec:time_aware_LM}),
                        color=red!80, fill=red!15, thick, text=black
                        [
                            \citet{dhingra-etal-2022-time}{,} \citet{rosin2022time}{,} \citet{wang2023bitimebert} \\ \citet{ren-etal-2023-time}{,} \citet{cole-etal-2023-salient}{,}  \citet{son-oh-2023-time}{,} \citet{cole-etal-2023-salient}{,} \\\citet{rosin-radinsky-2022-temporal}{,}   \citet{su-etal-2023-efficient},
                            ,color=red!80, fill=red!15, thick, text=black, text width=20em, leaf
                        ]
                    ]
                    [
                        Temporal RAG (~\S\ref{sec:temporal_rag}),
                        color=red!80, fill=red!15, thick, text=black
                        [
                            \citet{gade2024itsabouttime}{,} \citet{wu2024time}{,} \citet{wang2023bitimebert} \citet{abdallah2025tempretriever}{,} \\ \citet{siyue2024mrag}{,} \citet{qian-etal-2024-timer4}{,} \citet{vu-etal-2024-freshllms}{,} \citet{wang2025timerag}, 
                            ,color=red!80, fill=red!15, thick, text=black, text width=20em, leaf
                        ]
                    ]
                    [
                        Temporal Reasoning (~\S\ref{sec:temporal_reasoning}),
                        color=red!80, fill=red!15, thick, text=black
                        [
                            \citet{han-etal-2021-econet}{,} \citet{mathur-etal-2021-timers} {,} \citet{niu-etal-2024-contempo}{,} \\ \citet{chu-etal-2023-mtger}{,} \citet{zhu-etal-2023-question}{,} \citet{li2023unlocking}{,} \citet{wu2025chainoftimeline}{,} \\ \citet{yang-etal-2020-improving}{,} \citet{wang-zhao-2024-tram}{,} \citet{fatemi2024test}{,} \\ \citet{feng-etal-2023-generic}{,} \citet{zhang-etal-2024-narrative}{,} \citet{abdallah2026tempo},
                            ,color=red!80, fill=red!15, thick, text=black, text width=20em, leaf
                        ]
                    ]
                ]
            ]
        \end{forest}
    }
    \caption{Taxonomy of Temporal Datasets and Evaluation benchmarks, Tasks, and Approaches.}
    \label{fig:taxonomy_full}
\end{figure*}

Despite this progress, current systems still face challenges such as handling future-oriented questions, reasoning over temporally inconsistent documents, and mitigating knowledge decay. Addressing these problems requires a deeper understanding of the field’s current capabilities and limitations.

While existing surveys\footnote{See Appendix~\ref{appendix:Related_survey} for a detailed discussion.} either focus on general QA/IR \cite{kolomiyets2011survey, zhu2023large, zhu2021retrieving} or on narrow aspects of temporal processing \cite{kobayashi2000information, 10.1561/1500000043, campos2014survey}, to the best of our knowledge, no prior work systematically examines TQA as a full pipeline spanning corpus, question, and model temporality. The last survey reviewing TQA \cite{campos2014survey} predates modern temporal LMs, RAG systems, and today’s large temporal benchmarks, leaving a gap that we aim to fill.

In this survey, we provide a comprehensive synthesis of recent work in TQA over unstructured text. Our scope is limited to systems that operate on natural language questions and unstructured documents. We therefore exclude structured settings such as temporal KG-based QA \cite{su2024temporal, saxena-etal-2021-question} and temporal QA over semi-structured tables \cite{shankarampeta-etal-2025-transienttables}. These settings assume explicit temporal structure (e.g., timestamped rows or evolving attribute records) and involve reasoning challenges that differ from grounding temporal expressions in free text. We also do not provide full coverage of temporal-reasoning benchmarks that are not grounded in textual evidence; for an overview of such resources, we refer readers to \citet{wenzel2023overview}. This focus allows us to examine TQA as an end-to-end pipeline grounded in real-world text corpora, and to analyze how temporal reasoning interacts with retrieval, evidence selection, and evolving content.

We systematically review the evolution of TQA methods, benchmark datasets, evaluation strategies, and emerging challenges. To guide this discussion, we present a taxonomy, as illustrated in Figure~\ref{fig:taxonomy_full}, that organizes prior work into datasets and benchmarks, temporal tasks, and modeling approaches. Throughout the survey, we use the three-dimensional framework introduced above—corpus temporality, question temporality, and model temporality—as an analytical lens to compare methods, reveal mismatches, and identify open challenges.

\begingroup
\rowcolors{2}{blue!5}{white}
\begin{table*}[tb]
\centering
\small
\begin{adjustbox}{width=\textwidth,center}
\begin{tabular}{lcccccccc}
\hline
\textbf{Dataset} & \textbf{\#Questions} &\textbf{ \makecell[c]{Knowledge \\ Source}} & \textbf{\makecell[c]{Creation\\ Method}}  & \textbf{\makecell[c]{Answer \\ Type}}   & \textbf{Time Frame} & \textbf{\makecell[c]{Temporal \\ Metadata}} & \textbf{Multi-Hop}  \\ \hline
NewsQA \cite{trischler-etal-2017-newsqa}                    &119k &News &CS &Freeform  &2007-2015 &\xmark &\xmark \\ 
TDDiscourse \cite{naik-etal-2019-tddiscourse}               &6.1k &News &CS &Extractive  &Unspecified &\xmark &\xmark \\ 
TORQUE \cite{ning-etal-2020-torque}                         &21k &News &CS &Abstractive  &- &\xmark &\xmark \\ 
ArchivalQA \cite{wang2022archivalqa}                        &532k &News &AG &Extractive  &1987-2007 &\cmark &\xmark \\ 
TimeQA \cite{chen2dataset}                                  &41.2k &Wikipedia &AG &Extractive  &1367-2018 &\xmark &\xmark \\ 
TiQ \cite{jia2024tiq}                                       &10k &Wikipedia &AG &Freebase &Unspecified &\xmark &\xmark \\ 
%TimeBench \cite{chu-etal-2024-timebench}                    & & & & & & & & \\ 
TempQuestions \cite{jia2018tempquestions}                   &1.2k & Freebase &AG  & Extractive &Unspecified &\xmark &\cmark \\
TemporalQuestions \cite{wang2021improving}                  &1k &News &CS &Extractive  &1987-2007 &\cmark & \xmark \\ 
TempLAMA \cite{dhingra-etal-2022-time}                      &50k &News &CS &Extractive  &2010-2020 &\cmark &\xmark \\ 
ComplexTempQA \cite{gruber2024complextempqa}                &100.2M &Wikipedia &AG &Extractive  &1987-2023  &\cmark &\cmark \\ 
MenatQA \cite{wei-etal-2023-menatqa}                        &2.8k &Wikipedia &AG &Extractive  &1367-2018 &\xmark &\xmark \\ 
PAT-Question \cite{meem-etal-2024-pat}                      &6,1k &Wikipedia &CS &Extractive  &- &\xmark &\cmark \\
TempTabQA \cite{gupta-etal-2023-temptabqa}                  &11.4k &Wikipedia Info box &CS &Abstractive  &- &\xmark &\xmark \\
SituatedQA \cite{zhang-choi-2021-situatedqa}                &12.2k &Wikipedia &CS &- &$\leq2021$ &\xmark &\xmark \\ 
%TEMPWIKIBIO \cite{cao-wang-2022-time}                       & & & &  & & & \\ \hline
UnSeenTimeQA \cite{uddin2024unseentimeqa}                   &3.6k &Synthetic &AG &Abstractive  &- &\xmark &\cmark \\ 
ChroniclingAmericaQA \cite{piryani2024chroniclingamericaqa} &485k &News &AG & Extractive  &1800-1920 &\cmark &\xmark \\ 
FRESHQA \cite{vu-etal-2024-freshllms}                       &600 &Google Search &CS & - & - &\xmark &\cmark \\ 
COTEMPQA \cite{su-etal-2024-living}                         &4.7k &Wikidata &CS &Abstractive  & $\leq2023$ &\xmark &\cmark \\ 
Test of Time (ToT) \cite{fatemi2024test}                    & 1.8k &Synthetic &AG  &Abstractive &- &\xmark &\cmark \\ 
TIMEDAIL \cite{qin-etal-2021-timedial}                      &1.1k &DailyDialog &CS &Multiple-choice &- &\xmark  &\xmark \\ 
TEMPO \cite{abdallah2026tempo} & 1.7k& Stack Exchange & CS & Abstractive & $\leq2025$ &\cmark &\cmark \\ 
Complex-TR \cite{tan-etal-2024-towards}                     & 10.8k &Wikipedia+Google Search &AG  &Multi-answer & $\leq2023$ &\xmark &\cmark \\ 
StreamingQA \cite{liska22a-streaminqa}                      &147k &News &CS   &Extractive  &2007-2020 &\cmark &\cmark \\ 
TRACIE \cite{zhou-etal-2021-temporal}                       &5.4k &Wikipedia &CS  &Abstractive &$\leq2020$ &\xmark &\xmark \\ 
ForecastQA \cite{jin-etal-2021-forecastqa}                  &10.3k &News &CS  &Multiple-Choice &2015-2019 &\cmark &\cmark \\ 
TEMPREASON \cite{tan-etal-2023-towards}                     &52.8k &Wikipedia/Wikidata &SC &Abstractive  &634-2023 &\xmark &\xmark \\ 
TemporalAlignmentQA \cite{zhao-etal-2024-set}               &20k &Wikipedia &AG  &Abstractive &2000-2023  &\xmark &\xmark \\ 
ReaLTimeQA \cite{realtimeqa}              &5.1k &Search &CS & Multiple-choice &2020-2024 &\xmark &\xmark \\ \hline
\end{tabular}
\end{adjustbox}
\caption{Overview of Temporal QA datasets. Each dataset is characterized by the number of questions, the underlying knowledge source, the question creation method (CS = Crowdsourced, AG = Automatically Generated), the answer type, and the timeframe covered by the knowledge source. A "$\leq$" symbol indicates that the dataset uses a snapshot of Wikipedia and inherits its temporal scope. We also indicate whether temporal metadata is available and whether questions require multi-hop temporal reasoning.}
\label{tab:tqa_datasets}
\end{table*}
\endgroup

\section{Datasets and Evaluation Benchmarks}
\label{sec:dataset_evaluation_becnhmarks}

For readers unfamiliar with basic temporal concepts and terminology, such as \textbf{temporal expressions}, \textbf{temporal anchoring}, or \textbf{temporal signals}, we provide a concise overview in Appendix~\ref{appendix:key_concepts}. We now examine the datasets and evaluation benchmarks used to evaluate systems for temporally grounded question answering. From the three-dimensional perspective introduced earlier, existing benchmarks can be characterized along two primary dimensions: \textbf{corpus temporality} (diachronic vs. synchronic) and \textbf{question temporality} (explicit vs. implicit), which determine the temporal reasoning capabilities being evaluated. %The development of TQA systems fundamentally depends on temporally annotated datasets and robust evaluation methodologies for training, testing, and benchmarking time-aware models. 
This section provides an overview of temporal datasets, organized into two categories: (i) Temporal Document Collections, which serve as knowledge sources, and (ii) TQA datasets, which enable systematic evaluation of temporal reasoning capabilities.

\subsection{Temporal Document Collections}
\label{sec:temporal_collection}

Prior work has utilized three types of temporally structured document collections for temporal reasoning tasks, each offering distinct advantages.

% \textbf{Diachronic corpora} consist of time-stamped documents spanning extended time periods and support retrospective retrieval, diachronic analysis, and event-based reasoning. Representative examples include the \emph{New York Times Annotated Corpus} (1987--2007) \cite{sandhaus2008new}, which underpins \texttt{ArchivalQA} \cite{wang2022archivalqa}; the \emph{CNN/Daily Mail} corpus (2007--2015) \cite{NIPS2015__ccn/dailynews}, used in \texttt{NewsQA} \cite{trischler-etal-2017-newsqa}; and the \emph{Chronicling America} collection (1800--1920), which supports long-range historical QA via \texttt{ChroniclingAmericaQA} \cite{piryani2024chroniclingamericaqa}. More recent resources such as the \emph{Newswire} corpus \cite{NEURIPS2024_58f52a01_newswire} extend coverage to the late 19th and 20th centuries and provide rich metadata (e.g., geo-referenced datelines and entity links) for fine-grained spatio-temporal modeling, while \emph{CUSTOMNEWS} \cite{NEURIPS2021_customnews} (1969--2019) offers broad domain coverage across politics, finance, and sports. Beyond QA, diachronic corpora support related temporal tasks, including semantic drift detection \cite{hamilton-etal-2016-diachronic}, event burst modeling \cite{radinsky2013mining}, and timeline construction \cite{gutehrle-etal-2022-archive}.

\textbf{Diachronic corpora} consist of time-stamped documents spanning extensive time periods. They support retrospective retrieval, diachronic analysis, and event-based reasoning. Prominent examples include the \emph{New York Times Annotated Corpus} (1987--2007; 1.8M articles) \cite{sandhaus2008new}, which serves as the basis for the \texttt{ArchivalQA} dataset \cite{wang2022archivalqa}. Another widely used resource is the \emph{CNN/Daily Mail corpus} (2007--2015; 313K articles) \cite{NIPS2015__ccn/dailynews}, used in datasets such as \texttt{NewsQA} \cite{trischler-etal-2017-newsqa}. The \emph{Chronicling America} collection (1800--1920) offers digitized historical newspaper articles and supports long-range historical QA via \texttt{ChroniclingAmericaQA} \cite{piryani2024chroniclingamericaqa}. More recently, the \emph{Newswire} corpus \cite{NEURIPS2024_58f52a01_newswire} extends temporal coverage, providing 2.7 million newswire articles published between 1878 and 1977. It is enriched with metadata including geo-referenced datelines, Wikipedia/Wikidata entity links, and topical annotations, enabling fine-grained historical and spatio-temporal modeling. Another widely used corpus is \emph{CUSTOMNEWS} \cite{NEURIPS2021_customnews} (1969--2019), which consists of crawled English news sources spanning diverse domains such as politics, finance, and sports. Diachronic corpora are also used in related temporal tasks, including semantic drift detection \cite{hamilton-etal-2016-diachronic}, event burst modeling \cite{radinsky2013mining}, and timeline construction \cite{gutehrle-etal-2022-archive}.

\textbf{Synchronic corpora} represent a coherent snapshot of the world at a specific point in time. Unlike diachronic corpora, which typically span decades or years, synchronic collections capture a temporally aligned view, sometimes in conjunction with structured KBs. Wikipedia articles \cite{vrandevcic2014wikidata}, for example, reflect a particular version of world knowledge at a specific time (when the dump was made) and can be linked to Wikidata timestamps. Datasets such as \texttt{TimeQA} \cite{chen2dataset}, \texttt{TEMPREASON} \cite{tan-etal-2023-towards}, and \texttt{ComplexTempQA} \cite{gruber2024complextempqa} build on Wikipedia snapshots to support temporally-scoped QA grounded in a time-specific context. 

Finally, \textbf{Annotated temporal corpora} with explicit temporal annotations enable more structured temporal reasoning. \emph{TimeBank} \cite{pustejovsky2003timebank} introduced TimeML to annotate temporal expressions, events, and their temporal relations. Follow-up datasets like \texttt{WikiWars} \cite{mazur-dale-2010-wikiwars} and \texttt{RED} \cite{ogorman-etal-2016-richer} extended this framework to historical narratives and causal relations, respectively. These corpora serve as gold-standard resources for temporal tagging, relation extraction, and for developing models capable of explicit temporal reasoning over text. Unlike diachronic and synchronic corpora, these resources emphasize explicit temporal structure, making them particularly suitable for modeling fine-grained temporal relations.

%\vspace{-2mm}
\subsection{TQA Datasets}
\label{sec:tqa_dataset}

TQA datasets enable the evaluation of how well systems can answer questions that require temporal understanding and reasoning. They vary along multiple dimensions, including \textbf{Knowledge Source}, \textbf{Temporal Orientation}, \textbf{Temporal Explicitness}, and \textbf{Reasoning Complexity}. Table \ref{tab:tqa_datasets} provides a comprehensive comparison of major TQA datasets across these dimensions.

\textbf{Knowledge Source:} TQA datasets are commonly derived from diachronic or synchronic corpora. \textbf{Primary source datasets} derive from diachronic corpora, providing contemporaneous accounts written when events occurred. These datasets test models' ability to retrieve and reason over temporally anchored document collections with authentic historical perspectives. Datasets such as \texttt{NewsQA} \cite{trischler-etal-2017-newsqa}, \texttt{TDDiscourse} \cite{naik-etal-2019-tddiscourse}, \texttt{TORQUE} \cite{ning-etal-2020-torque}, \texttt{ArchivalQA} \cite{wang2022archivalqa}, \texttt{TKGQA} \cite{ong2023tkgqa},  \texttt{ChroniclingAmericaQA} \cite{piryani2024chroniclingamericaqa} are curated from historical news sources and support temporal reasoning over period-specific content. 

In contrast, synchronic corpora such as Wikipedia constitute \textbf{Secondary sources}, as they provide a retrospective view of the past. They have been used to build datasets such as \texttt{TimeQA} \cite{chen2dataset}, \texttt{TEMPREASON} \cite{tan-etal-2023-towards}, \texttt{TiQ} \cite{jia2024tiq}, and \texttt{ComplexTempQA} \cite{gruber2024complextempqa}, which support fine-grained reasoning across temporally scoped, consistent knowledge bases.

\textbf{Temporal Orientation:} While most datasets focus on past events, future-oriented QA datasets remain relatively rare. Still, they are increasingly important for evaluating models’ ability to perform predictive and hypothetical reasoning. \texttt{ForecastQA} \cite{jin-etal-2021-forecastqa}, \texttt{FutureContext} \cite{mutschlechner2025analyzing}, and \texttt{TimeBench} \cite{chu-etal-2024-timebench} are among the few benchmarks that include questions about future events, testing models' ability to perform timeline projections and forecast-based inference.

\textbf{Question Type:} Temporal questions can be broadly classified by their explicitness in referencing time.
Datasets like \texttt{TimeQA} \cite{chen2dataset}, \texttt{SituatedQA} \cite{zhang-choi-2021-situatedqa} and \texttt{TempQuestions} \cite{jia2018tempquestions} contain \textbf{Explicit Temporal Questions} with clear temporal markers, such as \textit{"What happened in 1947?"}, signaling temporal intent directly.

In contrast, \textbf{Implicit Temporal Questions} omit direct time references but still require temporal inference. For instance, \textit{"Who was Prime Minister of the UK when the Berlin Wall fell?"} requires inferring the date of the event and linking it to a temporally relevant fact. Datasets such as \texttt{TiQ} \cite{jia2024tiq} and \texttt{TORQUE} \cite{ning-etal-2020-torque} focus on implicit reasoning, testing event-event and event-time relationships. Other datasets, such as \texttt{ArchivalQA} \cite{wang2022archivalqa}, %\texttt{TemporalQuestions} \cite{wang2021improving}, 
and \texttt{ComplexTempQA} \cite{gruber2024complextempqa} combine both question types, offering a spectrum of temporal reasoning demands from explicit, time-anchored queries to implicit, event-based inference. % from surface-level retrieval to deep temporal inference.

\textbf{Temporal Reasoning Complexity:}
TQA tasks also vary in the depth of reasoning they require. \textbf{Simple Temporal Questions} typically involve direct lookups, such as identifying the date of a specific event or the state of the world at a given time. Early datasets like \texttt{NewsQA} \cite{trischler-etal-2017-newsqa} and \texttt{TempLAMA} \cite{dhingra-etal-2022-time} largely belong to this category. 
In contrast, \textbf{Complex Temporal Questions} demand more intricate processing, such as multi-hop reasoning, temporal filtering, or synthesizing information across events. For example, the question \textit{"What major international agreements were signed after World War I but before World War II?"} necessitates multi-hop temporal reasoning and contextual comparison. Datasets like \texttt{MenatQA} \cite{wei-etal-2023-menatqa}, \texttt{TempReason} \cite{tan-etal-2023-towards}, \texttt{Complex-TR} \cite{tan-etal-2024-towards}, and \texttt{ComplexTempQA} \cite{gruber2024complextempqa} are explicitly designed to evaluate these advanced reasoning capabilities. Others like \texttt{TimeBench} \cite{chu-etal-2024-timebench} span both simple and complex reasoning levels, including tasks such as timeline construction or event duration inference. %Table 

\textbf{Comparative perspective across TQA datasets:}
TQA benchmarks exhibit systematic trade-offs depending on their construction and source. \textit{Answer drift varies by dataset type}: crowdsourced datasets (e.g., \texttt{SituatedQA}, \texttt{FRESHQA}) mitigate ambiguity through explicit temporal clarification, whereas automatically generated datasets (e.g., \texttt{ArchivalQA}) better capture factual change over time, including entity evolution. 

\textit{Source type further shapes evaluation}: news-based, primary-source datasets emphasize temporal robustness under evolving evidence and event-centric reasoning, whereas Wikipedia-based datasets offer broader topical coverage with more stable, retrospective temporal framing. Finally, \textit{construction methods reflect scale–quality trade-offs}: crowdsourced datasets better capture natural language variation and implicit temporal intent, while automatically generated benchmarks enable large-scale evaluation of temporal sensitivity. A dataset selection guide summarizing these complementary strengths is provided in Table~\ref{tab:dataset_selection_guide} in the Appendix.

\paragraph{Summary:} TQA poses challenges that go beyond annotation quality and dataset scale. Unlike static QA tasks, the answers to temporal questions can shift over time, necessitating benchmarks that can account for answer drift and temporal volatility. Recent work such as RecencyQA \cite{piryani2026often} addresses this limitation by introducing a recency–stationarity taxonomy, enabling fine-grained evaluation of how answer validity evolves over time and across contexts. However, most existing datasets remain static snapshots: notable exceptions include \texttt{RealTimeQA} \cite{realtimeqa} and \texttt{FreshQA} \cite{vu-etal-2024-freshllms}, which incorporate periodic updates, though these updates are often resource-intensive, While datasets like \texttt{PATQA} \cite{meem-etal-2024-pat} explore scalable, automated update mechanisms. \textit{A persistent issue is temporal ambiguity} \cite{piryani-etal-2024-detecting}, where missing or implicit time references hinder both annotation and evaluation. Structural biases also emerge: \textit{diachronic corpora tend to have more data surrounding major events and contain detailed information on past events}, yet they reflect contemporaneous perspectives. In contrast, \textit{synchronic corpora provide broader coverage but often lack fine-grained event granularity.} Annotation strategies further involve trade-offs: \textit{crowdsourced datasets are typically small but high-quality, whereas automatically generated datasets are larger but noisier}. Moreover, most benchmarks fail to isolate specific temporal reasoning skills, such as duration inference or event ordering, thereby limiting diagnostic evaluation.

% While this section focuses on general-domain datasets, specialized domains such as medical, legal, and financial contexts present unique temporal reasoning challenges and dataset requirements, as shown in Appendix~\ref{appendix:domain_specific_QA}. 

While this section focuses on general-domain datasets, specialized domains such as medical, legal, and financial contexts introduce additional temporal reasoning challenges and require tailored dataset designs, as discussed in Appendix~\ref{appendix:domain_specific_QA}.

\section{Temporal Prediction Tasks}
\label{sec:temporal_prediction_tasks}

Temporal prediction tasks play a central role in the development of time-aware IR and QA systems. They focus on inferring implicit or missing temporal information from text, thereby improving alignment between queries, documents, and events. These tasks are especially important when explicit temporal metadata is sparse, noisy, or unavailable, and they support applications such as historical search, timeline construction, and time-sensitive retrieval.

Key tasks include \textbf{Event Dating}, \textbf{Document Dating}, \textbf{Focus Time Estimation}, \textbf{Query Time Profiling}, and \textbf{Event Occurrence Prediction}. Traditional methods rely on statistical language models and handcrafted rules, while more recent techniques employ transformer-based encoders, temporal embeddings, and graph-based reasoning to improve generalization and robustness \cite{yang-etal-2023-upon, abdallah2025tempretriever, liu-quan-2025-retrieval, yang-etal-2024-enhancing-temporal}. For a detailed review of task definitions, representative techniques, and evaluation strategies, we refer readers to Appendix~\ref{appendix:temp_prediction_task}.

%\vspace{-2mm}
\section{Approaches in Temporal QA}
\label{sec:temporal_qa_approaches}

TQA approaches differ in how they model temporal information in questions, retrieve time-relevant evidence, and reason over evolving knowledge. A wide range of methods has been developed to address these challenges, ranging from early rule-based and statistical systems to neural architectures and large language models (LLMs). Tables~\ref{tab:temporal_qa_part1} and~\ref{tab:temporal_qa_part2} (see Appendix) provide a comprehensive comparison of temporal QA approaches across architectural paradigms, temporal representations, and methodological strategies. They highlight differences in temporal modeling, reasoning strategies, and adaptation to changing world knowledge. These approaches differ in how they address the interaction between corpus, question, and model temporality introduced earlier.

% Many TQA systems additionally rely on auxiliary \textbf{temporal prediction tasks} that support temporal reasoning and representation, particularly when explicit temporal metadata is sparse or unreliable. These include \textbf{Event Dating}, \textbf{Document Dating}, \textbf{Document Focus Time Estimation}, \textbf{Query Time Profiling}, and \textbf{Event Occurrence Prediction}. Such tasks underpin downstream applications, including timeline construction, historical information retrieval, and time-sensitive question answering.

Many TQA systems additionally rely on auxiliary temporal prediction tasks (Section~\ref{sec:temporal_prediction_tasks}) to infer missing or implicit time information, particularly when explicit metadata is sparse or unreliable.

In this section, we focus on neural approaches, particularly temporal language models (TLMs), RAG, and reasoning for TQA. Detailed discussions of traditional rule-based and statistical QA methods are provided in Appendix~\ref{sec:temporal_rulebased}.

%\vspace{-15mm}
\subsection{Temporal Language Models}
\label{sec:time_aware_LM}

Deep learning has significantly advanced TQA by enabling models to capture temporal dependencies and contextual dynamics. Recent research has led to the development of \textbf{TLMs} that explicitly incorporate temporal signals during pretraining or fine-tuning. Models such as TempoT5 \cite{dhingra-etal-2022-time}, TempoBERT \cite{rosin2022time}, and BiTimeBERT \cite{wang2023bitimebert} integrate timestamps and temporal expressions into the training process, improving temporal generalization across downstream tasks, including semantic change detection and TQA. 
%inputs or used time-focused pretraining tasks. These strategies improve temporal generalization in downstream tasks such as semantic change detection and TQA.  

Beyond explicit timestamp conditioning, a second line of work introduces architectural and objective-level modifications to better internalize temporal structure. Approaches such as TALM \cite{ren-etal-2023-time}, SG-TLM \cite{su-etal-2023-efficient}, Temporal Span Masking \cite{cole-etal-2023-salient}, and Temporal Attention \cite{rosin-radinsky-2022-temporal} encourage models to attend to time-related cues, durations, and event structure through specialized masking, hierarchical representations, or time-aware attention mechanisms.

Some TLMs are designed for generation tasks that require explicit temporal grounding. For instance, \citet{cao-wang-2022-time} proposes temporal prompts, both textual and continuous vector-based, that guide generation with time-specific context. TCQA \cite{son-oh-2023-time} introduces a synthetic QA dataset and a span-selection task that aligns answers with their temporal context, enabling models to simulate historical reasoning and maintain timeline consistency.

\begin{tcolorbox}[
    colback=gray!3!white, 
    colframe=customblue, 
    boxsep=0.1mm,
    boxrule=0.3mm,
    width=\columnwidth,
]
\footnotesize
\textit{A common trend across TLMs is the shift from treating time as an auxiliary signal to making it a core component of the modeling process. Techniques such as timestamp conditioning, temporal pretraining tasks, and time-aware attention have improved temporal reasoning. However, most models still rely on static corpora and struggle with vague or implicit temporal cues. In practice, timestamp-conditioned or QA-supervised models align better with TQA than semantic change approaches focused on distributional drift. While these models represent a significant step forward, they still struggle with real-world temporal drift and complex reasoning.}
\end{tcolorbox}

\subsection{Temporal RAG}
\label{sec:temporal_rag}

While TLMs improve temporal understanding through pretraining, they remain constrained by the static nature of their training data. To address evolving information needs and mitigate temporal hallucinations, recent work has turned to \textbf{RAG} frameworks that couple neural retrieval with generation, enabling models to incorporate up-to-date, time-relevant evidence at inference time.

A key trend in this space is the integration of temporal signals directly into the retrieval process. Models such as TempRetriever \cite{abdallah2025tempretriever} and TsContriever \cite{wu2024time} extend dense retrievers to account for temporal relevance by encoding queries and passages with timestamp-aware embeddings. These approaches improve alignment between temporally scoped questions and temporally valid evidence. Similarly, TempRALM \cite{gade2024itsabouttime} enhances dense retrieval with temporal constraints to reduce factual drift and improve recency-grounded responses.

Beyond modifying retrieval scoring, several systems re-architect the retrieval–generation pipeline itself. TimeR4 \cite{qian-etal-2024-timer4} introduces a four-stage \textit{Retrieve–Rewrite–Retrieve–Rerank} framework that transforms underspecified temporal queries into explicitly time-anchored formulations before retrieving and re-ranking documents based on temporal fit. MRAG \cite{siyue2024mrag} further extends this direction by performing multi-hop reasoning across events, retrieving from multiple time-scoped sources, and aggregating cross-temporal evidence for complex QA. More recently, TimeRAG \cite{wang2025timerag} proposes an iterative temporal–semantic query decomposition and time-aware answer generation, enabling multi-step retrieval and reasoning over rapidly evolving information.

Compared to timestamp-conditioned language models, temporal RAG systems are most effective when knowledge is volatile, temporal intent is implicit, or answers depend on recent or evolving facts. In temporal RAG pipelines, retrieval quality is often the primary bottleneck for TQA. Unlike standard QA, temporal questions require evidence that is both topically relevant and temporally aligned with the query. Retrieval can fail due to noisy or missing timestamps, recency bias, or mismatches between document creation time and the event time described in the text. This propagates errors downstream and yields temporally inconsistent answers. Despite recent advances, robust temporal retrieval remains an open challenge, especially for queries with implicit temporal intent or noisy metadata.

\begin{tcolorbox}[
    colback=gray!3!white, 
    colframe=customblue, 
    boxsep=0.1mm,
    boxrule=0.3mm,
    width=\columnwidth,
]
\footnotesize

\textit{Together, these models signal a shift from static temporal representations to adaptive, retrieval-grounded temporal reasoning. While temporal RAG systems mitigate hallucinations and better reflect evolving knowledge, their effectiveness ultimately depends on reliable and temporally consistent evidence.}
\end{tcolorbox}

\subsection{Temporal Reasoning Capabilities}
\label{sec:temporal_reasoning}

While TLMs enhance time-aware representation, they remain largely limited to surface-level temporal associations. Many TQA tasks instead require \textbf{explicit temporal reasoning}-tracking event sequences, resolving durations, performing temporal arithmetic, and drawing logical inferences across time. These capabilities are essential for questions involving temporal transitions, causality, or evolving narratives, where retrieval alone cannot ensure temporally coherent answers.

Recent approaches have therefore sought to strengthen the reasoning capacity of pretrained language models (PLMs) through architectural innovations and specialized objectives. ECONET \cite{han-etal-2021-econet} pioneers continual temporal adaptation, preserving event consistency as models are updated with new information. TIMERS \cite{mathur-etal-2021-timers} and ConTempo \cite{niu-etal-2024-contempo} move beyond token-level reasoning, introducing structured inference layers that explicitly model temporal graphs, event hierarchies, and symbolic relations between actions. Despite these advances, such methods remain brittle in practice, improving local temporal ordering while still failing at multi-hop or abstract reasoning that spans long time.

Recent work has explored structured and logic-enhanced reasoning for text-based TQA, including temporal graph modeling over documents \cite{chu-etal-2023-mtger}, executable program induction for time-sensitive question answering \cite{zhu-etal-2023-question}, and logic- or timeline-based reasoning frameworks that explicitly formalize temporal inference for LLMs \cite{li2023unlocking, wu2025chainoftimeline}.

To expose these weaknesses more systematically, new benchmarks now evaluate LLMs along multiple reasoning dimensions. TRAM \cite{wang-zhao-2024-tram} assesses event frequency, duration estimation, and timeline ordering, consistently revealing large performance gaps between humans and even the strongest models such as GPT-4. Test of Time (ToT) \cite{fatemi2024test} isolates reasoning from memorization through synthetic tasks involving logical deduction, date arithmetic, and counterfactual inference—showing that models’ temporal accuracy often collapses once factual recall cues are removed. TEMPO \cite{abdallah2026tempo} focuses on reasoning-intensive temporal retrieval. Complementary benchmarks, such as TODAY \cite{feng-etal-2023-generic} and Narrative-of-Thought \cite{zhang-etal-2024-narrative}, further show that LLMs struggle to maintain temporal coherence over evolving contexts. 
% Recent work, such as RecencyQA \cite{piryani2026often}, highlights that non-stationary questions—where recency depends on context—are particularly challenging, with performance degrading as update frequency increases.

Beyond accuracy, \textbf{temporal robustness} has emerged as a critical yet underexplored weakness. Studies such as \cite{wallat2024temporal, wallat2025study} identify persistent \textit{temporal blind spots}, where even small perturbations to timestamps or event order cause sharp drops in performance. These findings suggest that models rely on shallow lexical or positional cues rather than genuine temporal reasoning, lacking invariance to time-related perturbations and failing to generalize across temporal shifts.

\begin{tcolorbox}[
    colback=gray!3!white, 
    colframe=customblue, 
    boxsep=0.1mm,
    boxrule=0.3mm,
    width=\columnwidth,
]
\footnotesize

\textit{Despite steady progress, current systems remain far from robust temporal reasoners. A core limitation is that most architectures entangle temporal reasoning with retrieval or pattern matching, rather than representing time as an abstract, manipulable variable. They exhibit limited understanding of implicit temporal cues, weak abstraction beyond explicit dates, and poor generalization under temporal drift. Bridging these gaps will require integrating symbolic and logical reasoning modules, incorporating explicit event memory, and designing evaluation frameworks that can disentangle genuine reasoning competence from pattern recall or retrieval bias.}

\end{tcolorbox}

\paragraph{Summary:} TQA has evolved from rule-based pipelines to neural architectures that integrate temporal information across input representations, model internals, and retrieval mechanisms. 

\textit{TLMs improved contextualization by embedding timestamps and temporal cues, yet they remain limited by static training corpora and an over-reliance on explicit time expressions, often failing on vague or underspecified temporal references. \emph{Temporal RAG systems} mitigate these issues by dynamically retrieving time-sensitive evidence at inference, but their performance depends critically on the temporal quality, relevance, and freshness of retrieved content. Building on these advances, recent \textit{reasoning-oriented architectures} extend capabilities toward event ordering, temporal inference, and multi-hop reasoning, yet even state-of-the-art models exhibit temporal blind spots and struggle to generalize across shifting temporal contexts.
}

Collectively, the paradigms alleviate specific weaknesses of their predecessors, but none fully resolves the core challenges of temporal ambiguity, factual drift, and reasoning consistency. Future research must systematically target these failure modes, balancing parametric and retrieved knowledge, handling implicit or underspecified timeframes, and developing evaluation methodologies that disentangle genuine temporal reasoning from statistical pattern matching.

\section{Future Research Directions}
\label{sec:future}

Despite recent advances in neural temporal reasoning, current TQA systems face fundamental limitations that require targeted research. We identify the following critical areas where future work is needed to build temporally aware, trustworthy QA systems. Detailed technical challenges and potential solutions for each direction are in Appendix~\ref{appendix:detailed_future_research_direction}.

\paragraph{Dynamic Temporal Knowledge Management.} Current TQA systems rely on static corpora, making them unable to respond to fast-evolving information needs. The temporal propagation problem, where small updates disrupt temporal dependencies across related events, exposes fundamental architectural limitations. Future systems must move beyond isolated fact updates toward scalable frameworks that track, edit, and reason over temporal dependencies in real time.

\paragraph{Temporally aware LLM agents.} While LLMs demonstrate impressive general reasoning capabilities, they consistently underperform in temporal reasoning tasks. LLMs exhibit temporal hallucinations and fail to resolve context-dependent expressions such as "last Tuesday" or "since our previous discussion."  Incorporating timeline tracking, event memory, and temporal reference resolution is essential for temporally coherent dialogue agents.

\paragraph{Diachronic and Synchronic Knowledge Integration.} Temporal questions often require combining time-sensitive sources that capture change over time with stable sources summarizing information at specific points. Most current systems treat these separately, limiting their ability to answer questions involving both historical trends and current facts. Future systems should develop temporal alignment algorithms and cross-source reasoning frameworks.

\paragraph{Temporal Uncertainty and Confidence Modeling.} Many historical events have unclear start and end dates, yet most TQA systems treat all dates as exact. This mismatch creates problems when systems give confident answers about uncertain information, with issues compounding in multi-step reasoning. Future systems must explicitly model temporal uncertainty and provide confidence for temporal answers.

\paragraph{Multilingual and Multimodal Temporal QA.} Temporal expressions, date formats, and cultural references vary widely across languages and modalities, yet most TQA systems are developed primarily for English textual input. Current systems struggle with diverse temporal signals due to limited cultural grounding and inadequate multimodal integration. Future research should focus on developing multilingual temporal tagging models and cross-modal alignment techniques.

\paragraph{Implicit Temporal Intent Understanding.} Many temporal questions hide their time requirements rather than stating them directly. Current systems struggle with this ambiguity because they lack mechanisms to infer intended timeframes from contextual clues or shared knowledge. As users increasingly expect models to infer nuanced, context-sensitive answers, detecting hidden temporal assumptions is critical to avoid plausible but misleading outputs.

% \paragraph{Evaluation and Benchmarking for Temporal Reasoning.} Standard metrics like accuracy and F1 fall short in capturing temporal coherence or reasoning depth. Developing temporally aware evaluation metrics remains an important challenge for effectively assessing and comparing future systems.

\paragraph{Evaluation and Benchmarking for Temporal Reasoning.} Standard metrics like accuracy and F1 fall short capturing temporal coherence or reasoning depth. Developing temporally aware evaluation metrics remains an important challenge for effectively assessing and comparing future systems.

\section{Conclusion}
TQA plays a critical role in retrieving and reasoning over time-sensitive information in dynamic contexts. %In this survey, 
We focus on recent advancements in the field, particularly modern TLMs and RAG approaches, examining core challenges including temporal expression recognition, event ordering, and implicit temporal reasoning. Despite advances, current systems struggle with answer drift, temporal uncertainty, and limited temporal granularity, while most benchmarks rely on static snapshots that hinder real-time and future-oriented reasoning. Viewed through the interaction among corpus temporality, question temporality, and model capabilities, these limitations highlight the need for more robust temporal representations, dynamic evaluation protocols, and adaptive learning mechanisms.

\section*{Limitations}
This survey aims to provide a comprehensive overview of TQA. There are a few important limitations to acknowledge.

We made our best efforts to be thorough, but it is possible that some relevant works may have been missed. We conducted an extensive literature review using forward and backward snowballing techniques, with particular attention to papers published in major venues such as  ACL, SIGIR, EMNLP, NeurIPS, ECIR, CIKM and preprints on arXiv. On the other hand, due to page limitations, we provide only a brief summary of each method, without providing exhaustive technical details.  

\section*{Acknowledgments}

The authors would like to acknowledge the financial support provided by the Austrian Research Agency (FFG) for the project “AI Enabled Sustainability Jurisdiction Demonstrator” (Project No. 915229).


\bibliography{custom}

@article{zhu2021retrieving,
  title={Retrieving and reading: A comprehensive survey on open-domain question answering},
  author={Zhu, Fengbin and Lei, Wenqiang and Wang, Chao and Zheng, Jianming and Poria, Soujanya and Chua, Tat-Seng},
  journal={arXiv preprint arXiv:2101.00774},
  year={2021}
}

@article{li2024matching,
author = {Li, Xiaoxi and Jin, Jiajie and Zhou, Yujia and Zhang, Yuyao and Zhang, Peitian and Zhu, Yutao and Dou, Zhicheng},
title = {From Matching to Generation: A Survey on Generative Information Retrieval},
year = {2025},
publisher = {Association for Computing Machinery},
address = {New York, NY, USA},
issn = {1046-8188},
url = {https://doi.org/10.1145/3722552},
doi = {10.1145/3722552},
note = {Just Accepted},
journal = {ACM Trans. Inf. Syst.},
month = mar
}

@inproceedings{kanhabua2016temporal,
author = {Kanhabua, Nattiya and Anand, Avishek},
title = {Temporal Information Retrieval},
year = {2016},
isbn = {9781450340694},
publisher = {Association for Computing Machinery},
address = {New York, NY, USA},
url = {https://doi.org/10.1145/2911451.2914805},
doi = {10.1145/2911451.2914805},
booktitle = {Proceedings of the 39th International ACM SIGIR Conference on Research and Development in Information Retrieval},
pages = {1235–1238},
numpages = {4},
location = {Pisa, Italy},
series = {SIGIR '16}
}

@inproceedings{piryani2024chroniclingamericaqa,
author = {Piryani, Bhawna and Mozafari, Jamshid and Jatowt, Adam},
title = {ChroniclingAmericaQA: A Large-scale Question Answering Dataset based on Historical American Newspaper Pages},
year = {2024},
isbn = {9798400704314},
publisher = {Association for Computing Machinery},
address = {New York, NY, USA},
url = {https://doi.org/10.1145/3626772.3657891},
doi = {10.1145/3626772.3657891},
booktitle = {Proceedings of the 47th International ACM SIGIR Conference on Research and Development in Information Retrieval},
pages = {2038–2048},
numpages = {11},
location = {Washington DC, USA},
series = {SIGIR '24}
}

@inproceedings{wei-etal-2023-menatqa,
    title = "{M}enat{QA}: A New Dataset for Testing the Temporal Comprehension and Reasoning Abilities of Large Language Models",
    author = "Wei, Yifan  and
      Su, Yisong  and
      Ma, Huanhuan  and
      Yu, Xiaoyan  and
      Lei, Fangyu  and
      Zhang, Yuanzhe  and
      Zhao, Jun  and
      Liu, Kang",
    editor = "Bouamor, Houda  and
      Pino, Juan  and
      Bali, Kalika",
    booktitle = "Findings of the Association for Computational Linguistics: EMNLP 2023",
    month = dec,
    year = "2023",
    address = "Singapore",
    publisher = "Association for Computational Linguistics",
    url = "https://aclanthology.org/2023.findings-emnlp.100/",
    doi = "10.18653/v1/2023.findings-emnlp.100",
    pages = "1434--1447"
}

@inproceedings{chen2dataset,
  title={A Dataset for Answering Time-Sensitive Questions},
  author={Chen, Wenhu and Wang, Xinyi and Wang, William Yang},
  booktitle={Thirty-fifth Conference on Neural Information Processing Systems Datasets and Benchmarks Track (Round 2)},
  year= "2021",
}

@inproceedings{ning-etal-2020-torque,
    title = "{TORQUE}: A Reading Comprehension Dataset of Temporal Ordering Questions",
    author = "Ning, Qiang  and
      Wu, Hao  and
      Han, Rujun  and
      Peng, Nanyun  and
      Gardner, Matt  and
      Roth, Dan",
    editor = "Webber, Bonnie  and
      Cohn, Trevor  and
      He, Yulan  and
      Liu, Yang",
    booktitle = "Proceedings of the 2020 Conference on Empirical Methods in Natural Language Processing (EMNLP)",
    month = nov,
    year = "2020",
    address = "Online",
    publisher = "Association for Computational Linguistics",
    url = "https://aclanthology.org/2020.emnlp-main.88/",
    doi = "10.18653/v1/2020.emnlp-main.88",
    pages = "1158--1172"
}

@inproceedings{tan-etal-2023-towards,
    title = "Towards Benchmarking and Improving the Temporal Reasoning Capability of Large Language Models",
    author = "Tan, Qingyu  and
      Ng, Hwee Tou  and
      Bing, Lidong",
    editor = "Rogers, Anna  and
      Boyd-Graber, Jordan  and
      Okazaki, Naoaki",
    booktitle = "Proceedings of the 61st Annual Meeting of the Association for Computational Linguistics (Volume 1: Long Papers)",
    month = jul,
    year = "2023",
    address = "Toronto, Canada",
    publisher = "Association for Computational Linguistics",
    url = "https://aclanthology.org/2023.acl-long.828/",
    doi = "10.18653/v1/2023.acl-long.828",
    pages = "14820--14835",
}

@inproceedings{jia2024tiq,
author = {Jia, Zhen and Christmann, Philipp and Weikum, Gerhard},
title = {TIQ: A Benchmark for Temporal Question Answering with Implicit Time Constraints},
year = {2024},
isbn = {9798400701726},
publisher = {Association for Computing Machinery},
address = {New York, NY, USA},
url = {https://doi.org/10.1145/3589335.3651895},
doi = {10.1145/3589335.3651895},
booktitle = {Companion Proceedings of the ACM Web Conference 2024},
pages = {1394–1399},
numpages = {6},
location = {Singapore, Singapore},
series = {WWW '24}
}

@inproceedings{chu-etal-2024-timebench,
    title = "{T}ime{B}ench: A Comprehensive Evaluation of Temporal Reasoning Abilities in Large Language Models",
    author = "Chu, Zheng  and
      Chen, Jingchang  and
      Chen, Qianglong  and
      Yu, Weijiang  and
      Wang, Haotian  and
      Liu, Ming  and
      Qin, Bing",
    editor = "Ku, Lun-Wei  and
      Martins, Andre  and
      Srikumar, Vivek",
    booktitle = "Proceedings of the 62nd Annual Meeting of the Association for Computational Linguistics (Volume 1: Long Papers)",
    month = aug,
    year = "2024",
    address = "Bangkok, Thailand",
    publisher = "Association for Computational Linguistics",
    url = "https://aclanthology.org/2024.acl-long.66/",
    doi = "10.18653/v1/2024.acl-long.66",
    pages = "1204--1228",
}

@inproceedings{jia2018tempquestions,
author = {Jia, Zhen and Abujabal, Abdalghani and Saha Roy, Rishiraj and Str\"{o}tgen, Jannik and Weikum, Gerhard},
title = {TempQuestions: A Benchmark for Temporal Question Answering},
year = {2018},
isbn = {9781450356404},
publisher = {International World Wide Web Conferences Steering Committee},
address = {Republic and Canton of Geneva, CHE},
url = {https://doi.org/10.1145/3184558.3191536},
doi = {10.1145/3184558.3191536},  
booktitle = {Companion Proceedings of the The Web Conference 2018},
pages = {1057–1062},
numpages = {6},
location = {Lyon, France},
series = {WWW '18}
}

@article{kumar2012dating,
  title={Dating texts without explicit temporal cues},
  author={Kumar, Abhimanu and Baldridge, Jason and Lease, Matthew and Ghosh, Joydeep},
  journal={arXiv preprint arXiv:1211.2290},
  year={2012}
}

@inproceedings{wang2021event,
author = {Wang, Jiexin and Jatowt, Adam and Yoshikawa, Masatoshi},
title = {Event Occurrence Date Estimation based on Multivariate Time Series Analysis over Temporal Document Collections},
year = {2021},
isbn = {9781450380379},
publisher = {Association for Computing Machinery},
address = {New York, NY, USA},
url = {https://doi.org/10.1145/3404835.3462885},
doi = {10.1145/3404835.3462885},
booktitle = {Proceedings of the 44th International ACM SIGIR Conference on Research and Development in Information Retrieval},
pages = {398–407},
numpages = {10},
location = {Virtual Event, Canada},
series = {SIGIR '21}
}

@inproceedings{shrivastava2017concept,
author = {Shrivastava, Shashank and Khapra, Mitesh and Chakraborti, Sutanu},
title = {A Concept Driven Graph Based Approach for Estimating the Focus Time of a Document},
year = {2017},
isbn = {978-3-319-71927-6},
publisher = {Springer-Verlag},
address = {Berlin, Heidelberg},
url = {https://doi.org/10.1007/978-3-319-71928-3_25},
doi = {10.1007/978-3-319-71928-3_25},
booktitle = {Mining Intelligence and Knowledge Exploration: 5th International Conference, MIKE 2017, Hyderabad, India, December 13–15, 2017, Proceedings},
pages = {250–260},
numpages = {11},
location = {Hyderabad, India}
}

@inproceedings{jatowt2013estimating,
author = {Jatowt, Adam and Au Yeung, Ching-Man and Tanaka, Katsumi},
title = {Estimating document focus time},
year = {2013},
isbn = {9781450322638},
publisher = {Association for Computing Machinery},
address = {New York, NY, USA},
url = {https://doi.org/10.1145/2505515.2505655},
doi = {10.1145/2505515.2505655},
booktitle = {Proceedings of the 22nd ACM International Conference on Information \& Knowledge Management},
pages = {2273–2278},
numpages = {6},
location = {San Francisco, California, USA},
series = {CIKM '13}
}

@inproceedings{dakka2008answering,
author = {Dakka, Wisam and Gravano, Luis and Ipeirotis, Panagiotis G.},
title = {Answering general time sensitive queries},
year = {2008},
isbn = {9781595939913},
publisher = {Association for Computing Machinery},
address = {New York, NY, USA},
url = {https://doi.org/10.1145/1458082.1458320},
doi = {10.1145/1458082.1458320},
booktitle = {Proceedings of the 17th ACM Conference on Information and Knowledge Management},
pages = {1437–1438},
numpages = {2},
location = {Napa Valley, California, USA},
series = {CIKM '08}
}

@article{jones2007temporal,
author = {Jones, Rosie and Diaz, Fernando},
title = {Temporal profiles of queries},
year = {2007},
issue_date = {July 2007},
publisher = {Association for Computing Machinery},
address = {New York, NY, USA},
volume = {25},
number = {3},
issn = {1046-8188},
url = {https://doi.org/10.1145/1247715.1247720},
doi = {10.1145/1247715.1247720},
journal = {ACM Trans. Inf. Syst.},
month = jul,
pages = {14–es},
numpages = {31}
}

@inproceedings{gupta2014identifying,
author = {Gupta, Dhruv and Berberich, Klaus},
title = {Identifying Time Intervals of Interest to Queries},
year = {2014},
isbn = {9781450325981},
publisher = {Association for Computing Machinery},
address = {New York, NY, USA},
url = {https://doi.org/10.1145/2661829.2661927},
doi = {10.1145/2661829.2661927},
booktitle = {Proceedings of the 23rd ACM International Conference on Conference on Information and Knowledge Management},
pages = {1835–1838},
numpages = {4},
location = {Shanghai, China},
series = {CIKM '14}
}

@inproceedings{das2017estimating,
author = {Das, Supratim and Mishra, Arunav and Berberich, Klaus and Setty, Vinay},
title = {Estimating Event Focus Time Using Neural Word Embeddings},
year = {2017},
isbn = {9781450349185},
publisher = {Association for Computing Machinery},
address = {New York, NY, USA},
url = {https://doi.org/10.1145/3132847.3133131},
doi = {10.1145/3132847.3133131},
booktitle = {Proceedings of the 2017 ACM on Conference on Information and Knowledge Management},
pages = {2039–2042},
numpages = {4},
location = {Singapore, Singapore},
series = {CIKM '17}
}

@inproceedings{morbidoni2018leveraging,
author = {Morbidoni, Christian and Cucchiarelli, Alessandro and Ursino, Domenico},
title = {Leveraging linked entities to estimate focus time of short texts},
year = {2018},
isbn = {9781450365277},
publisher = {Association for Computing Machinery},
address = {New York, NY, USA},
url = {https://doi.org/10.1145/3216122.3216158},
doi = {10.1145/3216122.3216158},
booktitle = {Proceedings of the 22nd International Database Engineering \& Applications Symposium},
pages = {282–286},
numpages = {5},
location = {Villa San Giovanni, Italy},
series = {IDEAS '18}
}

@inproceedings{Li-Croft,
author = {Li, Xiaoyan and Croft, W. Bruce},
title = {Time-based language models},
year = {2003},
isbn = {1581137230},
publisher = {Association for Computing Machinery},
address = {New York, NY, USA},
url = {https://doi.org/10.1145/956863.956951},
doi = {10.1145/956863.956951},
booktitle = {Proceedings of the Twelfth International Conference on Information and Knowledge Management},
pages = {469–475},
numpages = {7},
location = {New Orleans, LA, USA},
series = {CIKM '03}
}

@inproceedings{arikan2009time,
  title={Time will tell: Leveraging temporal expressions in ir},
  author={Arikan, Irem and Bedathur, Srikanta and Berberich, Klaus},
  booktitle={Second ACM International Conference on Web Search and Data Mining},
  year={2009},
  organization={ACM}
}

@inproceedings{honovich-etal-2020-machine,
    title = "Machine Reading of Historical Events",
    author = "Honovich, Or  and
      Torroba Hennigen, Lucas  and
      Abend, Omri  and
      Cohen, Shay B.",
    editor = "Jurafsky, Dan  and
      Chai, Joyce  and
      Schluter, Natalie  and
      Tetreault, Joel",
    booktitle = "Proceedings of the 58th Annual Meeting of the Association for Computational Linguistics",
    month = jul,
    year = "2020",
    address = "Online",
    publisher = "Association for Computational Linguistics",
    url = "https://aclanthology.org/2020.acl-main.668/",
    doi = "10.18653/v1/2020.acl-main.668",
    pages = "7486--7497"
}

@inproceedings{zhou-etal-2021-temporal,
    title = "Temporal Reasoning on Implicit Events from Distant Supervision",
    author = "Zhou, Ben  and
      Richardson, Kyle  and
      Ning, Qiang  and
      Khot, Tushar  and
      Sabharwal, Ashish  and
      Roth, Dan",
    editor = "Toutanova, Kristina  and
      Rumshisky, Anna  and
      Zettlemoyer, Luke  and
      Hakkani-Tur, Dilek  and
      Beltagy, Iz  and
      Bethard, Steven  and
      Cotterell, Ryan  and
      Chakraborty, Tanmoy  and
      Zhou, Yichao",
    booktitle = "Proceedings of the 2021 Conference of the North American Chapter of the Association for Computational Linguistics: Human Language Technologies",
    month = jun,
    year = "2021",
    address = "Online",
    publisher = "Association for Computational Linguistics",
    url = "https://aclanthology.org/2021.naacl-main.107/",
    doi = "10.18653/v1/2021.naacl-main.107",
    pages = "1361--1371",
}

@inproceedings{berberich2010language,
author = {Berberich, Klaus and Bedathur, Srikanta and Alonso, Omar and Weikum, Gerhard},
title = {A language modeling approach for temporal information needs},
year = {2010},
isbn = {3642122744},
publisher = {Springer-Verlag},
address = {Berlin, Heidelberg},
url = {https://doi.org/10.1007/978-3-642-12275-0_5},
doi = {10.1007/978-3-642-12275-0_5},
booktitle = {Proceedings of the 32nd European Conference on Advances in Information Retrieval},
pages = {13–25},
numpages = {13},
location = {Milton Keynes, UK},
series = {ECIR'2010}
}

@inproceedings{metzler2009improving,
author = {Metzler, Donald and Jones, Rosie and Peng, Fuchun and Zhang, Ruiqiang},
title = {Improving search relevance for implicitly temporal queries},
year = {2009},
isbn = {9781605584836},
publisher = {Association for Computing Machinery},
address = {New York, NY, USA},
url = {https://doi.org/10.1145/1571941.1572085},
doi = {10.1145/1571941.1572085},
booktitle = {Proceedings of the 32nd International ACM SIGIR Conference on Research and Development in Information Retrieval},
pages = {700–701},
numpages = {2},
location = {Boston, MA, USA},
series = {SIGIR '09}
}

@inproceedings{harabagiu2005question,
  title={Question answering based on temporal inference},
  author={Harabagiu, Sanda and Bejan, Cosmin Adrian},
  booktitle={Proceedings of the AAAI-2005 workshop on inference for textual question answering},
  pages={27--34},
  year={2005}
}

@inproceedings{jong2005temporal,
title = "Temporal Language Models for the Disclosure of Historical Text",
author = "de Jong, Franciska and Rode, Henning and Hiemstra, Djoerd",
year = "2005",
month = sep,
language = "Undefined",
isbn = "90-6984-456-7",
publisher = "Koninklijke Nederlandse Academie van Wetenschappen",
pages = "161--168",
booktitle = "Humanities, computers and cultural heritage: Proceedings of the XVIth International Conference of the Association for History and Computing (AHC 2005)",
}

@inproceedings{saquete2004splitting,
author = {Saquete, E. and Mart\'{\i}nez-Barco, P. and Mu\~{n}oz, R. and Vicedo, J. L.},
title = {Splitting complex temporal questions for question answering systems},
year = {2004},
publisher = {Association for Computational Linguistics},
address = {USA},
url = {https://doi.org/10.3115/1218955.1219027},
doi = {10.3115/1218955.1219027},
booktitle = {Proceedings of the 42nd Annual Meeting on Association for Computational Linguistics},
pages = {566–es},
location = {Barcelona, Spain},
series = {ACL '04}
}

@article{saquete2009enhancing,
author = {Saquete, Estela and Vicedo, Jose L. and Mart\'{\i}nez-Barco, Patricio and Mu\~{n}oz, Rafael and Llorens, Hector},
title = {Enhancing QA systems with complex temporal question processing capabilities},
year = {2009},
issue_date = {May 2009},
publisher = {AI Access Foundation},
address = {El Segundo, CA, USA},
volume = {35},
number = {1},
issn = {1076-9757},
journal = {Journal of Artificial Intelligence Research},
month = aug,
pages = {755–811},
numpages = {57}
}

@inproceedings{cole-etal-2023-salient,
    title = "Salient Span Masking for Temporal Understanding",
    author = "Cole, Jeremy R.  and
      Chaudhary, Aditi  and
      Dhingra, Bhuwan  and
      Talukdar, Partha",
    editor = "Vlachos, Andreas  and
      Augenstein, Isabelle",
    booktitle = "Proceedings of the 17th Conference of the European Chapter of the Association for Computational Linguistics",
    month = may,
    year = "2023",
    address = "Dubrovnik, Croatia",
    publisher = "Association for Computational Linguistics",
    url = "https://aclanthology.org/2023.eacl-main.222/",
    doi = "10.18653/v1/2023.eacl-main.222",
    pages = "3052--3060"
}

@inproceedings{su-etal-2023-efficient,
    title = "Efficient Continue Training of Temporal Language Model with Structural Information",
    author = "Su, Zhaochen  and
      Li, Juntao  and
      Zhang, Zikang  and
      Zhou, Zihan  and
      Zhang, Min",
    editor = "Bouamor, Houda  and
      Pino, Juan  and
      Bali, Kalika",
    booktitle = "Findings of the Association for Computational Linguistics: EMNLP 2023",
    month = dec,
    year = "2023",
    address = "Singapore",
    publisher = "Association for Computational Linguistics",
    url = "https://aclanthology.org/2023.findings-emnlp.418/",
    doi = "10.18653/v1/2023.findings-emnlp.418",
    pages = "6315--6329"
}

@inproceedings{rosin2022time,
author = {Rosin, Guy D. and Guy, Ido and Radinsky, Kira},
title = {Time Masking for Temporal Language Models},
year = {2022},
isbn = {9781450391320},
publisher = {Association for Computing Machinery},
address = {New York, NY, USA},
url = {https://doi.org/10.1145/3488560.3498529},
doi = {10.1145/3488560.3498529},
booktitle = {Proceedings of the Fifteenth ACM International Conference on Web Search and Data Mining},
pages = {833–841},
numpages = {9},
location = {Virtual Event, AZ, USA},
series = {WSDM '22}
}

@inproceedings{vashishth-etal-2018-dating,
    title = "Dating Documents using Graph Convolution Networks",
    author = "Vashishth, Shikhar  and
      Dasgupta, Shib Sankar  and
      Ray, Swayambhu Nath  and
      Talukdar, Partha",
    editor = "Gurevych, Iryna  and
      Miyao, Yusuke",
    booktitle = "Proceedings of the 56th Annual Meeting of the Association for Computational Linguistics (Volume 1: Long Papers)",
    month = jul,
    year = "2018",
    address = "Melbourne, Australia",
    publisher = "Association for Computational Linguistics",
    url = "https://aclanthology.org/P18-1149/",
    doi = "10.18653/v1/P18-1149",
    pages = "1605--1615"
}

@article{dhingra-etal-2022-time,
    title = "Time-Aware Language Models as Temporal Knowledge Bases",
    author = "Dhingra, Bhuwan  and
      Cole, Jeremy R.  and
      Eisenschlos, Julian Martin  and
      Gillick, Daniel  and
      Eisenstein, Jacob  and
      Cohen, William W.",
    editor = "Roark, Brian  and
      Nenkova, Ani",
    journal = "Transactions of the Association for Computational Linguistics",
    volume = "10",
    year = "2022",
    address = "Cambridge, MA",
    publisher = "MIT Press",
    url = "https://aclanthology.org/2022.tacl-1.15/",
    doi = "10.1162/tacl_a_00459",
    pages = "257--273",
}

@inproceedings{rosin-radinsky-2022-temporal,
    title = "Temporal Attention for Language Models",
    author = "Rosin, Guy D.  and
      Radinsky, Kira",
    editor = "Carpuat, Marine  and
      de Marneffe, Marie-Catherine  and
      Meza Ruiz, Ivan Vladimir",
    booktitle = "Findings of the Association for Computational Linguistics: NAACL 2022",
    month = jul,
    year = "2022",
    address = "Seattle, United States",
    publisher = "Association for Computational Linguistics",
    url = "https://aclanthology.org/2022.findings-naacl.112/",
    doi = "10.18653/v1/2022.findings-naacl.112",
    pages = "1498--1508"
}

@inproceedings{yang-etal-2024-enhancing-temporal,
    title = "Enhancing Temporal Sensitivity and Reasoning for Time-Sensitive Question Answering",
    author = "Yang, Wanqi  and
      Li, Yanda  and
      Fang, Meng  and
      Chen, Ling",
    editor = "Al-Onaizan, Yaser  and
      Bansal, Mohit  and
      Chen, Yun-Nung",
    booktitle = "Findings of the Association for Computational Linguistics: EMNLP 2024",
    month = nov,
    year = "2024",
    address = "Miami, Florida, USA",
    publisher = "Association for Computational Linguistics",
    url = "https://aclanthology.org/2024.findings-emnlp.848/",
    doi = "10.18653/v1/2024.findings-emnlp.848",
    pages = "14495--14508"
}

@inproceedings{wang2022archivalqa,
author = {Wang, Jiexin and Jatowt, Adam and Yoshikawa, Masatoshi},
title = {ArchivalQA: A Large-scale Benchmark Dataset for Open-Domain Question Answering over Historical News Collections},
year = {2022},
isbn = {9781450387323},
publisher = {Association for Computing Machinery},
address = {New York, NY, USA},
url = {https://doi.org/10.1145/3477495.3531734},
doi = {10.1145/3477495.3531734},
booktitle = {Proceedings of the 45th International ACM SIGIR Conference on Research and Development in Information Retrieval},
pages = {3025–3035},
numpages = {11},
location = {Madrid, Spain},
series = {SIGIR '22}
}

@inproceedings{wang2023bitimebert,
author = {Wang, Jiexin and Jatowt, Adam and Yoshikawa, Masatoshi and Cai, Yi},
title = {BiTimeBERT: Extending Pre-Trained Language Representations with Bi-Temporal Information},
year = {2023},
isbn = {9781450394086},
publisher = {Association for Computing Machinery},
address = {New York, NY, USA},
url = {https://doi.org/10.1145/3539618.3591686},
doi = {10.1145/3539618.3591686},
booktitle = {Proceedings of the 46th International ACM SIGIR Conference on Research and Development in Information Retrieval},
pages = {812–821},
numpages = {10},
location = {Taipei, Taiwan},
series = {SIGIR '23}
}

@article{campos2014survey,
author = {Campos, Ricardo and Dias, Ga\"{e}l and Jorge, Al\'{\i}pio M. and Jatowt, Adam},
title = {Survey of Temporal Information Retrieval and Related Applications},
year = {2014},
issue_date = {January 2015},
publisher = {Association for Computing Machinery},
address = {New York, NY, USA},
volume = {47},
number = {2},
issn = {0360-0300},
url = {https://doi.org/10.1145/2619088},
doi = {10.1145/2619088},
journal = {ACM Computing Survey},
month = aug,
articleno = {15},
numpages = {41}
}

@inproceedings{kanhabua2008improving,
author = {Kanhabua, Nattiya and N\o{}rv\r{a}g, Kjetil},
title = {Improving Temporal Language Models for Determining Time of Non-timestamped Documents},
year = {2008},
isbn = {9783540875987},
publisher = {Springer-Verlag},
address = {Berlin, Heidelberg},
url = {https://doi.org/10.1007/978-3-540-87599-4_37},
doi = {10.1007/978-3-540-87599-4_37},
booktitle = {Proceedings of the 12th European Conference on Research and Advanced Technology for Digital Libraries},
pages = {358–370},
numpages = {13},
location = {Aarhus, Denmark},
series = {ECDL '08}
}

@inproceedings{kanhabua2010determining,
author = {Kanhabua, Nattiya and N\o{}rv\r{a}g, Kjetil},
title = {Determining time of queries for re-ranking search results},
year = {2010},
isbn = {3642154638},
publisher = {Springer-Verlag},
address = {Berlin, Heidelberg},
booktitle = {Proceedings of the 14th European Conference on Research and Advanced Technology for Digital Libraries},
url = {https://doi.org/10.1007/978-3-642-15464-5_27},
doi= {10.1007/978-3-642-15464-5_27},
pages = {261–272},
numpages = {12},
location = {Glasgow, UK},
series = {ECDL'10}
}

@inproceedings{han-etal-2021-econet,
    title = "{ECONET}: Effective Continual Pretraining of Language Models for Event Temporal Reasoning",
    author = "Han, Rujun  and
      Ren, Xiang  and
      Peng, Nanyun",
    editor = "Moens, Marie-Francine  and
      Huang, Xuanjing  and
      Specia, Lucia  and
      Yih, Scott Wen-tau",
    booktitle = "Proceedings of the 2021 Conference on Empirical Methods in Natural Language Processing",
    month = nov,
    year = "2021",
    address = "Online and Punta Cana, Dominican Republic",
    publisher = "Association for Computational Linguistics",
    url = "https://aclanthology.org/2021.emnlp-main.436/",
    doi = "10.18653/v1/2021.emnlp-main.436",
    pages = "5367--5380"
}

@inproceedings{wu2024time,
author = {Wu, Feifan and Liu, Lingyuan and He, Wentao and Liu, Ziqi and Zhang, Zhiqiang and Wang, Haofen and Wang, Meng},
title = {Time-Sensitve Retrieval-Augmented Generation for Question Answering},
year = {2024},
isbn = {9798400704369},
publisher = {Association for Computing Machinery},
address = {New York, NY, USA},
url = {https://doi.org/10.1145/3627673.3679800},
doi = {10.1145/3627673.3679800},
booktitle = {Proceedings of the 33rd ACM International Conference on Information and Knowledge Management},
pages = {2544–2553},
numpages = {10},
location = {Boise, ID, USA},
series = {CIKM '24}
}

@article{gade2024itsabouttime,
  title={It's About Time: Incorporating Temporality in Retrieval Augmented Language Models},
  author={Gade, Anoushka and Jetcheva, Jorjeta G},
  journal={CoRR},
  year={2024}
}

@inproceedings{jatowt2005temporal,
  title={Temporal ranking of search engine results},
  author={Jatowt, Adam and Kawai, Yukiko and Tanaka, Katsumi},
  booktitle={Web Information Systems Engineering--WISE 2005: 6th International Conference on Web Information Systems Engineering, New York, NY, USA, November 20-22, 2005. Proceedings 6},
  pages={43--52},
  url="https://link.springer.com/chapter/10.1007/11581062_4#citeas",
  year={2005},
  organization={Springer}
}

@inproceedings{dalli-2006-temporal,
    title = "Temporal Classification of Text and Automatic Document Dating",
    author = "Dalli, Angelo",
    editor = "Moore, Robert C.  and
      Bilmes, Jeff  and
      Chu-Carroll, Jennifer  and
      Sanderson, Mark",
    booktitle = "Proceedings of the Human Language Technology Conference of the {NAACL}, Companion Volume: Short Papers",
    month = jun,
    year = "2006",
    address = "New York City, USA",
    publisher = "Association for Computational Linguistics",
    url = "https://aclanthology.org/N06-2008/",
    pages = "29--32"
}

@inproceedings{jatowt2007detecting,
author = {Jatowt, Adam and Kawai, Yukiko and Tanaka, Katsumi},
title = {Detecting age of page content},
year = {2007},
isbn = {9781595938299},
publisher = {Association for Computing Machinery},
address = {New York, NY, USA},
url = {https://doi.org/10.1145/1316902.1316925},
doi = {10.1145/1316902.1316925},
booktitle = {Proceedings of the 9th Annual ACM International Workshop on Web Information and Data Management},
pages = {137–144},
numpages = {8},
location = {Lisbon, Portugal},
series = {WIDM '07}
}

@article{alonso2007value,
author = {Alonso, Omar and Gertz, Michael and Baeza-Yates, Ricardo},
title = {On the value of temporal information in information retrieval},
year = {2007},
issue_date = {December 2007},
publisher = {Association for Computing Machinery},
address = {New York, NY, USA},
volume = {41},
number = {2},
issn = {0163-5840},
url = {https://doi.org/10.1145/1328964.1328968},
doi = {10.1145/1328964.1328968},
journal = {SIGIR Forum},
month = dec,
pages = {35–41},
numpages = {7}
}

@inproceedings{alonso2009clustering,
author = {Alonso, Omar and Gertz, Michael and Baeza-Yates, Ricardo},
title = {Clustering and exploring search results using timeline constructions},
year = {2009},
isbn = {9781605585123},
publisher = {Association for Computing Machinery},
address = {New York, NY, USA},
url = {https://doi.org/10.1145/1645953.1645968},
doi = {10.1145/1645953.1645968},
booktitle = {Proceedings of the 18th ACM Conference on Information and Knowledge Management},
pages = {97–106},
numpages = {10},
location = {Hong Kong, China},
series = {CIKM '09}
}

@inproceedings{dong2010time,
author = {Dong, Anlei and Zhang, Ruiqiang and Kolari, Pranam and Bai, Jing and Diaz, Fernando and Chang, Yi and Zheng, Zhaohui and Zha, Hongyuan},
title = {Time is of the essence: improving recency ranking using Twitter data},
year = {2010},
isbn = {9781605587998},
publisher = {Association for Computing Machinery},
address = {New York, NY, USA},
url = {https://doi.org/10.1145/1772690.1772725},
doi = {10.1145/1772690.1772725},
booktitle = {Proceedings of the 19th International Conference on World Wide Web},
pages = {331–340},
numpages = {10},
location = {Raleigh, North Carolina, USA},
series = {WWW '10}
}

@inproceedings{kanhabua2011comparison,
author = {Kanhabua, Nattiya and N\o{}rv\r{a}g, Kjetil},
title = {A comparison of time-aware ranking methods},
year = {2011},
isbn = {9781450307574},
publisher = {Association for Computing Machinery},
address = {New York, NY, USA},
url = {https://doi.org/10.1145/2009916.2010147},
doi = {10.1145/2009916.2010147},
booktitle = {Proceedings of the 34th International ACM SIGIR Conference on Research and Development in Information Retrieval},
pages = {1257–1258},
numpages = {2},
location = {Beijing, China},
series = {SIGIR '11}
}

@inproceedings{styskin2011recency,
author = {Styskin, Andrey and Romanenko, Fedor and Vorobyev, Fedor and Serdyukov, Pavel},
title = {Recency ranking by diversification of result set},
year = {2011},
isbn = {9781450307178},
publisher = {Association for Computing Machinery},
address = {New York, NY, USA},
url = {https://doi.org/10.1145/2063576.2063862},
doi = {10.1145/2063576.2063862},
booktitle = {Proceedings of the 20th ACM International Conference on Information and Knowledge Management},
pages = {1949–1952},
numpages = {4},
location = {Glasgow, Scotland, UK},
series = {CIKM '11}
}

@inproceedings{alonso2011temporal,
  title={Temporal Information Retrieval: Challenges and Opportunities},
  author={Alonso, Omar and Str{\"o}tgen, Jannik and Baeza-Yates, Ricardo and Gertz, Michael},
  booktitle={Temporal Web Analytics Workshop TWAW 2011},
  url="https://ceur-ws.org/Vol-707/TWAW2011-paper1.pdf",
  pages={1},
  year={2011}
}

@inproceedings{kulkarni2011understanding,
author = {Kulkarni, Anagha and Teevan, Jaime and Svore, Krysta M. and Dumais, Susan T.},
title = {Understanding temporal query dynamics},
year = {2011},
isbn = {9781450304931},
publisher = {Association for Computing Machinery},
address = {New York, NY, USA},
url = {https://doi.org/10.1145/1935826.1935862},
doi = {10.1145/1935826.1935862},
booktitle = {Proceedings of the Fourth ACM International Conference on Web Search and Data Mining},
pages = {167–176},
numpages = {10},
location = {Hong Kong, China},
series = {WSDM '11}
}

@inproceedings{kanhabua2012learning,
author = {Kanhabua, Nattiya and Berberich, Klaus and N\o{}rv\r{a}g, Kjetil},
title = {Learning to select a time-aware retrieval model},
year = {2012},
isbn = {9781450314725},
publisher = {Association for Computing Machinery},
address = {New York, NY, USA},
url = {https://doi.org/10.1145/2348283.2348488},
doi = {10.1145/2348283.2348488},
booktitle = {Proceedings of the 35th International ACM SIGIR Conference on Research and Development in Information Retrieval},
pages = {1099–1100},
numpages = {2},
location = {Portland, Oregon, USA},
series = {SIGIR '12}
}

@article{wang2021improving,
author = {Wang, Jiexin and Jatowt, Adam and F\"{a}rber, Michael and Yoshikawa, Masatoshi},
title = {Improving question answering for event-focused questions in temporal collections of news articles},
year = {2021},
issue_date = {Feb 2021},
publisher = {Kluwer Academic Publishers},
address = {USA},
volume = {24},
number = {1},
issn = {1386-4564},
url = {https://doi.org/10.1007/s10791-020-09387-9},
doi = {10.1007/s10791-020-09387-9},
journal = {Inf. Retr.},
month = feb,
pages = {29–54},
numpages = {26}
}

@inproceedings{wang2020answering,
  title={Answering event-related questions over long-term news article archives},
  author={Wang, Jiexin and Jatowt, Adam and F{\"a}rber, Michael and Yoshikawa, Masatoshi},
  booktitle={Advances in Information Retrieval: 42nd European Conference on IR Research, ECIR 2020, Lisbon, Portugal, April 14--17, 2020, Proceedings, Part I 42},
  pages={774--789},
  year={2020},
  organization={Springer}
}

@inproceedings{yang-etal-2020-improving,
    title = "Improving Event Duration Prediction via Time-aware Pre-training",
    author = "Yang, Zonglin  and
      Du, Xinya  and
      Rush, Alexander  and
      Cardie, Claire",
    editor = "Cohn, Trevor  and
      He, Yulan  and
      Liu, Yang",
    booktitle = "Findings of the Association for Computational Linguistics: EMNLP 2020",
    month = nov,
    year = "2020",
    address = "Online",
    publisher = "Association for Computational Linguistics",
    url = "https://aclanthology.org/2020.findings-emnlp.302/",
    doi = "10.18653/v1/2020.findings-emnlp.302",
    pages = "3370--3378"
}

@inproceedings{zhang-choi-2021-situatedqa,
    title = "{S}ituated{QA}: Incorporating Extra-Linguistic Contexts into {QA}",
    author = "Zhang, Michael  and
      Choi, Eunsol",
    editor = "Moens, Marie-Francine  and
      Huang, Xuanjing  and
      Specia, Lucia  and
      Yih, Scott Wen-tau",
    booktitle = "Proceedings of the 2021 Conference on Empirical Methods in Natural Language Processing",
    month = nov,
    year = "2021",
    address = "Online and Punta Cana, Dominican Republic",
    publisher = "Association for Computational Linguistics",
    url = "https://aclanthology.org/2021.emnlp-main.586/",
    doi = "10.18653/v1/2021.emnlp-main.586",
    pages = "7371--7387"
}

@inproceedings{jin-etal-2021-forecastqa,
    title = "{F}orecast{QA}: A Question Answering Challenge for Event Forecasting with Temporal Text Data",
    author = "Jin, Woojeong  and
      Khanna, Rahul  and
      Kim, Suji  and
      Lee, Dong-Ho  and
      Morstatter, Fred  and
      Galstyan, Aram  and
      Ren, Xiang",
    editor = "Zong, Chengqing  and
      Xia, Fei  and
      Li, Wenjie  and
      Navigli, Roberto",
    booktitle = "Proceedings of the 59th Annual Meeting of the Association for Computational Linguistics and the 11th International Joint Conference on Natural Language Processing (Volume 1: Long Papers)",
    month = aug,
    year = "2021",
    address = "Online",
    publisher = "Association for Computational Linguistics",
    url = "https://aclanthology.org/2021.acl-long.357/",
    doi = "10.18653/v1/2021.acl-long.357",
    pages = "4636--4650",
}

@inproceedings{lin-etal-2021-conditional,
    title = "Conditional Generation of Temporally-ordered Event Sequences",
    author = "Lin, Shih-Ting  and
      Chambers, Nathanael  and
      Durrett, Greg",
    editor = "Zong, Chengqing  and
      Xia, Fei  and
      Li, Wenjie  and
      Navigli, Roberto",
    booktitle = "Proceedings of the 59th Annual Meeting of the Association for Computational Linguistics and the 11th International Joint Conference on Natural Language Processing (Volume 1: Long Papers)",
    month = aug,
    year = "2021",
    address = "Online",
    publisher = "Association for Computational Linguistics",
    url = "https://aclanthology.org/2021.acl-long.555/",
    doi = "10.18653/v1/2021.acl-long.555",
    pages = "7142--7157",
}

@inproceedings{cao-wang-2022-time,
    title = "Time-aware Prompting for Text Generation",
    author = "Cao, Shuyang  and
      Wang, Lu",
    editor = "Goldberg, Yoav  and
      Kozareva, Zornitsa  and
      Zhang, Yue",
    booktitle = "Findings of the Association for Computational Linguistics: EMNLP 2022",
    month = dec,
    year = "2022",
    address = "Abu Dhabi, United Arab Emirates",
    publisher = "Association for Computational Linguistics",
    url = "https://aclanthology.org/2022.findings-emnlp.535/",
    doi = "10.18653/v1/2022.findings-emnlp.535",
    pages = "7231--7246",
}

@inproceedings{son-oh-2023-time,
    title = "Time-Aware Representation Learning for Time-Sensitive Question Answering",
    author = "Son, Jungbin  and
      Oh, Alice",
    editor = "Bouamor, Houda  and
      Pino, Juan  and
      Bali, Kalika",
    booktitle = "Findings of the Association for Computational Linguistics: EMNLP 2023",
    month = dec,
    year = "2023",
    address = "Singapore",
    publisher = "Association for Computational Linguistics",
    url = "https://aclanthology.org/2023.findings-emnlp.6/",
    doi = "10.18653/v1/2023.findings-emnlp.6",
    pages = "70--77",
}

@inproceedings{xiong-etal-2024-large,
    title = "Large Language Models Can Learn Temporal Reasoning",
    author = "Xiong, Siheng  and
      Payani, Ali  and
      Kompella, Ramana  and
      Fekri, Faramarz",
    editor = "Ku, Lun-Wei  and
      Martins, Andre  and
      Srikumar, Vivek",
    booktitle = "Proceedings of the 62nd Annual Meeting of the Association for Computational Linguistics (Volume 1: Long Papers)",
    month = aug,
    year = "2024",
    address = "Bangkok, Thailand",
    publisher = "Association for Computational Linguistics",
    url = "https://aclanthology.org/2024.acl-long.563/",
    doi = "10.18653/v1/2024.acl-long.563",
    pages = "10452--10470",
}

@inproceedings{su-etal-2024-living,
    title = "Living in the Moment: Can Large Language Models Grasp Co-Temporal Reasoning?",
    author = "Su, Zhaochen  and
      Li, Juntao  and
      Zhang, Jun  and
      Zhu, Tong  and
      Qu, Xiaoye  and
      Zhou, Pan  and
      Bowen, Yan  and
      Cheng, Yu  and
      Zhang, Min",
    editor = "Ku, Lun-Wei  and
      Martins, Andre  and
      Srikumar, Vivek",
    booktitle = "Proceedings of the 62nd Annual Meeting of the Association for Computational Linguistics (Volume 1: Long Papers)",
    month = aug,
    year = "2024",
    address = "Bangkok, Thailand",
    publisher = "Association for Computational Linguistics",
    url = "https://aclanthology.org/2024.acl-long.703/",
    doi = "10.18653/v1/2024.acl-long.703",
    pages = "13014--13033",

}

@inproceedings{barik-etal-2024-time,
    title = "Time Matters: An End-to-End Solution for Temporal Claim Verification",
    author = "Barik, Anab Maulana  and
      Hsu, Wynne  and
      Lee, Mong-Li",
    editor = "Dernoncourt, Franck  and
      Preo{\c{t}}iuc-Pietro, Daniel  and
      Shimorina, Anastasia",
    booktitle = "Proceedings of the 2024 Conference on Empirical Methods in Natural Language Processing: Industry Track",
    month = nov,
    year = "2024",
    address = "Miami, Florida, US",
    publisher = "Association for Computational Linguistics",
    url = "https://aclanthology.org/2024.emnlp-industry.48/",
    doi = "10.18653/v1/2024.emnlp-industry.48",
    pages = "657--664",
}

@inproceedings{qian-etal-2024-timer4,
    title = "{T}ime{R}$^4$ : Time-aware Retrieval-Augmented Large Language Models for Temporal Knowledge Graph Question Answering",
    author = "Qian, Xinying  and
      Zhang, Ying  and
      Zhao, Yu  and
      Zhou, Baohang  and
      Sui, Xuhui  and
      Zhang, Li  and
      Song, Kehui",
    editor = "Al-Onaizan, Yaser  and
      Bansal, Mohit  and
      Chen, Yun-Nung",
    booktitle = "Proceedings of the 2024 Conference on Empirical Methods in Natural Language Processing",
    month = nov,
    year = "2024",
    address = "Miami, Florida, USA",
    publisher = "Association for Computational Linguistics",
    url = "https://aclanthology.org/2024.emnlp-main.394/",
    doi = "10.18653/v1/2024.emnlp-main.394",
    pages = "6942--6952",
}

@inproceedings{meem-etal-2024-pat,
    title = "{PAT}-Questions: A Self-Updating Benchmark for Present-Anchored Temporal Question-Answering",
    author = "Meem, Jannat  and
      Rashid, Muhammad  and
      Dong, Yue  and
      Hristidis, Vagelis",
    editor = "Ku, Lun-Wei  and
      Martins, Andre  and
      Srikumar, Vivek",
    booktitle = "Findings of the Association for Computational Linguistics: ACL 2024",
    month = aug,
    year = "2024",
    address = "Bangkok, Thailand",
    publisher = "Association for Computational Linguistics",
    url = "https://aclanthology.org/2024.findings-acl.777/",
    doi = "10.18653/v1/2024.findings-acl.777",
    pages = "13129--13148",
}

@inproceedings{qin-etal-2021-timedial,
    title = "{TIMEDIAL}: Temporal Commonsense Reasoning in Dialog",
    author = "Qin, Lianhui  and
      Gupta, Aditya  and
      Upadhyay, Shyam  and
      He, Luheng  and
      Choi, Yejin  and
      Faruqui, Manaal",
    editor = "Zong, Chengqing  and
      Xia, Fei  and
      Li, Wenjie  and
      Navigli, Roberto",
    booktitle = "Proceedings of the 59th Annual Meeting of the Association for Computational Linguistics and the 11th International Joint Conference on Natural Language Processing (Volume 1: Long Papers)",
    month = aug,
    year = "2021",
    address = "Online",
    publisher = "Association for Computational Linguistics",
    url = "https://aclanthology.org/2021.acl-long.549/",
    doi = "10.18653/v1/2021.acl-long.549",
    pages = "7066--7076",
}

@inproceedings{mathur-etal-2021-timers,
    title = "{TIMERS}: Document-level Temporal Relation Extraction",
    author = "Mathur, Puneet  and
      Jain, Rajiv  and
      Dernoncourt, Franck  and
      Morariu, Vlad  and
      Tran, Quan Hung  and
      Manocha, Dinesh",
    editor = "Zong, Chengqing  and
      Xia, Fei  and
      Li, Wenjie  and
      Navigli, Roberto",
    booktitle = "Proceedings of the 59th Annual Meeting of the Association for Computational Linguistics and the 11th International Joint Conference on Natural Language Processing (Volume 2: Short Papers)",
    month = aug,
    year = "2021",
    address = "Online",
    publisher = "Association for Computational Linguistics",
    url = "https://aclanthology.org/2021.acl-short.67/",
    doi = "10.18653/v1/2021.acl-short.67",
    pages = "524--533",
}

@article{rizzo2022ranking,
author = {Rizzo, Stefano Giovanni and Brucato, Matteo and Montesi, Danilo},
title = {Ranking Models for the Temporal Dimension of Text},
year = {2022},
issue_date = {April 2023},
publisher = {Association for Computing Machinery},
address = {New York, NY, USA},
volume = {41},
number = {2},
issn = {1046-8188},
url = {https://doi.org/10.1145/3565481},
doi = {10.1145/3565481},
journal = {ACM Trans. Inf. Syst.},
month = dec,
articleno = {49},
numpages = {34}
}

@inproceedings{ren-etal-2023-time,
    title = "Time-Aware Language Modeling for Historical Text Dating",
    author = "Ren, Han  and
      Wang, Hai  and
      Zhao, Yajie  and
      Ren, Yafeng",
    editor = "Bouamor, Houda  and
      Pino, Juan  and
      Bali, Kalika",
    booktitle = "Findings of the Association for Computational Linguistics: EMNLP 2023",
    month = dec,
    year = "2023",
    address = "Singapore",
    publisher = "Association for Computational Linguistics",
    url = "https://aclanthology.org/2023.findings-emnlp.911/",
    doi = "10.18653/v1/2023.findings-emnlp.911",
    pages = "13646--13656",
}

@inproceedings{niu-etal-2024-contempo,
    title = "{C}on{T}empo: A Unified Temporally Contrastive Framework for Temporal Relation Extraction",
    author = "Niu, Jingcheng  and
      Liao, Saifei  and
      Ng, Victoria  and
      De Montigny, Simon  and
      Penn, Gerald",
    editor = "Ku, Lun-Wei  and
      Martins, Andre  and
      Srikumar, Vivek",
    booktitle = "Findings of the Association for Computational Linguistics: ACL 2024",
    month = aug,
    year = "2024",
    address = "Bangkok, Thailand",
    publisher = "Association for Computational Linguistics",
    url = "https://aclanthology.org/2024.findings-acl.89/",
    doi = "10.18653/v1/2024.findings-acl.89",
    pages = "1521--1533",
}

@inproceedings{vu-etal-2024-freshllms,
    title = "{F}resh{LLM}s: Refreshing Large Language Models with Search Engine Augmentation",
    author = "Vu, Tu  and
      Iyyer, Mohit  and
      Wang, Xuezhi  and
      Constant, Noah  and
      Wei, Jerry  and
      Wei, Jason  and
      Tar, Chris  and
      Sung, Yun-Hsuan  and
      Zhou, Denny  and
      Le, Quoc  and
      Luong, Thang",
    editor = "Ku, Lun-Wei  and
      Martins, Andre  and
      Srikumar, Vivek",
    booktitle = "Findings of the Association for Computational Linguistics: ACL 2024",
    month = aug,
    year = "2024",
    address = "Bangkok, Thailand",
    publisher = "Association for Computational Linguistics",
    url = "https://aclanthology.org/2024.findings-acl.813/",
    doi = "10.18653/v1/2024.findings-acl.813",
    pages = "13697--13720",
}

@inproceedings{trischler-etal-2017-newsqa,
    title = "{N}ews{QA}: A Machine Comprehension Dataset",
    author = "Trischler, Adam  and
      Wang, Tong  and
      Yuan, Xingdi  and
      Harris, Justin  and
      Sordoni, Alessandro  and
      Bachman, Philip  and
      Suleman, Kaheer",
    editor = "Blunsom, Phil  and
      Bordes, Antoine  and
      Cho, Kyunghyun  and
      Cohen, Shay  and
      Dyer, Chris  and
      Grefenstette, Edward  and
      Hermann, Karl Moritz  and
      Rimell, Laura  and
      Weston, Jason  and
      Yih, Scott",
    booktitle = "Proceedings of the 2nd Workshop on Representation Learning for {NLP}",
    month = aug,
    year = "2017",
    address = "Vancouver, Canada",
    publisher = "Association for Computational Linguistics",
    url = "https://aclanthology.org/W17-2623/",
    doi = "10.18653/v1/W17-2623",
    pages = "191--200"
}

@inproceedings{naik-etal-2019-tddiscourse,
    title = "{TDD}iscourse: A Dataset for Discourse-Level Temporal Ordering of Events",
    author = "Naik, Aakanksha  and
      Breitfeller, Luke  and
      Rose, Carolyn",
    editor = "Nakamura, Satoshi  and
      Gasic, Milica  and
      Zukerman, Ingrid  and
      Skantze, Gabriel  and
      Nakano, Mikio  and
      Papangelis, Alexandros  and
      Ultes, Stefan  and
      Yoshino, Koichiro",
    booktitle = "Proceedings of the 20th Annual SIGdial Meeting on Discourse and Dialogue",
    month = sep,
    year = "2019",
    address = "Stockholm, Sweden",
    publisher = "Association for Computational Linguistics",
    url = "https://aclanthology.org/W19-5929/",
    doi = "10.18653/v1/W19-5929",
    pages = "239--249",
}

@Article{ong2023tkgqa,
AUTHOR = {Ong, Ryan and Sun, Jiahao and Șerban, Ovidiu and Guo, Yi-Ke},
TITLE = {TKGQA Dataset: Using Question Answering to Guide and Validate the Evolution of Temporal Knowledge Graph},
JOURNAL = {Data},
VOLUME = {8},
YEAR = {2023},
NUMBER = {3},
ARTICLE-NUMBER = {61},
URL = {https://www.mdpi.com/2306-5729/8/3/61},
ISSN = {2306-5729},
DOI = {10.3390/data8030061}
}

@inproceedings{strotgen-gertz-2010-heideltime,
    title = "{H}eidel{T}ime: High Quality Rule-Based Extraction and Normalization of Temporal Expressions",
    author = {Str{\"o}tgen, Jannik  and
      Gertz, Michael},
    editor = "Erk, Katrin  and
      Strapparava, Carlo",
    booktitle = "Proceedings of the 5th International Workshop on Semantic Evaluation",
    month = jul,
    year = "2010",
    address = "Uppsala, Sweden",
    publisher = "Association for Computational Linguistics",
    url = "https://aclanthology.org/S10-1071/",
    pages = "321--324"
}

@inproceedings{chang-manning-2012-sutime,
    title = "{SUT}ime: A library for recognizing and normalizing time expressions",
    author = "Chang, Angel X.  and
      Manning, Christopher",
    editor = "Calzolari, Nicoletta  and
      Choukri, Khalid  and
      Declerck, Thierry  and
      Do{\u{g}}an, Mehmet U{\u{g}}ur  and
      Maegaard, Bente  and
      Mariani, Joseph  and
      Moreno, Asuncion  and
      Odijk, Jan  and
      Piperidis, Stelios",
    booktitle = "Proceedings of the Eighth International Conference on Language Resources and Evaluation ({LREC}`12)",
    month = may,
    year = "2012",
    address = "Istanbul, Turkey",
    publisher = "European Language Resources Association (ELRA)",
    url = "https://aclanthology.org/L12-1122/",
    pages = "3735--3740",
}

@inproceedings{radinsky2013mining,
author = {Radinsky, Kira and Horvitz, Eric},
title = {Mining the web to predict future events},
year = {2013},
isbn = {9781450318693},
publisher = {Association for Computing Machinery},
address = {New York, NY, USA},
url = {https://doi.org/10.1145/2433396.2433431},
doi = {10.1145/2433396.2433431},
booktitle = {Proceedings of the Sixth ACM International Conference on Web Search and Data Mining},
pages = {255–264},
numpages = {10},
location = {Rome, Italy},
series = {WSDM '13}
}

@article{sandhaus2008new,
  title={The new york times annotated corpus},
  author={Sandhaus, Evan},
  journal={Linguistic Data Consortium, Philadelphia},
  volume={6},
  url = {https://doi.org/ 11272.1/AB2/GZC6PL},
  number={12},
  pages={e26752},
  year={2008}
}

@inproceedings{pustejovsky2003timebank,
  title={The timebank corpus},
  author={Pustejovsky, James and Hanks, Patrick and Sauri, Roser and See, Andrew and Gaizauskas, Robert and Setzer, Andrea and Radev, Dragomir and Sundheim, Beth and Day, David and Ferro, Lisa and others},
  booktitle={Corpus linguistics},
  volume={2003},
  pages={40},
  year={2003},
  organization={Lancaster, UK}
}

@inproceedings{mazur-dale-2010-wikiwars,
    title = "{W}iki{W}ars: A New Corpus for Research on Temporal Expressions",
    author = "Mazur, Pawel  and
      Dale, Robert",
    editor = "Li, Hang  and
      M{\`a}rquez, Llu{\'i}s",
    booktitle = "Proceedings of the 2010 Conference on Empirical Methods in Natural Language Processing",
    month = oct,
    year = "2010",
    address = "Cambridge, MA",
    publisher = "Association for Computational Linguistics",
    url = "https://aclanthology.org/D10-1089/",
    pages = "913--922"
}

@article{kobayashi2000information,
author = {Kobayashi, Mei and Takeda, Koichi},
title = {Information retrieval on the web},
year = {2000},
issue_date = {June 2000},
publisher = {Association for Computing Machinery},
address = {New York, NY, USA},
volume = {32},
number = {2},
issn = {0360-0300},
url = {https://doi.org/10.1145/358923.358934},
doi = {10.1145/358923.358934},
journal = {ACM Comput. Surv.},
month = jun,
pages = {144–173},
numpages = {30}
}

@article{10.1561/1500000043,
author = {Kanhabua, Nattiya and Blanco, Roi and N\o{}rv\r{a}g, Kjetil},
title = {Temporal Information Retrieval},
year = {2015},
issue_date = {Jul 2015},
publisher = {Now Publishers Inc.},
address = {Hanover, MA, USA},
volume = {9},
number = {2},
issn = {1554-0669},
url = {https://doi.org/10.1561/1500000043},
doi = {10.1561/1500000043},
journal = {Found. Trends Inf. Retr.},
month = jul,
pages = {91–208},
numpages = {121}
}

@inproceedings{ogorman-etal-2016-richer,
    title = "Richer Event Description: Integrating event coreference with temporal, causal and bridging annotation",
    author = "O{'}Gorman, Tim  and
      Wright-Bettner, Kristin  and
      Palmer, Martha",
    editor = "Caselli, Tommaso  and
      Miller, Ben  and
      van Erp, Marieke  and
      Vossen, Piek  and
      Caswell, David",
    booktitle = "Proceedings of the 2nd Workshop on Computing News Storylines ({CNS} 2016)",
    month = nov,
    year = "2016",
    address = "Austin, Texas",
    publisher = "Association for Computational Linguistics",
    url = "https://aclanthology.org/W16-5706/",
    doi = "10.18653/v1/W16-5706",
    pages = "47--56"
}

@inproceedings{miller-1992-wordnet,
    title = "{W}ord{N}et: A Lexical Database for {E}nglish",
    author = "Miller, George A.",
    booktitle = "Speech and Natural Language: Proceedings of a Workshop Held at Harriman, New York, {F}ebruary 23-26, 1992",
    year = "1992",
    url = "https://aclanthology.org/H92-1116/"
}

@inproceedings{anand2011temporal,
author = {Anand, Avishek and Bedathur, Srikanta and Berberich, Klaus and Schenkel, Ralf},
title = {Temporal index sharding for space-time efficiency in archive search},
year = {2011},
isbn = {9781450307574},
publisher = {Association for Computing Machinery},
address = {New York, NY, USA},
url = {https://doi.org/10.1145/2009916.2009991},
doi = {10.1145/2009916.2009991},
booktitle = {Proceedings of the 34th International ACM SIGIR Conference on Research and Development in Information Retrieval},
pages = {545–554},
numpages = {10},
location = {Beijing, China},
series = {SIGIR '11}
}

@inproceedings{abdallah2025tempretriever,
author = {Abdallah, Abdelrahman and Piryani, Bhawna and Wallat, Jonas and Anand, Avishek and Jatowt, Adam},
title = {TempRetriever: Fusion-based Temporal Dense Passage Retrieval for Time-Sensitive Questions},
year = {2026},
isbn = {9798400722929},
publisher = {Association for Computing Machinery},
address = {New York, NY, USA},
url = {https://doi.org/10.1145/3773966.3777938},
doi = {10.1145/3773966.3777938},
booktitle = {Proceedings of the Nineteenth ACM International Conference on Web Search and Data Mining},
pages = {5–15},
numpages = {11},
location = {USA},
series = {WSDM '26}
}

@inproceedings{joho2013survey,
author = {Joho, Hideo and Jatowt, Adam and Roi, Blanco},
title = {A survey of temporal web search experience},
year = {2013},
isbn = {9781450320382},
publisher = {Association for Computing Machinery},
address = {New York, NY, USA},
url = {https://doi.org/10.1145/2487788.2488126},
doi = {10.1145/2487788.2488126},
booktitle = {Proceedings of the 22nd International Conference on World Wide Web},
pages = {1101–1108},
numpages = {8},
location = {Rio de Janeiro, Brazil},
series = {WWW '13 Companion}
}

@inproceedings{salaheldeen2013carbon,
author = {SalahEldeen, Hany M. and Nelson, Michael L.},
title = {Carbon dating the web: estimating the age of web resources},
year = {2013},
isbn = {9781450320382},
publisher = {Association for Computing Machinery},
address = {New York, NY, USA},
url = {https://doi.org/10.1145/2487788.2488121},
doi = {10.1145/2487788.2488121},
booktitle = {Proceedings of the 22nd International Conference on World Wide Web},
pages = {1075–1082},
numpages = {8},
location = {Rio de Janeiro, Brazil},
series = {WWW '13 Companion}
}

@inproceedings{niculae-etal-2014-temporal,
    title = "Temporal Text Ranking and Automatic Dating of Texts",
    author = "Niculae, Vlad  and
      Zampieri, Marcos  and
      Dinu, Liviu  and
      Ciobanu, Alina Maria",
    editor = "Wintner, Shuly  and
      Riezler, Stefan  and
      Goldwater, Sharon",
    booktitle = "Proceedings of the 14th Conference of the {E}uropean Chapter of the Association for Computational Linguistics, volume 2: Short Papers",
    month = apr,
    year = "2014",
    address = "Gothenburg, Sweden",
    publisher = "Association for Computational Linguistics",
    url = "https://aclanthology.org/E14-4004/",
    doi = "10.3115/v1/E14-4004",
    pages = "17--21"
}

@inproceedings{kuzey2016temponym,
author = {Kuzey, Erdal and Str\"{o}tgen, Jannik and Setty, Vinay and Weikum, Gerhard},
title = {Temponym Tagging: Temporal Scopes for Textual Phrases},
year = {2016},
isbn = {9781450341448},
publisher = {International World Wide Web Conferences Steering Committee},
address = {Republic and Canton of Geneva, CHE},
url = {https://doi.org/10.1145/2872518.2889289},
doi = {10.1145/2872518.2889289},
booktitle = {Proceedings of the 25th International Conference Companion on World Wide Web},
pages = {841–842},
numpages = {2},
location = {Montr\'{e}al, Qu\'{e}bec, Canada},
series = {WWW '16 Companion}
}

@inproceedings{ning-etal-2018-cogcomptime,
    title = "{C}og{C}omp{T}ime: A Tool for Understanding Time in Natural Language",
    author = "Ning, Qiang  and
      Zhou, Ben  and
      Feng, Zhili  and
      Peng, Haoruo  and
      Roth, Dan",
    editor = "Blanco, Eduardo  and
      Lu, Wei",
    booktitle = "Proceedings of the 2018 Conference on Empirical Methods in Natural Language Processing: System Demonstrations",
    month = nov,
    year = "2018",
    address = "Brussels, Belgium",
    publisher = "Association for Computational Linguistics",
    url = "https://aclanthology.org/D18-2013/",
    doi = "10.18653/v1/D18-2013",
    pages = "72--77"
}

@inproceedings{wallat2024temporal,
author = {Wallat, Jonas and Jatowt, Adam and Anand, Avishek},
title = {Temporal Blind Spots in Large Language Models},
year = {2024},
isbn = {9798400703713},
publisher = {Association for Computing Machinery},
address = {New York, NY, USA},
url = {https://doi.org/10.1145/3616855.3635818},
doi = {10.1145/3616855.3635818},
booktitle = {Proceedings of the 17th ACM International Conference on Web Search and Data Mining},
pages = {683–692},
numpages = {10},
location = {Merida, Mexico},
series = {WSDM '24}
}

@article{mutschlechner2025analyzing,
  title={Analyzing the role of context in forecasting with large language models},
  author={Mutschlechner, Gerrit and Jatowt, Adam},
  journal={arXiv preprint arXiv:2501.06496},
  year={2025}
}

@article{mavi2024multi,
  title={Multi-hop question answering},
  author={Mavi, Vaibhav and Jangra, Anubhav and Jatowt, Adam and others},
  journal={Foundations and Trends{\textregistered} in Information Retrieval},
  volume={17},
  number={5},
  pages={457--586},
  year={2024},
  publisher={Now Publishers, Inc.}
}

@inproceedings{wallat2025study,
    title = "A Study into Investigating Temporal Robustness of {LLM}s",
    author = "Wallat, Jonas  and
      Abdallah, Abdelrahman  and
      Jatowt, Adam  and
      Anand, Avishek",
    editor = "Che, Wanxiang  and
      Nabende, Joyce  and
      Shutova, Ekaterina  and
      Pilehvar, Mohammad Taher",
    booktitle = "Findings of the Association for Computational Linguistics: ACL 2025",
    month = jul,
    year = "2025",
    address = "Vienna, Austria",
    publisher = "Association for Computational Linguistics",
    url = "https://aclanthology.org/2025.findings-acl.810/",
    doi = "10.18653/v1/2025.findings-acl.810",
    pages = "15685--15705",
    ISBN = "979-8-89176-256-5"
}

@article{abdallah2026tempo,
  title={TEMPO: A Realistic Multi-Domain Benchmark for Temporal Reasoning-Intensive Retrieval},
  author={Abdallah, Abdelrahman and Ali, Mohammed and Abdul-Mageed, Muhammad and Jatowt, Adam},
  journal={arXiv preprint arXiv:2601.09523},
  year={2026}
}

@InProceedings{liska22a-streaminqa,
  title = 	 {{S}treaming{QA}: A Benchmark for Adaptation to New Knowledge over Time in Question Answering Models},
  author =       {Liska, Adam and Kocisky, Tomas and Gribovskaya, Elena and Terzi, Tayfun and Sezener, Eren and Agrawal, Devang and De Masson D'Autume, Cyprien and Scholtes, Tim and Zaheer, Manzil and Young, Susannah and Gilsenan-Mcmahon, Ellen and Austin, Sophia and Blunsom, Phil and Lazaridou, Angeliki},
  booktitle = 	 {Proceedings of the 39th International Conference on Machine Learning},
  pages = 	 {13604--13622},
  year = 	 {2022},
  editor = 	 {Chaudhuri, Kamalika and Jegelka, Stefanie and Song, Le and Szepesvari, Csaba and Niu, Gang and Sabato, Sivan},
  volume = 	 {162},
  series = 	 {Proceedings of Machine Learning Research},
  month = 	 {17--23 Jul},
  publisher =    {PMLR},
  url = 	 {https://proceedings.mlr.press/v162/liska22a.html},
}

@inproceedings{gupta-etal-2023-temptabqa,
    title = "{T}emp{T}ab{QA}: Temporal Question Answering for Semi-Structured Tables",
    author = "Gupta, Vivek  and
      Kandoi, Pranshu  and
      Vora, Mahek  and
      Zhang, Shuo  and
      He, Yujie  and
      Reinanda, Ridho  and
      Srikumar, Vivek",
    editor = "Bouamor, Houda  and
      Pino, Juan  and
      Bali, Kalika",
    booktitle = "Proceedings of the 2023 Conference on Empirical Methods in Natural Language Processing",
    month = dec,
    year = "2023",
    address = "Singapore",
    publisher = "Association for Computational Linguistics",
    url = "https://aclanthology.org/2023.emnlp-main.149/",
    doi = "10.18653/v1/2023.emnlp-main.149",
    pages = "2431--2453",
}

@inproceedings{NEURIPS2024_58f52a01_newswire,
 author = {Silcock, Emily and Arora, Abhishek and D\textquotesingle Amico-Wong, Luca and Dell, Melissa},
 booktitle = {Advances in Neural Information Processing Systems},
 editor = {A. Globerson and L. Mackey and D. Belgrave and A. Fan and U. Paquet and J. Tomczak and C. Zhang},
 pages = {49768--49779},
 publisher = {Curran Associates, Inc.},
 title = {Newswire: A Large-Scale Structured Database of a Century of Historical News},
 url = {https://proceedings.neurips.cc/paper_files/paper/2024/file/58f52a01a516609336da78ff17e6f81f-Paper-Datasets_and_Benchmarks_Track.pdf},
 volume = {37},
 year = {2024}
}

@inproceedings{hamilton-etal-2016-diachronic,
    title = "Diachronic Word Embeddings Reveal Statistical Laws of Semantic Change",
    author = "Hamilton, William L.  and
      Leskovec, Jure  and
      Jurafsky, Dan",
    editor = "Erk, Katrin  and
      Smith, Noah A.",
    booktitle = "Proceedings of the 54th Annual Meeting of the Association for Computational Linguistics (Volume 1: Long Papers)",
    month = aug,
    year = "2016",
    address = "Berlin, Germany",
    publisher = "Association for Computational Linguistics",
    url = "https://aclanthology.org/P16-1141/",
    doi = "10.18653/v1/P16-1141",
    pages = "1489--1501"
}

@inproceedings{NIPS2015__ccn/dailynews,
 author = {Hermann, Karl Moritz and Kocisky, Tomas and Grefenstette, Edward and Espeholt, Lasse and Kay, Will and Suleyman, Mustafa and Blunsom, Phil},
 booktitle = {Advances in Neural Information Processing Systems},
 editor = {C. Cortes and N. Lawrence and D. Lee and M. Sugiyama and R. Garnett},
 pages = {},
 publisher = {Curran Associates, Inc.},
 title = {Teaching Machines to Read and Comprehend},
 url = {https://proceedings.neurips.cc/paper_files/paper/2015/file/afdec7005cc9f14302cd0474fd0f3c96-Paper.pdf},
 volume = {28},
 year = {2015}
}

@inproceedings{jia2021complex,
author = {Jia, Zhen and Pramanik, Soumajit and Saha Roy, Rishiraj and Weikum, Gerhard},
title = {Complex Temporal Question Answering on Knowledge Graphs},
year = {2021},
isbn = {9781450384469},
publisher = {Association for Computing Machinery},
address = {New York, NY, USA},
url = {https://doi.org/10.1145/3459637.3482416},
doi = {10.1145/3459637.3482416},
booktitle = {Proceedings of the 30th ACM International Conference on Information \& Knowledge Management},
pages = {792–802},
numpages = {11},
location = {Virtual Event, Queensland, Australia},
series = {CIKM '21}
}

@inproceedings{saxena-etal-2021-question,
    title = "Question Answering Over Temporal Knowledge Graphs",
    author = "Saxena, Apoorv  and
      Chakrabarti, Soumen  and
      Talukdar, Partha",
    editor = "Zong, Chengqing  and
      Xia, Fei  and
      Li, Wenjie  and
      Navigli, Roberto",
    booktitle = "Proceedings of the 59th Annual Meeting of the Association for Computational Linguistics and the 11th International Joint Conference on Natural Language Processing (Volume 1: Long Papers)",
    month = aug,
    year = "2021",
    address = "Online",
    publisher = "Association for Computational Linguistics",
    url = "https://aclanthology.org/2021.acl-long.520/",
    doi = "10.18653/v1/2021.acl-long.520",
    pages = "6663--6676",
}

@inproceedings{chen-etal-2023-multi,
    title = "Multi-granularity Temporal Question Answering over Knowledge Graphs",
    author = "Chen, Ziyang  and
      Liao, Jinzhi  and
      Zhao, Xiang",
    editor = "Rogers, Anna  and
      Boyd-Graber, Jordan  and
      Okazaki, Naoaki",
    booktitle = "Proceedings of the 61st Annual Meeting of the Association for Computational Linguistics (Volume 1: Long Papers)",
    month = jul,
    year = "2023",
    address = "Toronto, Canada",
    publisher = "Association for Computational Linguistics",
    url = "https://aclanthology.org/2023.acl-long.637/",
    doi = "10.18653/v1/2023.acl-long.637",
    pages = "11378--11392",
}

@article{Xiong_Yang_Payani_Kerce_Fekri_2024, 
    title={TEILP: Time Prediction over Knowledge Graphs via Logical Reasoning}, 
    volume={38}, 
    url={https://ojs.aaai.org/index.php/AAAI/article/view/29544},
    DOI={10.1609/aaai.v38i14.29544},
    number={14}, 
    journal={Proceedings of the AAAI Conference on Artificial Intelligence}, 
    author={Xiong, Siheng and Yang, Yuan and Payani, Ali and Kerce, James C and Fekri, Faramarz}, 
    year={2024}, 
    month={Mar.}, 
    pages ={16112-16119} ,
}

@inproceedings{gutehrle-etal-2022-archive,
    title = "Archive {T}ime{L}ine Summarization ({ATLS}): Conceptual Framework for Timeline Generation over Historical Document Collections",
    author = "Gutehrl{\'e}, Nicolas  and
      Doucet, Antoine  and
      Jatowt, Adam",
    editor = "Degaetano, Stefania  and
      Kazantseva, Anna  and
      Reiter, Nils  and
      Szpakowicz, Stan",
    booktitle = "Proceedings of the 6th Joint SIGHUM Workshop on Computational Linguistics for Cultural Heritage, Social Sciences, Humanities and Literature",
    month = oct,
    year = "2022",
    address = "Gyeongju, Republic of Korea",
    publisher = "International Conference on Computational Linguistics",
    url = "https://aclanthology.org/2022.latechclfl-1.3/",
    pages = "13--23",
}

@inproceedings{tan-etal-2024-towards,
    title = "Towards Robust Temporal Reasoning of Large Language Models via a Multi-Hop {QA} Dataset and Pseudo-Instruction Tuning",
    author = "Tan, Qingyu  and
      Ng, Hwee Tou  and
      Bing, Lidong",
    editor = "Ku, Lun-Wei  and
      Martins, Andre  and
      Srikumar, Vivek",
    booktitle = "Findings of the Association for Computational Linguistics: ACL 2024",
    month = aug,
    year = "2024",
    address = "Bangkok, Thailand",
    publisher = "Association for Computational Linguistics",
    url = "https://aclanthology.org/2024.findings-acl.374/",
    doi = "10.18653/v1/2024.findings-acl.374",
    pages = "6272--6286",
}

@inproceedings{jain-etal-2023-language-models,
    title = "Do Language Models Have a Common Sense regarding Time? Revisiting Temporal Commonsense Reasoning in the Era of Large Language Models",
    author = "Jain, Raghav  and
      Sojitra, Daivik  and
      Acharya, Arkadeep  and
      Saha, Sriparna  and
      Jatowt, Adam  and
      Dandapat, Sandipan",
    editor = "Bouamor, Houda  and
      Pino, Juan  and
      Bali, Kalika",
    booktitle = "Proceedings of the 2023 Conference on Empirical Methods in Natural Language Processing",
    month = dec,
    year = "2023",
    address = "Singapore",
    publisher = "Association for Computational Linguistics",
    url = "https://aclanthology.org/2023.emnlp-main.418/",
    doi = "10.18653/v1/2023.emnlp-main.418",
    pages = "6750--6774",
}

@inproceedings{sojitra2024timeline,
author = {Sojitra, Daivik and Jain, Raghav and Saha, Sriparna and Jatowt, Adam and Gupta, Manish},
title = {Timeline Summarization in the Era of LLMs},
year = {2024},
isbn = {9798400704314},
publisher = {Association for Computing Machinery},
address = {New York, NY, USA},
url = {https://doi.org/10.1145/3626772.3657899},
doi = {10.1145/3626772.3657899},
booktitle = {Proceedings of the 47th International ACM SIGIR Conference on Research and Development in Information Retrieval},
pages = {2657–2661},
numpages = {5},
location = {Washington DC, USA},
series = {SIGIR '24}
}

@inproceedings{kuzey2016time,
author = {Kuzey, Erdal and Setty, Vinay and Str\"{o}tgen, Jannik and Weikum, Gerhard},
title = {As Time Goes By: Comprehensive Tagging of Textual Phrases with Temporal Scopes},
year = {2016},
isbn = {9781450341431},
publisher = {International World Wide Web Conferences Steering Committee},
address = {Republic and Canton of Geneva, CHE},
url = {https://doi.org/10.1145/2872427.2883055},
doi = {10.1145/2872427.2883055},
booktitle = {Proceedings of the 25th International Conference on World Wide Web},
pages = {915–925},
numpages = {11},
location = {Montr\'{e}al, Qu\'{e}bec, Canada},
series = {WWW '16}
}

@inproceedings{bedi-etal-2017-event,
    title = "Event Timeline Generation from History Textbooks",
    author = "Bedi, Harsimran  and
      Patil, Sangameshwar  and
      Hingmire, Swapnil  and
      Palshikar, Girish",
    editor = "Tseng, Yuen-Hsien  and
      Chen, Hsin-Hsi  and
      Lee, Lung-Hao  and
      Yu, Liang-Chih",
    booktitle = "Proceedings of the 4th Workshop on Natural Language Processing Techniques for Educational Applications ({NLPTEA} 2017)",
    month = dec,
    year = "2017",
    address = "Taipei, Taiwan",
    publisher = "Asian Federation of Natural Language Processing",
    url = "https://aclanthology.org/W17-5912/",
    pages = "69--77",
}

@inproceedings{piryani-etal-2024-detecting,
    title = "Detecting Temporal Ambiguity in Questions",
    author = "Piryani, Bhawna  and
      Abdallah, Abdelrahman  and
      Mozafari, Jamshid  and
      Jatowt, Adam",
    editor = "Al-Onaizan, Yaser  and
      Bansal, Mohit  and
      Chen, Yun-Nung",
    booktitle = "Findings of the Association for Computational Linguistics: EMNLP 2024",
    month = nov,
    year = "2024",
    address = "Miami, Florida, USA",
    publisher = "Association for Computational Linguistics",
    url = "https://aclanthology.org/2024.findings-emnlp.562/",
    doi = "10.18653/v1/2024.findings-emnlp.562",
    pages = "9620--9634",
}

@article{leeuwenberg2019survey,
  title={A survey on temporal reasoning for temporal information extraction from text},
  author={Leeuwenberg, Artuur and Moens, Marie-Francine},
  journal={Journal of Artificial Intelligence Research},
  volume={66},
  pages={341--380},
  year={2019}
}

@inproceedings{saquete2003terseo,
  title={Terseo: Temporal expression resolution system applied to event ordering},
  author={Saquete, Estela and Munoz, Rafael and Mart{\'\i}nez-Barco, Patricio},
  booktitle={International Conference on Text, Speech and Dialogue},
  pages={220--228},
  year={2003},
  organization={Springer}
}

@inproceedings{feng-etal-2023-generic,
    title = "Generic Temporal Reasoning with Differential Analysis and Explanation",
    author = "Feng, Yu  and
      Zhou, Ben  and
      Wang, Haoyu  and
      Jin, Helen  and
      Roth, Dan",
    editor = "Rogers, Anna  and
      Boyd-Graber, Jordan  and
      Okazaki, Naoaki",
    booktitle = "Proceedings of the 61st Annual Meeting of the Association for Computational Linguistics (Volume 1: Long Papers)",
    month = jul,
    year = "2023",
    address = "Toronto, Canada",
    publisher = "Association for Computational Linguistics",
    url = "https://aclanthology.org/2023.acl-long.671/",
    doi = "10.18653/v1/2023.acl-long.671",
    pages = "12013--12029"
}

@inproceedings{wang-zhao-2024-tram,
    title = "{TRAM}: Benchmarking Temporal Reasoning for Large Language Models",
    author = "Wang, Yuqing  and
      Zhao, Yun",
    editor = "Ku, Lun-Wei  and
      Martins, Andre  and
      Srikumar, Vivek",
    booktitle = "Findings of the Association for Computational Linguistics: ACL 2024",
    month = aug,
    year = "2024",
    address = "Bangkok, Thailand",
    publisher = "Association for Computational Linguistics",
    url = "https://aclanthology.org/2024.findings-acl.382/",
    doi = "10.18653/v1/2024.findings-acl.382",
    pages = "6389--6415",
}

@inproceedings{zhang-etal-2024-narrative,
    title = "Narrative-of-Thought: Improving Temporal Reasoning of Large Language Models via Recounted Narratives",
    author = "Zhang, Xinliang Frederick  and
      Beauchamp, Nicholas  and
      Wang, Lu",
    editor = "Al-Onaizan, Yaser  and
      Bansal, Mohit  and
      Chen, Yun-Nung",
    booktitle = "Findings of the Association for Computational Linguistics: EMNLP 2024",
    month = nov,
    year = "2024",
    address = "Miami, Florida, USA",
    publisher = "Association for Computational Linguistics",
    url = "https://aclanthology.org/2024.findings-emnlp.963/",
    doi = "10.18653/v1/2024.findings-emnlp.963",
    pages = "16507--16530",
}

@inproceedings{anand2012index,
  title={Index maintenance for time-travel text search},
  author={Anand, Avishek and Bedathur, Srikanta and Berberich, Klaus and Schenkel, Ralf},
  booktitle={Proceedings of the 35th international ACM SIGIR conference on Research and development in Information Retrieval},
  pages={235--244},
  year={2012}
}

@inproceedings{singh2016history,
  title={History by diversity: Helping historians search news archives},
  author={Singh, Jaspreet and Nejdl, Wolfgang and Anand, Avishek},
  booktitle={Proceedings of the 2016 ACM on conference on human information interaction and retrieval},
  pages={183--192},
  year={2016}
}

@inproceedings{setty2017modeling,
  title={Modeling event importance for ranking daily news events},
  author={Setty, Vinay and Anand, Abhijit and Mishra, Arunav and Anand, Avishek},
  booktitle={Proceedings of the Tenth ACM International Conference on Web Search and Data Mining},
  pages={231--240},
  year={2017}
}

@inproceedings{holzmann2016tempas,
  title={Tempas: Temporal archive search based on tags},
  author={Holzmann, Helge and Anand, Avishek},
  booktitle={Proceedings of the 25th International Conference Companion on World Wide Web},
  pages={207--210},
  year={2016}
}

@article{jatowt2015generic,
title = {Generic method for detecting focus time of documents},
journal = {Information Processing \& Management},
volume = {51},
number = {6},
pages = {851-868},
year = {2015},
issn = {0306-4573},
doi = {https://doi.org/10.1016/j.ipm.2015.05.001},
url = {https://www.sciencedirect.com/science/article/pii/S0306457315000527},
author = {Adam Jatowt and Ching Man {Au Yeung} and Katsumi Tanaka}
}

@inproceedings{zhao-etal-2024-set,
    title = "Set the Clock: Temporal Alignment of Pretrained Language Models",
    author = "Zhao, Bowen  and
      Brumbaugh, Zander  and
      Wang, Yizhong  and
      Hajishirzi, Hannaneh  and
      Smith, Noah",
    editor = "Ku, Lun-Wei  and
      Martins, Andre  and
      Srikumar, Vivek",
    booktitle = "Findings of the Association for Computational Linguistics: ACL 2024",
    month = aug,
    year = "2024",
    address = "Bangkok, Thailand",
    publisher = "Association for Computational Linguistics",
    url = "https://aclanthology.org/2024.findings-acl.892/",
    doi = "10.18653/v1/2024.findings-acl.892",
    pages = "15015--15040",
}

@inproceedings{realtimeqa,
 author = {Kasai, Jungo and Sakaguchi, Keisuke and takahashi, yoichi and Le Bras, Ronan and Asai, Akari and Yu, Xinyan and Radev, Dragomir and Smith, Noah A and Choi, Yejin and Inui, Kentaro},
 booktitle = {Advances in Neural Information Processing Systems},
 editor = {A. Oh and T. Naumann and A. Globerson and K. Saenko and M. Hardt and S. Levine},
 pages = {49025--49043},
 publisher = {Curran Associates, Inc.},
 title = {RealTime QA: What\textquotesingle s the Answer Right Now?},
 url = {https://proceedings.neurips.cc/paper_files/paper/2023/file/9941624ef7f867a502732b5154d30cb7-Paper-Datasets_and_Benchmarks.pdf},
 volume = {36},
 year = {2023}
}

@inproceedings{yang-etal-2023-upon,
    title = "Once Upon a ${\it Time}$ in ${\it Graph}$: Relative-Time Pretraining for Complex Temporal Reasoning",
    author = "Yang, Sen  and
      Li, Xin  and
      Bing, Lidong  and
      Lam, Wai",
    editor = "Bouamor, Houda  and
      Pino, Juan  and
      Bali, Kalika",
    booktitle = "Proceedings of the 2023 Conference on Empirical Methods in Natural Language Processing",
    month = dec,
    year = "2023",
    address = "Singapore",
    publisher = "Association for Computational Linguistics",
    url = "https://aclanthology.org/2023.emnlp-main.728/",
    doi = "10.18653/v1/2023.emnlp-main.728",
    pages = "11879--11895",
}

@inproceedings{liu-quan-2025-retrieval,
    title = "Retrieval of Temporal Event Sequences from Textual Descriptions",
    author = "Liu, Zefang  and
      Quan, Yinzhu",
    editor = "Shi, Weijia  and
      Yu, Wenhao  and
      Asai, Akari  and
      Jiang, Meng  and
      Durrett, Greg  and
      Hajishirzi, Hannaneh  and
      Zettlemoyer, Luke",
    booktitle = "Proceedings of the 4th International Workshop on Knowledge-Augmented Methods for Natural Language Processing",
    month = may,
    year = "2025",
    address = "Albuquerque, New Mexico, USA",
    publisher = "Association for Computational Linguistics",
    url = "https://aclanthology.org/2025.knowledgenlp-1.3/",
    pages = "37--49",
    ISBN = "979-8-89176-229-9",
}

@article{vrandevcic2014wikidata,
author = {Vrande\v{c}i\'{c}, Denny and Kr\"{o}tzsch, Markus},
title = {Wikidata: a free collaborative knowledgebase},
year = {2014},
issue_date = {October 2014},
publisher = {Association for Computing Machinery},
address = {New York, NY, USA},
volume = {57},
number = {10},
issn = {0001-0782},
url = {https://doi.org/10.1145/2629489},
doi = {10.1145/2629489},
journal = {Commun. ACM},
month = sep,
pages = {78–85},
numpages = {8}
}

@inproceedings{NEURIPS2021_customnews,
 author = {Lazaridou, Angeliki and Kuncoro, Adhi and Gribovskaya, Elena and Agrawal, Devang and Liska, Adam and Terzi, Tayfun and Gimenez, Mai and de Masson d\textquotesingle Autume, Cyprien and Kocisky, Tomas and Ruder, Sebastian and Yogatama, Dani and Cao, Kris and Young, Susannah and Blunsom, Phil},
 booktitle = {Advances in Neural Information Processing Systems},
 editor = {M. Ranzato and A. Beygelzimer and Y. Dauphin and P.S. Liang and J. Wortman Vaughan},
 pages = {29348--29363},
 publisher = {Curran Associates, Inc.},
 title = {Mind the Gap: Assessing Temporal Generalization in Neural Language Models},
 url = {https://proceedings.neurips.cc/paper_files/paper/2021/file/f5bf0ba0a17ef18f9607774722f5698c-Paper.pdf},
 volume = {34},
 year = {2021}
}

@article{su2024temporal,
  title={Temporal Knowledge Graph Question Answering: A Survey},
  author={Su, Miao and Li, Zixuan and Chen, Zhuo and Bai, Long and Jin, Xiaolong and Guo, Jiafeng},
  journal={arXiv preprint arXiv:2406.14191},
  year={2024}
}

@article{zhou2008evaluation,
  title={The evaluation of a temporal reasoning system in processing clinical discharge summaries},
  author={Zhou, Li and Parsons, Simon and Hripcsak, George},
  journal={Journal of the American Medical Informatics Association},
  volume={15},
  number={1},
  pages={99--106},
  year={2008},
  publisher={BMJ Group BMA House, Tavistock Square, London, WC1H 9JR}
}

@inproceedings{tao2010time,
  title={Time-oriented question answering from clinical narratives sing semantic-web techniques},
  author={Tao, Cui and Solbrig, Harold R and Sharma, Deepak K and Wei, Wei-Qi and Savova, Guergana K and Chute, Christopher G},
  booktitle={Proceedings of the 9th international semantic web conference on The semantic web-Volume Part II},
  pages={241--256},
  year={2010}
}

@article{zhou2007temporal,
  title={Temporal reasoning with medical data—a review with emphasis on medical natural language processing},
  author={Zhou, Li and Hripcsak, George},
  journal={Journal of biomedical informatics},
  volume={40},
  number={2},
  pages={183--202},
  year={2007},
  publisher={Elsevier}
}

@article{OLEX2021103784,
title = {Review of Temporal Reasoning in the Clinical Domain for Timeline Extraction: Where we are and where we need to be},
journal = {Journal of Biomedical Informatics},
volume = {118},
pages = {103784},
year = {2021},
issn = {1532-0464},
doi = {https://doi.org/10.1016/j.jbi.2021.103784},
url = {https://www.sciencedirect.com/science/article/pii/S1532046421001131},
author = {Amy L. Olex and Bridget T. McInnes}
}

@inproceedings{vladika-matthes-2024-improving,
    title = "Improving Health Question Answering with Reliable and Time-Aware Evidence Retrieval",
    author = "Vladika, Juraj  and
      Matthes, Florian",
    editor = "Duh, Kevin  and
      Gomez, Helena  and
      Bethard, Steven",
    booktitle = "Findings of the Association for Computational Linguistics: NAACL 2024",
    month = jun,
    year = "2024",
    address = "Mexico City, Mexico",
    publisher = "Association for Computational Linguistics",
    url = "https://aclanthology.org/2024.findings-naacl.295/",
    doi = "10.18653/v1/2024.findings-naacl.295",
    pages = "4752--4763"
}

@inproceedings{zhong2020jec,
  title={JEC-QA: a legal-domain question answering dataset},
  author={Zhong, Haoxi and Xiao, Chaojun and Tu, Cunchao and Zhang, Tianyang and Liu, Zhiyuan and Sun, Maosong},
  booktitle={Proceedings of the AAAI conference on artificial intelligence},
  volume={34},
  number={05},
  pages={9701--9708},
  year={2020}
}

@inproceedings{chen2023equals,
author = {Chen, Andong and Yao, Feng and Zhao, Xinyan and Zhang, Yating and Sun, Changlong and Liu, Yun and Shen, Weixing},
title = {EQUALS: A Real-world Dataset for Legal Question Answering via Reading Chinese Laws},
year = {2023},
isbn = {9798400701979},
publisher = {Association for Computing Machinery},
address = {New York, NY, USA},
url = {https://doi.org/10.1145/3594536.3595159},
doi = {10.1145/3594536.3595159},
booktitle = {Proceedings of the Nineteenth International Conference on Artificial Intelligence and Law},
pages = {71–80},
numpages = {10},
location = {Braga, Portugal},
series = {ICAIL '23}
}

@inproceedings{buttner-habernal-2024-answering,
    title = "Answering legal questions from laymen in {G}erman civil law system",
    author = {B{\"u}ttner, Marius  and
      Habernal, Ivan},
    editor = "Graham, Yvette  and
      Purver, Matthew",
    booktitle = "Proceedings of the 18th Conference of the European Chapter of the Association for Computational Linguistics (Volume 1: Long Papers)",
    month = mar,
    year = "2024",
    address = "St. Julian{'}s, Malta",
    publisher = "Association for Computational Linguistics",
    url = "https://aclanthology.org/2024.eacl-long.122/",
    doi = "10.18653/v1/2024.eacl-long.122",
    pages = "2015--2027"
}

@inproceedings{kien-etal-2020-answering,
    title = "Answering Legal Questions by Learning Neural Attentive Text Representation",
    author = "Kien, Phi Manh  and
      Nguyen, Ha-Thanh  and
      Bach, Ngo Xuan  and
      Tran, Vu  and
      Nguyen, Minh Le  and
      Phuong, Tu Minh",
    editor = "Scott, Donia  and
      Bel, Nuria  and
      Zong, Chengqing",
    booktitle = "Proceedings of the 28th International Conference on Computational Linguistics",
    month = dec,
    year = "2020",
    address = "Barcelona, Spain (Online)",
    publisher = "International Committee on Computational Linguistics",
    url = "https://aclanthology.org/2020.coling-main.86/",
    doi = "10.18653/v1/2020.coling-main.86",
    pages = "988--998"
}

@inproceedings{t-y-s-s-etal-2024-chronoslex,
    title = "{C}hronos{L}ex: Time-aware Incremental Training for Temporal Generalization of Legal Classification Tasks",
    author = "T.y.s.s, Santosh  and
      Vuong, Tuan-Quang  and
      Grabmair, Matthias",
    editor = "Ku, Lun-Wei  and
      Martins, Andre  and
      Srikumar, Vivek",
    booktitle = "Proceedings of the 62nd Annual Meeting of the Association for Computational Linguistics (Volume 1: Long Papers)",
    month = aug,
    year = "2024",
    address = "Bangkok, Thailand",
    publisher = "Association for Computational Linguistics",
    url = "https://aclanthology.org/2024.acl-long.166/",
    doi = "10.18653/v1/2024.acl-long.166",
    pages = "3022--3039"
}

@inproceedings{chen-etal-2021-finqa,
    title = "{F}in{QA}: A Dataset of Numerical Reasoning over Financial Data",
    author = "Chen, Zhiyu  and
      Chen, Wenhu  and
      Smiley, Charese  and
      Shah, Sameena  and
      Borova, Iana  and
      Langdon, Dylan  and
      Moussa, Reema  and
      Beane, Matt  and
      Huang, Ting-Hao  and
      Routledge, Bryan  and
      Wang, William Yang",
    editor = "Moens, Marie-Francine  and
      Huang, Xuanjing  and
      Specia, Lucia  and
      Yih, Scott Wen-tau",
    booktitle = "Proceedings of the 2021 Conference on Empirical Methods in Natural Language Processing",
    month = nov,
    year = "2021",
    address = "Online and Punta Cana, Dominican Republic",
    publisher = "Association for Computational Linguistics",
    url = "https://aclanthology.org/2021.emnlp-main.300/",
    doi = "10.18653/v1/2021.emnlp-main.300",
    pages = "3697--3711"
}

@inproceedings{chen-etal-2024-fintextqa,
    title = "{F}in{T}ext{QA}: A Dataset for Long-form Financial Question Answering",
    author = "Chen, Jian  and
      Zhou, Peilin  and
      Hua, Yining  and
      Xin, Loh  and
      Chen, Kehui  and
      Li, Ziyuan  and
      Zhu, Bing  and
      Liang, Junwei",
    editor = "Ku, Lun-Wei  and
      Martins, Andre  and
      Srikumar, Vivek",
    booktitle = "Proceedings of the 62nd Annual Meeting of the Association for Computational Linguistics (Volume 1: Long Papers)",
    month = aug,
    year = "2024",
    address = "Bangkok, Thailand",
    publisher = "Association for Computational Linguistics",
    url = "https://aclanthology.org/2024.acl-long.328/",
    doi = "10.18653/v1/2024.acl-long.328",
    pages = "6025--6047"
}

@article{wenzel2023overview,
  title={An overview of temporal commonsense reasoning and acquisition},
  author={Wenzel, Georg and Jatowt, Adam},
  journal={arXiv preprint arXiv:2308.00002},
  year={2023}
}

@inproceedings{vashishtha-etal-2019-fine,
    title = "Fine-Grained Temporal Relation Extraction",
    author = "Vashishtha, Siddharth  and
      Van Durme, Benjamin  and
      White, Aaron Steven",
    editor = "Korhonen, Anna  and
      Traum, David  and
      M{\`a}rquez, Llu{\'i}s",
    booktitle = "Proceedings of the 57th Annual Meeting of the Association for Computational Linguistics",
    month = jul,
    year = "2019",
    address = "Florence, Italy",
    publisher = "Association for Computational Linguistics",
    url = "https://aclanthology.org/P19-1280/",
    doi = "10.18653/v1/P19-1280",
    pages = "2906--2919"
}

@inproceedings{allein-etal-2023-implicit,
    title = "Implicit Temporal Reasoning for Evidence-Based Fact-Checking",
    author = "Allein, Liesbeth  and
      Saelens, Marlon  and
      Cartuyvels, Ruben  and
      Moens, Marie-Francine",
    editor = "Vlachos, Andreas  and
      Augenstein, Isabelle",
    booktitle = "Findings of the Association for Computational Linguistics: EACL 2023",
    month = may,
    year = "2023",
    address = "Dubrovnik, Croatia",
    publisher = "Association for Computational Linguistics",
    url = "https://aclanthology.org/2023.findings-eacl.13/",
    doi = "10.18653/v1/2023.findings-eacl.13",
    pages = "176--189"
}

@inproceedings{wu2025videoqa,
  title={VideoQA-TA: Temporal-Aware Multi-Modal Video Question Answering},
  author={Wu, Zhixuan and Cheng, Bo and Han, Jiale and Ma, Jiabao and Zhang, Shuhao and Chen, Yuli and Li, Changbo},
  booktitle={Proceedings of the 31st International Conference on Computational Linguistics},
  pages={7239--7252},
  year={2025}
}

@article{kolomiyets2011survey,
  title={A survey on question answering technology from an information retrieval perspective},
  author={Kolomiyets, Oleksandr and Moens, Marie-Francine},
  journal={Information Sciences},
  volume={181},
  number={24},
  pages={5412--5434},
  year={2011},
  publisher={Elsevier}
}

@inproceedings{deng-etal-2025-enhancing,
    title = "Enhancing Temporal Understanding in {LLM}s for Semi-structured Tables",
    author = "Deng, Irwin  and
      Dixit, Kushagra  and
      Roth, Dan  and
      Gupta, Vivek",
    editor = "Chiruzzo, Luis  and
      Ritter, Alan  and
      Wang, Lu",
    booktitle = "Findings of the Association for Computational Linguistics: NAACL 2025",
    month = apr,
    year = "2025",
    address = "Albuquerque, New Mexico",
    publisher = "Association for Computational Linguistics",
    url = "https://aclanthology.org/2025.findings-naacl.278/",
    doi = "10.18653/v1/2025.findings-naacl.278",
    pages = "4936--4955",
    ISBN = "979-8-89176-195-7"
}

@inproceedings{kulkarni-etal-2025-llm,
    title = "{LLM}-Symbolic Integration for Robust Temporal Tabular Reasoning",
    author = "Kulkarni, Atharv  and
      Dixit, Kushagra  and
      Srikumar, Vivek  and
      Roth, Dan  and
      Gupta, Vivek",
    editor = "Che, Wanxiang  and
      Nabende, Joyce  and
      Shutova, Ekaterina  and
      Pilehvar, Mohammad Taher",
    booktitle = "Findings of the Association for Computational Linguistics: ACL 2025",
    month = jul,
    year = "2025",
    address = "Vienna, Austria",
    publisher = "Association for Computational Linguistics",
    url = "https://aclanthology.org/2025.findings-acl.1022/",
    doi = "10.18653/v1/2025.findings-acl.1022",
    pages = "19914--19940",
    ISBN = "979-8-89176-256-5",
    
}

@inproceedings{shankarampeta-etal-2025-transienttables,
    title = "{TRANSIENTTABLES}: Evaluating {LLM}s' Reasoning on Temporally Evolving Semi-structured Tables",
    author = "Shankarampeta, Abhilash  and
      Mahajan, Harsh  and
      Kataria, Tushar  and
      Roth, Dan  and
      Gupta, Vivek",
    editor = "Chiruzzo, Luis  and
      Ritter, Alan  and
      Wang, Lu",
    booktitle = "Proceedings of the 2025 Conference of the Nations of the Americas Chapter of the Association for Computational Linguistics: Human Language Technologies (Volume 1: Long Papers)",
    month = apr,
    year = "2025",
    address = "Albuquerque, New Mexico",
    publisher = "Association for Computational Linguistics",
    url = "https://aclanthology.org/2025.naacl-long.332/",
    doi = "10.18653/v1/2025.naacl-long.332",
    pages = "6526--6544",
    ISBN = "979-8-89176-189-6"
}

@inproceedings{wang2025timerag,
author = {Wang, Zhao and Zhao, Ziliang and Dou, Zhicheng},
title = {TimeRAG: Enhancing Complex Temporal Reasoning with Search Engine Augmentation},
year = {2025},
isbn = {9798400720406},
publisher = {Association for Computing Machinery},
address = {New York, NY, USA},
url = {https://doi.org/10.1145/3746252.3761425},
doi = {10.1145/3746252.3761425},
booktitle = {Proceedings of the 34th ACM International Conference on Information and Knowledge Management},
pages = {3230–3239},
numpages = {10},
location = {Seoul, Republic of Korea},
series = {CIKM '25}
}

@inproceedings{zhu-etal-2023-question,
    title = "Question Answering as Programming for Solving Time-Sensitive Questions",
    author = "Zhu, Xinyu  and
      Yang, Cheng  and
      Chen, Bei  and
      Li, Siheng  and
      Lou, Jian-Guang  and
      Yang, Yujiu",
    editor = "Bouamor, Houda  and
      Pino, Juan  and
      Bali, Kalika",
    booktitle = "Proceedings of the 2023 Conference on Empirical Methods in Natural Language Processing",
    month = dec,
    year = "2023",
    address = "Singapore",
    publisher = "Association for Computational Linguistics",
    url = "https://aclanthology.org/2023.emnlp-main.787/",
    doi = "10.18653/v1/2023.emnlp-main.787",
    pages = "12775--12790"
}

@inproceedings{bazaga-etal-2025-learning,
    title = "Learning to Reason Over Time: Timeline Self-Reflection for Improved Temporal Reasoning in Language Models",
    author = "Bazaga, Adri{\'a}n  and
      Blloshmi, Rexhina  and
      Byrne, Bill  and
      de Gispert, Adri{\`a}",
    editor = "Che, Wanxiang  and
      Nabende, Joyce  and
      Shutova, Ekaterina  and
      Pilehvar, Mohammad Taher",
    booktitle = "Proceedings of the 63rd Annual Meeting of the Association for Computational Linguistics (Volume 1: Long Papers)",
    month = jul,
    year = "2025",
    address = "Vienna, Austria",
    publisher = "Association for Computational Linguistics",
    url = "https://aclanthology.org/2025.acl-long.1358/",
    doi = "10.18653/v1/2025.acl-long.1358",
    pages = "28014--28033",
    ISBN = "979-8-89176-251-0"
}

@inproceedings{
wu2025chainoftimeline,
title={Chain-of-Timeline: Enhancing {LLM} Zero-Shot Temporal Reasoning with {SQL}-Style Timeline Formalization},
author={Jiaying Wu and Bryan Hooi},
booktitle={Workshop on Reasoning and Planning for Large Language Models},
year={2025},
}

@article{li2023unlocking,
  title={Unlocking Temporal Question Answering for Large Language Models with Tailor-Made Reasoning Logic},
  author={Li, Xingxuan and Cheng, Liying and Tan, Qingyu and Ng, Hwee Tou and Joty, Shafiq and Bing, Lidong},
  journal={arXiv preprint arXiv:2305.15014},
  year={2023}
}

@inproceedings{chu-etal-2023-mtger,
    title = "{MTGER}: Multi-view Temporal Graph Enhanced Temporal Reasoning over Time-Involved Document",
    author = "Chu, Zheng  and
      Wang, Zekun  and
      Liang, Jiafeng  and
      Liu, Ming  and
      Qin, Bing",
    editor = "Bouamor, Houda  and
      Pino, Juan  and
      Bali, Kalika",
    booktitle = "Findings of the Association for Computational Linguistics: EMNLP 2023",
    month = dec,
    year = "2023",
    address = "Singapore",
    publisher = "Association for Computational Linguistics",
    url = "https://aclanthology.org/2023.findings-emnlp.1016/",
    doi = "10.18653/v1/2023.findings-emnlp.1016",
    pages = "15218--15233",
}

@inproceedings{choi2025finder,
author = {Choi, Chanyeol and Kwon, Jihoon and Ha, Jaeseon and Choi, Hojun and Kim, Chaewoon and Lee, Yongjae and Sohn, Jy-yong and Lopez-Lira, Alejandro},
title = {FinDER: Financial Dataset for Question Answering and Evaluating Retrieval-Augmented Generation},
year = {2025},
isbn = {9798400722202},
publisher = {Association for Computing Machinery},
address = {New York, NY, USA},
url = {https://doi.org/10.1145/3768292.3770361},
doi = {10.1145/3768292.3770361},
booktitle = {Proceedings of the 6th ACM International Conference on AI in Finance},
pages = {638–646},
numpages = {9},
location = {
},
series = {ICAIF '25}
}

@inproceedings{
fatemi2024test,
title={Test of Time: A Benchmark for Evaluating {LLM}s on Temporal Reasoning},
author={Bahare Fatemi and Mehran Kazemi and Anton Tsitsulin and Karishma Malkan and Jinyeong Yim and John Palowitch and Sungyong Seo and Jonathan Halcrow and Bryan Perozzi},
booktitle={The Thirteenth International Conference on Learning Representations},
year={2025},
url={https://openreview.net/forum?id=44CoQe6VCq}
}

@inproceedings{ge2025tremu,
    title = "{TR}e{M}u: Towards Neuro-Symbolic Temporal Reasoning for {LLM}-Agents with Memory in Multi-Session Dialogues",
    author = "Ge, Yubin  and
      Romeo, Salvatore  and
      Cai, Jason  and
      Shu, Raphael  and
      Benajiba, Yassine  and
      Sunkara, Monica  and
      Zhang, Yi",
    editor = "Che, Wanxiang  and
      Nabende, Joyce  and
      Shutova, Ekaterina  and
      Pilehvar, Mohammad Taher",
    booktitle = "Findings of the Association for Computational Linguistics: ACL 2025",
    month = jul,
    year = "2025",
    address = "Vienna, Austria",
    publisher = "Association for Computational Linguistics",
    url = "https://aclanthology.org/2025.findings-acl.972/",
    doi = "10.18653/v1/2025.findings-acl.972",
    pages = "18974--18988",
    ISBN = "979-8-89176-256-5",
}

@inproceedings{gruber2024complextempqa,
    title = "{C}omplex{T}emp{QA}: A 100m Dataset for Complex Temporal Question Answering",
    author = {Gruber, Raphael  and
      Abdallah, Abdelrahman  and
      F{\"a}rber, Michael  and
      Jatowt, Adam},
    editor = "Christodoulopoulos, Christos  and
      Chakraborty, Tanmoy  and
      Rose, Carolyn  and
      Peng, Violet",
    booktitle = "Proceedings of the 2025 Conference on Empirical Methods in Natural Language Processing",
    month = nov,
    year = "2025",
    address = "Suzhou, China",
    publisher = "Association for Computational Linguistics",
    url = "https://aclanthology.org/2025.emnlp-main.463/",
    doi = "10.18653/v1/2025.emnlp-main.463",
    pages = "9100--9112",
    ISBN = "979-8-89176-332-6",
}

@article{liang2025nestr, 
    title={NeSTR: A Neuro-Symbolic Abductive Framework for Temporal Reasoning in Large Language Models}, 
    volume={40}, 
    url={https://ojs.aaai.org/index.php/AAAI/article/view/40460}, 
    DOI={10.1609/aaai.v40i38.40460},  
    number={38}, 
    journal={Proceedings of the AAAI Conference on Artificial Intelligence}, 
    author={Liang, Feng and Zeng, Weixin and Zhao, Runhao and Zhao, Xiang}, 
    year={2026}, 
    month={Mar.}, 
    pages={31907-31915} 
}

@inproceedings{siyue2024mrag,
    title = "{MRAG}: A Modular Retrieval Framework for Time-Sensitive Question Answering",
    author = "Zhang, Siyue  and
      Xue, Yuxiang  and
      Zhang, Yiming  and
      Wu, Xiaobao  and
      Luu, Anh Tuan  and
      Zhao, Chen",
    editor = "Christodoulopoulos, Christos  and
      Chakraborty, Tanmoy  and
      Rose, Carolyn  and
      Peng, Violet",
    booktitle = "Findings of the Association for Computational Linguistics: EMNLP 2025",
    month = nov,
    year = "2025",
    address = "Suzhou, China",
    publisher = "Association for Computational Linguistics",
    url = "https://aclanthology.org/2025.findings-emnlp.167/",
    doi = "10.18653/v1/2025.findings-emnlp.167",
    pages = "3080--3118",
    ISBN = "979-8-89176-335-7",
}

@inproceedings{
shangguan2024tomato,
title={{TOMATO}: Assessing Visual Temporal Reasoning Capabilities in Multimodal Foundation Models},
author={Ziyao Shangguan and Chuhan Li and Yuxuan Ding and Yanan Zheng and Yilun Zhao and Tesca Fitzgerald and Arman Cohan},
booktitle={The Thirteenth International Conference on Learning Representations},
year={2025},
url={https://openreview.net/forum?id=fCi4o83Mfs}
}

@article{zhu2023large,
author = {Zhu, Yutao and Yuan, Huaying and Wang, Shuting and Liu, Jiongnan and Liu, Wenhan and Deng, Chenlong and Chen, Haonan and Liu, Zheng and Dou, Zhicheng and Wen, Ji-Rong},
title = {Large Language Models for Information Retrieval: A Survey},
year = {2025},
issue_date = {January 2026},
publisher = {Association for Computing Machinery},
address = {New York, NY, USA},
volume = {44},
number = {1},
issn = {1046-8188},
url = {https://doi.org/10.1145/3748304},
doi = {10.1145/3748304},
journal = {ACM Trans. Inf. Syst.},
month = nov,
articleno = {12},
numpages = {54}
}

@inproceedings{uddin2024unseentimeqa,
    title = "{U}n{S}een{T}ime{QA}: Time-Sensitive Question-Answering Beyond {LLM}s' Memorization",
    author = "Uddin, Md Nayem  and
      Saeidi, Amir  and
      Handa, Divij  and
      Seth, Agastya  and
      Son, Tran Cao  and
      Blanco, Eduardo  and
      Corman, Steven  and
      Baral, Chitta",
    editor = "Che, Wanxiang  and
      Nabende, Joyce  and
      Shutova, Ekaterina  and
      Pilehvar, Mohammad Taher",
    booktitle = "Proceedings of the 63rd Annual Meeting of the Association for Computational Linguistics (Volume 1: Long Papers)",
    month = jul,
    year = "2025",
    address = "Vienna, Austria",
    publisher = "Association for Computational Linguistics",
    url = "https://aclanthology.org/2025.acl-long.94/",
    doi = "10.18653/v1/2025.acl-long.94",
    pages = "1873--1913",
    ISBN = "979-8-89176-251-0"
}

@article{piryani2026often,
  title={How often do Answers Change? Estimating Recency Requirements in Question Answering},
  author={Piryani, Bhawna and Mert, Zehra and Jatowt, Adam},
  journal={arXiv preprint arXiv:2603.16544},
  year={2026}
}

\newpage

\appendix

% \section{Appendix}
% \label{sec:appendix}

\section{Related Surveys}
\label{appendix:Related_survey}

Advances in temporal datasets, time-aware models, and temporal reasoning techniques have enabled systems capable of retrieving time-relevant documents, ordering events, and answering temporally constrained questions across applications such as historical analysis, fact-checking, and intelligent assistants. Although temporal information processing has received considerable attention in the literature, existing surveys have not provided a comprehensive, end-to-end analysis of TQA systems over unstructured text. Prior work can be grouped into broad areas, each with limitations relative to our focus.

While QA has been widely surveyed, most existing reviews focus on general techniques, often neglecting temporal aspects. IR surveys emphasize ranking functions, neural retrieval models, and query understanding \cite{li2024matching, zhu2023large}, whereas QA surveys center on extractive, abstractive, or multi-hop answering \cite{zhu2021retrieving, mavi2024multi}. These works rarely consider temporal intent, dynamic or evolving information needs, or event sequencing, highlighting a key gap in understanding temporally conditioned QA.

Several earlier works provided foundational insights into TIR. \citet{alonso2011temporal} discusses challenges such as real-time streams, exploratory temporal search, and spatio-temporal retrieval. \citet{campos2014survey} offers a broad overview of document dating, time-aware ranking, and query understanding, covering both explicit and implicit time signals. \citet{kanhabua2016temporal} complements these with a tutorial on temporal indexing and ranking, emphasizing the detection of temporal query intent. However, despite substantial progress in temporal reasoning with modern LLMs, there has been no recent systematic overview that integrates contemporary text-based TQA datasets, models, and evaluation paradigms.

Crucially, existing temporal IR and temporal processing surveys adopt a largely component-level perspective, focusing on tasks such as temporal tagging, document dating, or time-aware ranking, rather than temporal question answering as an end-to-end problem. Likewise, general QA surveys typically treat time as an implicit or secondary factor and do not systematically analyze temporally constrained queries or diachronic evidence selection. Our survey fills this gap by explicitly framing text-based TQA as a distinct problem setting and synthesizing recent datasets, models, and evaluation paradigms that capture the interaction between temporal reasoning, retrieval, and answer generation.

Temporal QA has also been studied in structured and semi-structured settings, including temporal knowledge graphs \cite{jia2021complex, saxena-etal-2021-question, chen-etal-2023-multi, Xiong_Yang_Payani_Kerce_Fekri_2024} and, more recently, evolving tables. Tabular temporal QA approaches reason over time-indexed records or attribute updates \cite{shankarampeta-etal-2025-transienttables,deng-etal-2025-enhancing,kulkarni-etal-2025-llm}, relying on explicit temporal schemata, row-level timestamps, and structured event representations. Their central challenges, such as temporal joins, value-update tracking, and schema-aware inference, differ fundamentally from grounding implicit temporal expressions, resolving document-relative timestamps, or retrieving diachronic evidence in natural language. Because our survey focuses on temporality as expressed in unstructured text, we treat KG-based and table-based TQA as out of scope while acknowledging their relevance as complementary research areas. Hybrid datasets such as TempTabQA \cite{gupta-etal-2023-temptabqa} sit at the boundary by extracting semi-structured information (e.g., infoboxes) from text, yet still rely primarily on natural-language context.

Our survey focuses on temporally aware QA over unstructured text. We review both traditional and neural approaches to core tasks such as temporal tagging, event dating, time-aware retrieval, and temporal reasoning. To our knowledge, no prior survey brings together recent developments across these tasks in the context of text-based TQA. Other related topics, including temporal fact verification \cite{barik-etal-2024-time} and timeline summarization \cite{sojitra2024timeline}, are discussed only when directly relevant.

%As a parallel research line, TQA over knowledge graphs has gained considerable attention \cite{jia2021complex, saxena-etal-2021-question, chen-etal-2023-multi, Xiong_Yang_Payani_Kerce_Fekri_2024}. Our survey focuses on temporally aware QA over text. We review both traditional and neural approaches to core tasks such as temporal tagging, event dating, time-aware retrieval, and temporal reasoning. To our knowledge, no prior survey brings together recent developments across these tasks in the context of text-based QA. Other related topics, including temporal fact verification \cite{barik-etal-2024-time} and timeline summarization \cite{sojitra2024timeline}, are discussed only when directly relevant.

\section{Key Concepts}
\label{appendix:key_concepts}
We first introduce the core concepts related to TQA.
%, establishing a vocabulary commonly used throughout the field. 
\textbf{Temporal Information Retrieval (TIR)} aims to retrieve documents that are not only topically relevant but also aligned with the query's \textbf{temporal intent}. This intent may be explicit, such as \textit{“Olympics 2024”}, or implicit, such as \textit{“latest Apple earnings”}. TIR relies on different \textbf{temporal signals}, including \textbf{document timestamps} (publication dates), \textbf{temporal expressions} (\textit{“March 2023”}), and \textbf{event mentions} (\textit{“2024 Olympics”}) to assess a document’s \textbf{temporal relevance}, which indicates how well its content aligns with the query's intended time frame \cite{kanhabua2008improving,singh2016history}.
While TIR facilitates time-sensitive access to documents, it often lacks mechanisms for performing deeper temporal reasoning over the content itself. 

On the other hand, \textbf{Temporal Question Answering (TQA)} focuses on answering questions with \textbf{temporal constraints}, either explicitly stated, such as \textit{"Who won the Nobel Prize in Physics in 2020?"}, or implied, for instance, \textit{"What are the latest US climate policies?"}. Success in TQA requires understanding the \textbf{question's temporal intent} and retrieving documents relevant to the corresponding time frame, or published around that time. Unlike TIR, TQA often involves more sophisticated processing, including interpreting \textbf{temporal expressions}, \textbf{ordering events}, and grounding answers in the appropriate time frame. As a result, it frequently requires multi-hop reasoning across documents and temporal signals.

TQA relies on diverse temporal cues. \textit{Temporal signals} include \textbf{explicit temporal expressions} like \textit{"March 2023"}, \textbf{relative expressions} such as \textit{"last week"}, \textbf{implicit cues} like \textit{"recently"}, and event-based references like \textit{"2024 Olympics"}. These require contextual interpretation and sometimes external temporal knowledge. Temporal metadata, such as \textbf{document timestamps}, indicate publication time and often serve as proxies for judging the freshness of content.

A critical yet often underutilized concept is that of \textit{document focus time}, the time period a document primarily discusses, which may differ from its publication timestamp. For example, a news article published in 2013 about the 2010 Academy Awards has a publication date of 2013 but a focus time of 2010. This distinction is crucial in TQA, where users often seek information tied to specific historical or future contexts rather than content merely anchored to when it was written.

Estimating focus time is non-trivial, often requiring temporal expression normalization, burst detection, or event timestamping. When accurately modeled, it improves retrieval quality in time-sensitive settings, especially for long-range archives or temporally anchored questions. Prior work has shown that focus time estimation can enhance performance in tasks such as historical QA, event ordering, and temporally grounded search \cite{jatowt2013estimating, wang2020answering}.

\textbf{Temporal taggers} are essential tools in temporal information processing; they identify and standardize time expressions in text, such as \textit{“March 15, 2021”} or \textit{“yesterday,”} converting them into formats like YYYY-MM-DD and categorizing them (e.g., DATE, DURATION). Popular taggers like HeidelTime \cite{strotgen-gertz-2010-heideltime}, SUTime \cite{chang-manning-2012-sutime}, Temponym tagger \cite{kuzey2016temponym}, CogCompTime \cite{ning-etal-2018-cogcomptime} support a range of languages and domains, forming the foundation for downstream tasks including TQA, event ordering, and timeline construction.

Additionally, \textbf{Temponyms} \cite{kuzey2016time} are free-text phrases that implicitly refer to specific time periods or events but are not recognized as standard temporal expressions, for Instance, \textit{"Greek referendum"} or \textit{“Clinton’s presidency”}. Recognizing and resolving these expressions is essential for comprehensive temporal understanding.
Other 
%important temporal signals 
related concepts include \textbf{temporal granularity} (typically ranging from day to decade), \textbf{temporal proximity} (the temporal closeness of a document to the query's target time, influencing ranking), and \textbf{temporal distribution} patterns in retrieval results. Effectively leveraging these signals is key to building time-aware systems \cite{campos2014survey}.

\textbf{Temporal Disambiguation} resolves ambiguous time references (e.g., identifying which "Tuesday" is being discussed), addressing \textbf{temporal ambiguity} in both queries and documents \cite{piryani-etal-2024-detecting}. \textbf{Temporal Co-reference} involves identifying and linking different mentions of the same temporal entity within or across documents, such as connecting “that year” to “2020” \cite{ning-etal-2018-cogcomptime}.
%textual references to the same time-related entity within a document or across multiple documents, such as connecting "that year" to a previously mentioned "2020" \cite{ning-etal-2018-cogcomptime}.
\textbf{Timeline Extraction} automatically constructs a chronological sequence of events or facts from text, to answer questions requiring event ordering, such as constructing a historical timeline \cite{bedi-etal-2017-event}.

More advanced reasoning tasks include \textbf{Temporal Reasoning}, which infers time-related relationships, such as determining the order of events or calculating durations between them. It is crucial for answering complex questions like \textit{"What happened in Poland after World War II and before 1960?"} \cite{leeuwenberg2019survey}. \textbf{Temporal Aggregation} synthesizes information from multiple time periods to answer broad or comparative questions (e.g., \textit{“How has climate policy evolved over the last decade?”}). %\textbf{Temporal Constraints}, which are explicit or implicit time-based limitations within a query (e.g., \textit{“news from last week”}) \cite{berberich2010language}. 
\textbf{Temporal Robustness} \cite{wallat2025study} refers to the resiliency of systems to adversarial changes in time-related elements (e.g., altering a date in a query, or its position in a sentence) in the form of \textbf{temporal perturbations}. It is used in evaluation to assess temporal reasoning stability.

\section{Temporal Prediction Tasks}
\label{appendix:temp_prediction_task}
Temporal prediction tasks are crucial in understanding and organizing time-sensitive textual data. Despite sharing the common objective of grounding text in time, these tasks differ in focus, granularity, and application. In this section, we explore related temporal prediction tasks—document dating, document focus time estimation, temporal query profiling, and event occurrence time estimation, which provide complementary insights and support distinct applications. Each task addresses unique aspects of temporal analysis, from inferring document creation times to profiling query intent. Below, we review these tasks, their methodologies, and key contributions, emphasizing their roles in temporal IR and QA.

\subsection{Document Dating}
\label{appendix:document_dating_task}

Document dating refers to the task of estimating a document’s creation time (e.g., publication date) based on its textual content, especially when metadata is missing, unreliable, or unavailable. The input is the full document text, and the output is a timestamp, typically at year or month granularity.

 Early approaches, such as that by \citet{jong2005temporal}, leveraged unigram language models trained over distinct time periods to determine when a document's vocabulary was most prevalent. Building on this, \citet{kanhabua2008improving} integrated additional linguistic features such as part-of-speech tags, tf-idf scores, and collocations to better capture temporal patterns. \citet{dalli-2006-temporal} introduced an unsupervised method for automatic document dating using periodic word usage. \citet{kumar2012dating} trained language models over discretized time intervals (chronons) using Wikipedia biographies. \citet{niculae-etal-2014-temporal} model document dating as a pairwise ranking problem using logistic regression. More recently, \citet{vashishth-etal-2018-dating} introduced a neural method employing Graph Convolutional Networks (GCNs) to model syntactic and temporal relations jointly. 

 Document dating is crucial in temporal indexing, digital preservation, and metadata recovery, particularly for historical or noisy corpora. Beyond textual content analysis, several methods estimate the creation date of web resources. \citet{jatowt2007detecting} was the first approach for dating content of web pages. The authors estimated timestamps of individual content elements of web pages using their archived snapshots. \citet{salaheldeen2013carbon} developed Carbon Date, a tool that aggregates signals from multiple online sources, such as first tweets, archive snapshots, URL shorteners, and search engine crawls, to estimate a webpage's creation date.

\subsection{Document Focus Time Estimation}
\label{appendix:document_focus_time_estiamtion_task}

Document focus time estimation aims to identify the historical time periods that a document discusses, which may differ from its actual publication date. For example, a news article published in 2021 that analyzes the 9/11 attacks would have a focus time centered around September 2001. The input to this task is the document's full text, and the output consists of one or more temporal intervals that represent the document's narrative temporal scope. \citet{jatowt2013estimating} proposed a graph-based method that models co-occurrences between terms and dates to identify salient temporal associations within the text. Building on this, \citet{jatowt2015generic} introduced a method that estimates focus time using statistical evidence from external corpora, even when explicit temporal expressions are limited. \citet{shrivastava2017concept} further advanced this line of work by linking documents to Wikipedia concepts, leveraging their temporal relations to estimate focus times. This task supports historical analysis, event-centric retrieval, and timeline generation, providing insights into the temporal context of textual content.

\subsection{Temporal Query Profiling}
\label{appendix:temporal_query_profiling_task}

Temporal query profiling determines a query's temporal intent and time of interest, such as whether it refers to the past, future, or is atemporal. The input is a short keyword query (e.g., "Ukraine-Russia war"), and the output is an inferred time or temporal distribution. \citet{kanhabua2010determining} estimated query time by analyzing timestamps of top-k retrieved documents, while \citet{dakka2008answering} and \citet{jones2007temporal} modeled temporal distributions of relevant documents.  \citet{kanhabua2011comparison} conducted a comparative evaluation of five temporal ranking approaches (LMT, LMTU, TS, TSU, FuzzySet), evaluating their ability to model uncertainty and adapt to temporal variance. \citet{gupta2014identifying} combined timestamp metadata with temporal expressions in document content to infer precise time intervals. 
Temporal query profiling is essential for time-aware IR, as it enables query disambiguation, improves temporal relevance ranking, and supports applications such as event-centric search and timeline construction.

\subsection{Event Occurrence Time Estimation}
\label{appendix:Event_Occurrence_Time_Estimation_task}

Event occurrence time estimation aims to predict the specific date on which an event occurred, given a short textual description (e.g., "Plane crash in Armenia kills 36"). Unlike document-centric tasks, this focuses on the event mention itself and typically requires high-granularity outputs—such as day- or month-level timestamps. 

\citet{das2017estimating} introduced time vectors combining word and global temporal embeddings, estimating dates via cosine similarity. \citet{morbidoni2018leveraging} leveraged structured knowledge bases such as DBpedia and Wikipedia to link event descriptions to temporally grounded entities. \citet{honovich-etal-2020-machine} proposed a neural approach with sentence extraction, LSTM with attention, and an MLP classifier for date prediction. More recently, \citet{wang2021event} introduced TEP-Trans, a Transformer-based model 
that formulates event time prediction as a multivariate time series forecasting problem using features extracted from temporal news collections.

\paragraph{Summary:} While these temporal prediction tasks are highly interrelated, each aiming to anchor textual information within a temporal context, they address distinct facets of temporal understanding. Document dating predicts when a document was created, whereas document focus time estimation identifies when the content is about, which may precede or differ from the creation time. Temporal query profiling focuses on the user's intent, inferring when the query is directed in time rather than analyzing any specific document. Finally, event occurrence time estimation deals with precise, often fine-grained dating of event mentions, requiring models to infer real-world event timelines from sparse input. Together, these tasks form a complementary suite of temporal reasoning capabilities, enabling robust time-aware information retrieval and question answering systems.

\section{Rule-based \& Statistical Methods}
\label{sec:temporal_rulebased}

Early work in Temporal QA was dominated by rule-based systems and statistical models that laid the groundwork for core temporal tasks such as time expression normalization, event ordering, and temporal ranking. While limited in scalability and adaptability, they introduced many foundational concepts that remain relevant today.

In TIR, rule-based systems focused on extracting and normalizing time expressions to improve retrieval for time-sensitive queries \cite{arikan2009time, alonso2007value}. 
Models like TCluster \cite{alonso2009clustering} and time-based language models \cite{Li-Croft} used document timestamps and decay functions to model recency, while others like \citet{berberich2010language} combined metadata and vague expressions in probabilistic ranking models. To handle implicit temporal intent, techniques such as median timestamp analysis \cite{kanhabua2010determining} and query log mining \cite{metzler2009improving} were introduced. 

Other strategies focused on enhancing recency-aware retrieval. \citet{jatowt2005temporal} proposed re-ranking methods using archived web snapshots to favor fresher content, while \citet{dong2010time} incorporated real-time Twitter signals, and \citet{setty2017modeling} used news signals into crawling and ranking to support time-sensitive queries. 
Efficient indexing methods were also developed to support temporal queries over evolving corpora such as Wikipedia and web archives \cite{anand2011temporal, anand2012index, holzmann2016tempas}.  \citet{styskin2011recency} introduced a machine learning model to predict recency sensitivity, combining it with greedy diversification to balance freshness and topical relevance.
%\cite{anand2011temporal,anand2012index} proposed efficient indexing methods to support temporal queries on temporal collections such as Wikipedia version history and Web archives~\cite{holzmann2016tempas}.

As TIR matured, researchers began modeling the temporal dynamics of both queries and documents. \citet{kulkarni2011understanding} analyzed how user intents evolve over time, highlighting the need for adaptive retrieval strategies that can respond to temporal drift in query behavior. \citet{joho2013survey} studied the prevalence of different temporal orientations of user queries, and the strategies user apply to find temporally relevant content from the past, future or present. Later systems adapted ranking strategies to temporal query profiles using machine learning \cite{kanhabua2012learning} or temporal interval representations \cite{rizzo2022ranking}.

Early QA systems like \citet{harabagiu2005question} relied on TimeML and lexical resources like WordNet \cite{miller-1992-wordnet} for event reasoning. To handle complex temporal questions more effectively, \citet{saquete2004splitting, saquete2009enhancing} introduced a multi-layered QA architecture that decomposed questions into temporally constrained sub-questions using temporal expression taggers like TERSEO \cite{saquete2003terseo}. These approaches showed improved precision and generalizability across languages.

%Despite their simplicity, rule-based and statistical methods introduced key mechanisms of temporal intent modeling, expression normalization, and timeline reasoning that continue to influence more advanced systems.
\textit{In retrospect, rule-based and statistical systems introduced many principles that remain relevant: time expression normalization, explicit modeling of document recency and user intent, and the integration of temporal structure into retrieval and reasoning pipelines. While their lack of adaptability to diverse and large-scale contexts eventually limited their impact, they provided a crucial foundation that modern neural and retrieval-augmented models continue to build upon.}

\section{Domain-Specific Temporal QA}
\label{appendix:domain_specific_QA}
While much of the existing research in TQA focuses on open-domain settings using Wikipedia and news corpora, domain-specific applications present unique temporal reasoning challenges that remain largely underexplored. Unlike general TQA systems that handle diverse topics with explicit temporal markers, specialized domains require an understanding of domain-specific temporal conventions, implicit temporal relationships, and field-specific accuracy requirements. 

\paragraph{Medical Domain}
Temporal reasoning in clinical narratives has been explored in foundational work. The TimeText system, for instance, demonstrated early success in temporal clinical QA, achieving 84\% accuracy on 147 time-oriented questions about discharge summaries \cite{zhou2008evaluation}. Other efforts have used semantic web techniques to answer time-based clinical questions, such as event sequencing in patient histories \cite{tao2010time}.

Despite comprehensive reviews of temporal reasoning in medical text processing \cite{OLEX2021103784, zhou2007temporal}, modern temporal QA systems for medical applications remain limited. Recent work, such as Improving Health Question Answering with Reliable and Time-Aware Evidence Retrieval \cite{vladika-matthes-2024-improving}, incorporates recency and citation count for prioritizing reliable evidence, but falls short of full temporal reasoning. Open challenges remain in multi-hop reasoning over patient timelines, temporally scoped symptom progression, and treatment-event linking—tasks crucial to clinical decision support yet largely unexplored from a QA perspective.

\paragraph{Legal Domain}

Legal systems present inherent temporal complexity, shaped by evolving statutes, shifting precedents, and regulatory amendments. While several legal QA datasets exist, such as \citet{buttner-habernal-2024-answering, kien-etal-2020-answering, chen2023equals, zhong2020jec}, most are focused on static question answering and do not address the temporal dynamics of legal knowledge. Recent advances like ChronosLex  \cite{t-y-s-s-etal-2024-chronoslex} introduce time-aware incremental training for better generalization across legal epochs, signaling early momentum toward temporal legal understanding.

However, most legal QA systems do not yet perform event ordering, precedent timeline alignment, or reasoning over legislative changes. For example, questions such as \textit{“Which case law was applicable prior to the 2015 amendment of the privacy statute?”} remain out of scope for current systems. There is a pressing need for temporal-aware retrieval and reasoning architectures capable of navigating legal timelines and citations.

\paragraph{Financial Domain} In the financial domain, temporal modeling is well-established through time-series forecasting methods used for stock price prediction, volatility estimation, and risk assessment. However, these approaches typically operate on structured numerical data and do not engage in temporal reasoning over natural language. As a result, financial temporal QA remains a largely unexplored frontier.

Several financial QA datasets have emerged in recent years, primarily focused on numerical reasoning. Notable examples include FinQA \cite{chen-etal-2021-finqa}, FinTextQA \cite{chen-etal-2024-fintextqa}, and FinDER \cite{choi2025finder}.

While these resources advance quantitative and factual QA, they fall short of supporting temporal inference. Complex questions such as \textit{“How have regulatory changes since 2008 affected current banking policies?”} require systems to track policy evolution, integrate document publication dates, and synthesize trends across multiple temporal anchors (capabilities not currently addressed by existing models or datasets). Bridging this gap would require designing temporal-aware retrieval systems and multi-hop reasoning frameworks tailored to financial text, capable of answering questions grounded in historical context and regulatory shifts.

\paragraph{Real-World Deployment and Domain-Specific Challenges}

Deploying temporal QA systems in high-stakes, real-world settings presents challenges that require specialized reasoning capabilities. In the medical domain, errors in temporal interpretation can adversely affect clinical decision-making, demanding high standards of accuracy and interpretability. Legal information systems must navigate jurisdiction-specific temporal expressions (e.g., “Pre-Brexit regulatory regime”) and shifting hierarchies of legal precedent that differ across regions and evolve over time. Financial applications require real-time temporal understanding for tasks such as regulatory compliance and market analysis, where even minor misalignments in temporal reasoning can lead to significant operational or economic consequences.

Unlike general-purpose systems that rely on clearly stated temporal cues, domain-specific QA often involves inferring timeframes from implicit, field-specific conventions such as the “perioperative period” in medicine, the “discovery phase” in legal contexts, or “earnings season” in financial reporting. These terms reflect temporally bounded phases that are well understood by domain experts but opaque to models lacking domain awareness. Addressing such challenges demands tailored temporal reasoning approaches that incorporate expert knowledge, event timelines, and contextual interpretation.

\section{Detailed Future Research Directions}
\label{appendix:detailed_future_research_direction}

This appendix provides a comprehensive analysis of the research challenges outlined in Section ~\ref{sec:future}, detailing specific technical problems, examples, and potential research approaches for each identified direction.

\subsection{Dynamic Temporal Knowledge Management}

TQA systems face a persistent temporal consistency problem: their reliance on static corpora makes them unable to respond to fast-evolving information needs \citep{dhingra-etal-2022-time, wallat2024temporal}.
As real-world facts change, even small updates can disrupt temporal dependencies across related events, durations, and causal chains, a challenge known as the temporal propagation problem \citep{vu-etal-2024-freshllms}. This exposes a fundamental limitation in current architectures: they lack the modularity and temporal reasoning structure needed to adapt efficiently \citep{han-etal-2021-econet}. Future systems must move beyond isolated fact updates toward scalable frameworks that track, edit, and reason over temporal dependencies in real time.

For example, if a political leader's term end's date changes, this affects not only direct questions about their tenure but also questions about policies enacted \textit{“during their presidency}” or events that occurred \textit{“before they left office.”} Current systems often fail to propagate such updates, leading to inconsistencies. To address this, future systems should: (1) identify temporal dependency chains when facts change, (2) propagate updates through related facts, (3) maintain multi-hop consistency, and (4) reconcile conflicting temporal evidence. Integrated temporal knowledge graphs offer a promising direction for such dynamic reasoning.

\subsection{Temporally-Aware LLM Agents}

Current LLM agents face severe limitations in temporal reasoning that become pronounced in interactive settings. They consistently exhibit temporal hallucinations, generating plausible but temporally incorrect information, and fail to resolve context-dependent expressions such as \textit{"last Tuesday"} or "\textit{since our previous discussion}" \cite{xiong-etal-2024-large,bazaga-etal-2025-learning}. These failures often lead to incoherent or contradictory answers across dialogue turns.

These failures stem from architectural constraints: transformers lack persistent temporal working memory and struggle with cross-turn temporal reference resolution \cite{ge2025tremu}. As a result, agents treat dialogue turns largely independently, causing temporal anchors to drift over extended interactions.

Addressing these issues requires explicit temporal modeling mechanisms, including timeline tracking across turns, event memory that preserves temporal relations, multi-turn consistency enforcement, and robust temporal reference resolution.

Concretely, a minimally viable temporal agent would extend a base LLM with three explicit components:
(i) a persistent temporal memory that stores events and dialogue states indexed by time,
(ii) a timeline tracking module that maintains and updates temporal anchors across interactions, and
(iii) a consistency mechanism that verifies newly generated responses against stored temporal constraints.
Even without full symbolic reasoning, such an agent would mitigate common failure modes such as temporal drift, inconsistent reuse of temporal references, and hallucinated event ordering.

Existing multi-session memory and neuro-symbolic approaches already implement individual components of this pipeline—e.g., timeline construction and self-verification \cite{bazaga-etal-2025-learning}, symbolic temporal constraint checking \cite{liang2025nestr}, and executable temporal reasoning \cite{zhu-etal-2023-question}—but integrating them into a unified temporal agent remains an open research challenge.

\subsection{Diachronic and Synchronic Knowledge Integration}

Temporal questions often require combining diachronic sources (which capture change over time) and synchronic sources (which reflect knowledge at a specific moment). Most TQA systems treat these separately \cite{wang2022archivalqa, gruber2024complextempqa, piryani2024chroniclingamericaqa}, limiting their ability to fully answer questions that span both historical context and present-day facts.

For instance, the question \textit{"How has unemployment changed since 2008, and what is the current rate?"} requires integrating long-term trends from news archives (diachronic) with recent statistical summaries (synchronic). Current models often handle only one type of source effectively.

This integration requires more than merging sources; it demands alignment across different temporal granularities and anchoring schemes. Conflicts between historical accounts and retrospective summaries can further complicate the reasoning process.

Future systems should develop: (1) temporal alignment algorithms that can map events across different temporal representations, (2) cross-source reasoning frameworks that can weight and combine evidence from different temporal paradigms, (3) conflict resolution mechanisms for handling disagreements between diachronic and synchronic accounts, and (4) unified temporal representations that can accommodate both evolving and static knowledge seamlessly.

\subsection{Temporal Uncertainty and Confidence Modeling}

Many historical events involve uncertain or approximate dates, yet most TQA systems treat all temporal information as exact. This creates a mismatch with real-world complexity, particularly in multi-step reasoning where uncertainty can propagate and amplify.

For instance, the question \textit{"What happened between the fall of Rome and the beginning of the Renaissance?"} involves fuzzy boundaries: the fall of Rome (often dated to 476 AD) and the Renaissance (spanning the 14th to 16th centuries) are debated and regionally variable. Systems must represent and reason over this imprecision, not resolve it artificially.

Temporal uncertainty also arises in natural language expressions like \textit{“around that time”} or \textit{“in the late period,”} and from conflicting historical accounts. Ignoring these leads to overconfident and potentially misleading answers.

Future systems should incorporate: (1) probabilistic temporal representations that can model uncertain dates and durations, (2) confidence propagation mechanisms for multi-step temporal reasoning, (3) uncertainty-aware answer generation that communicates temporal confidence to users, (4) conflict-aware reasoning that can handle and explain disagreements between sources, and (5) evaluation frameworks that reward appropriate uncertainty rather than false precision \citep{vashishtha-etal-2019-fine, zhou-etal-2021-temporal, fatemi2024test}.

\subsection{Multilingual and Multimodal Temporal QA}

Temporal expressions, date formats, and cultural references vary widely across languages and modalities, yet most TQA systems are developed primarily for English and textual input. This poses challenges in multilingual and multimodal settings where temporal cues appear in varied forms such as non-Gregorian date formats in documents, seasonal imagery in videos, or handwritten timestamps in scanned texts. Current systems struggle to interpret these diverse signals due to limited cultural grounding and inadequate multimodal integration. For example, a model may fail to recognize a lunar calendar reference in Arabic or temporal context from a video of a snowstorm, suggesting winter. 

Recent benchmarks like TimeDial \cite{qin-etal-2021-timedial} and TOMATO \cite{shangguan2024tomato} highlight the need for culturally and visually grounded temporal reasoning. Future research should focus on developing multilingual temporal taggers, temporally annotated datasets in low-resource languages, and cross-modal alignment techniques that jointly reason over text, images, and video to capture time-related meaning in diverse cultural contexts \cite{qin-etal-2021-timedial,shangguan2024tomato, wu2025videoqa}.

\subsection{Implicit Temporal Intent Understanding}

Many temporal questions conceal their intended timeframes, making implicit temporal intent one of the most difficult challenges in TQA. For example, the question, \textit{"What caused the economic crisis during Trump’s presidency?"} could refer to events between 2017-2021 (e.g., COVID-19 or trade-related downturns) or more recent crises if the question is asked retrospectively. Current systems often fail to detect such ambiguity, either assuming the most recent crisis, misinterpreting the temporal anchor, or overlooking the ambiguity entirely.

This challenge stems from various factors: (1) conversational context, where prior dialogue establishes implicit timeframes; (2) cultural assumptions, such as shared historical reference points; (3) domain conventions, where temporal scope is understood implicitly; and (4) individual user perspectives or real-time situational context (e.g., ongoing events).

Existing models rely heavily on explicit temporal expressions and struggle with contextual inference \citep{zhou-etal-2021-temporal, allein-etal-2023-implicit, chu-etal-2024-timebench}. Future work should focus on: (1) contextual intent detection models, (2) user and domain-aware temporal reasoning, (3) clarification strategies for resolving ambiguity, and (4) evaluation frameworks for temporal intent detection across diverse scenarios.

\subsection{Evaluation and Benchmarking for Temporal Reasoning}

Current evaluation practices for TQA largely rely on standard IR/NLP metrics such as accuracy, F1, MRR, or NDCG. While useful, these metrics often overlook the specific demands of temporal reasoning, such as correct temporal anchoring, event ordering, consistency across multi-hop inferences, and robustness to temporal ambiguity.

In Temporal QA, benchmarks such as TimeQA \citep{chen2dataset}, MenatQA \citep{wei-etal-2023-menatqa}, ComplexTempQA \cite{gruber2024complextempqa}, TimeBench \citep{chu-etal-2024-timebench} and TEMPO \cite{abdallah2026tempo} provide test sets that target temporally sensitive tasks, including temporal entailment, counterfactual reasoning, and timeline ordering. However, evaluation often reduces to span-level accuracy or multiple-choice correctness. These benchmarks represent early attempts to probe temporal reasoning beyond static fact recall, but their evaluation protocols remain largely inherited from non-temporal QA.

In Temporal IR, metrics sometimes integrate recency weighting or time decay functions \citep{berberich2010language}, but still conflate temporal and topical relevance.

Future work should aim to develop: (1) evaluation metrics that capture temporal grounding and coherence, (2) metrics sensitive to ambiguous or conflicting time signals, and (3) unified protocols for comparing temporal capabilities across models and tasks.

An important methodological concern in Temporal QA is the potential overlap between benchmark datasets and large language model pretraining corpora. Many TQA datasets (e.g., TimeQA \cite{chen2dataset}, ArchivalQA \cite{wang2022archivalqa}) are derived from Wikipedia or news archives that are likely included in model training data, raising questions about whether reported performance reflects genuine temporal reasoning or memorization. Some benchmarks attempt to mitigate this issue through time-restricted or post-training evaluation: RealTimeQA \cite{realtimeqa} and FreshQA \cite{vu-etal-2024-freshllms} focus on events occurring after model training cutoffs, while Test of Time \cite{fatemi2024test} employs synthetic tasks designed to minimize reliance on memorized world knowledge. However, most existing benchmarks lack systematic contamination analysis or explicit controls for pretraining overlap. Addressing this gap will require temporal decontamination protocols and evaluation on temporally out-of-distribution data.

\begingroup
\rowcolors{2}{gray!10}{white}
\begin{table*}[htbp]
\centering
\small
%\begin{tabular}{p{2cm}p{1.5cm}p{3cm}p{3cm}p{2.5cm}p{3cm}}
\begin{tabularx}{\textwidth}{>{\hsize=1.2\hsize}X >{\hsize=0.8\hsize}X >{\hsize=0.8\hsize}X >{\hsize=0.9\hsize}X >{\hsize=1.1\hsize}X >{\hsize=1.2\hsize}X}
\hline
\textbf{Method} & \textbf{Type} & \textbf{Task} & \textbf{Temporal Representation} & \textbf{Temporal Signals} & \textbf{Architecture} \\
\hline

\textbf{TempoT5 \cite{dhingra-etal-2022-time}} &Encoder-Decoder &TQA & Explicit timestamps & Document timestamps + temporal context & T5 + temporal conditioning \\

\textbf{TempoBERT \cite{rosin2022time}} & Encoder & Semantic Change Detection, Sentence Time PredictionSemantic Change Detection, Sentence Time Prediction & Explicit timestamps & Time tokens & BERT + temporal masking \\

\textbf{BiTimeBERT \cite{wang2023bitimebert}} & Encoder & Event Time Estimation, Document Dating,  TQA  & Explicit timestamps and content temporal expressions  &Document timestamps + temporal expressions in text & BERT + dual temporal encoding \\

\textbf{TALM \cite{ren-etal-2023-time}} & Temporal adaptation + Hierarchical modeling & Historical Dating & Implicit (Time-specific word variants) & Time-specific representations & BERT + Temporal Adaptation + Hierarchical Document Encoder
 \\

\textbf{SG-TLM \cite{su-etal-2023-efficient}} &TLM  & Temporal Language Modeling, TQA &Implicit Temporal Information & Syntactic role distributions + timestamps & BERT + syntax-guided masking + temporal-aware masking  \\

\textbf{TSM \cite{cole-etal-2023-salient}} &Encoder &Temporal language modelling, TQA & Implicit temporal spans &Temporal expressions (dates, durations) &T5 + temporal span masking \\

\textbf{Temporal Attention \cite{rosin-radinsky-2022-temporal}} &Encoder &Semantic Change Detection &Explicit Document timestamp & Document timestamp & Transformer encoder with time-aware attention via time matrix T \\

\textbf{TempRetriever \cite{abdallah2025tempretriever}} & Neural Embedding  & TQA & Explict+Implicit temporal information & Query timestamps + document timestamps & Semantic + Temporal Encoder \\

\textbf{TsContriever \cite{wu2024time}} &Neural Retrieval &TQA &Explict+Implicit temporal information &Query timestamps + document timestamps & Semantic + Temporal Encoder\\

\textbf{TempRALM \cite{gade2024itsabouttime}} & RAG & TQA & Explicit temporal Information & Query + document timestamps & RAG + temporal enhancement \\

\textbf{TimeR4 \cite{qian-etal-2024-timer4}} &RAG &TQA &Explicit temporal facts & Question constraints + TKG timestamps& Retrieve-Rewrite-Retrieve-Rerank with contrastive time-aware retriever + LLaMA2 reasoning\\

\textbf{MRAG \cite{siyue2024mrag}} & Neural + Symbolic & TQA & Explicit temporal information & Query constraints + Document timestamps & Modular RAG framework ( Question Processing, Retrieval+ Summarization +Semantic-Temporal Hybrid Ranking \\

\hline

\end{tabularx}

\caption{Comprehensive summary of prominent approaches to TQA and temporally-aware language understanding. The table classifies each method by: \textit{Type}, indicating the architectural paradigm, \textit{Task}, specifying the temporal reasoning or language understanding objective; \textit{Temporal Representation}, describing whether temporal information is explicitly or implicitly modeled; \textit{Temporal Signals}, referring to the temporal cues utilized, and \textit{Architecture}, summarizing the model components and any temporal-specific adaptations. This overview highlights the variety of strategies employed to capture, encode, and reason over temporal context.}
\label{tab:temporal_qa_part1}
\end{table*}

\endgroup

%\textbf{TCluster} & Query + Docs & Ranked docs & Web archives & Timeline construction & Time-based clustering \\
%\hline

%\textbf{Saquete et al.} & Question & Answer & News + synthetic & Event ordering & Question decomposition \\
%\hline

%\textbf{NeuralDater} & Document text & Timestamp & News articles & Structural reasoning & Joint syntactic-temporal \\
%\hline
\begingroup
\rowcolors{2}{green!5}{white}
\begin{table*}[htbp]
\centering
\small
\begin{tabularx}{\textwidth}{>{\hsize=1.2\hsize}X >{\hsize=0.8\hsize}X >{\hsize=0.8\hsize}X >{\hsize=0.9\hsize}X >{\hsize=1.1\hsize}X >{\hsize=1.2\hsize}X}
\hline
\textbf{Method} & \textbf{Input} & \textbf{Output} & \textbf{Knowledge} & \textbf{Processing} & \textbf{Methodology} \\
\hline

\textbf{TempoT5 \cite{dhingra-etal-2022-time}} &Text + temporal prefix &Masked span prediction &CUSTOMNEWS  &Temporal conditioning + uniform sampling & Input prefixing for time-aware modeling and efficient updates\\

\textbf{TempoBERT \cite{rosin2022time}} & Text + timestamps & Time tokens & News corpus & Implicit temporal reasoning via time masking & Time-aware pretraining with masking \\

\textbf{BiTimeBERT \cite{wang2023bitimebert}} & News articles with timestamps & Time predictions (day/month/year granularity) & NYT corpus & Bi-temporal reasoning using both timestamp and content time
 & Time-Aware Masked Language Modeling, Document Dating \\

\textbf{TALM \cite{ren-etal-2023-time}} & Historical text & Time period classification
 & Chinese Twenty-Four Histories + English Royal Society Corpus & Temporal word adaptation and contextual learning
& Learn separate word representations per time period + alignment

 \\

\textbf{SG-TLM \cite{su-etal-2023-efficient}} &Text + timestamp & Masked token + time prediction &WMT News Crawl + Reddit Time Corpus &Syntax-guided masking (SGM) + temporal-aware masking (TAM) &Syntax-based lexicon selection for efficient temporal adaptation \\

\textbf{TSM \cite{cole-etal-2023-salient}} &Text with temporal expressions & Masked span prediction & Wikipedia &Temporal span masking using SUTIME parser + intermediate training & SUTIME parser identifies temporal expressions (Time, Duration, Set, Date) for targeted masking during intermediate pretraining \\

\textbf{Temporal Attention \cite{rosin-radinsky-2022-temporal}} &Text + Document timestamp &Time-specific contextualized embeddings & Time-annotated corpora (SemEval datasets) & Time-aware self-attention mechanism with additional time matrix & Extends self-attention with time matrix T to condition attention weights on temporal context via time-aware dot products. \\

\textbf{TempRetriever \cite{abdallah2025tempretriever}} & Query + Query timestamp + Document + Document timestamp & passages with temporal +semantic relevance  & NYT + ChroniclingAmerica Corpus & Temporal alignment + Temporal understanding & Learned temporal embeddings with fusion techniques + Time-based negative sampling

  \\

\textbf{TsContriever \cite{wu2024time}} &Time-sensitive questions & Top-k temporally relevant documents& Nobel Prize benchmark corpus + Wikidata QA pairs &Contrastive learning + Query-side fine-tuning + Query routing & Supervised contrastive learning with time-aware query tuning and routing \\

\textbf{TempRALM \cite{gade2024itsabouttime}} &  Time-sensitive queries & Answer & Knowledge base & Temporal proximity-based document ranking & Temporal scoring \\

\textbf{TimeR4 \cite{qian-etal-2024-timer4}} &Time-sensitive queries &Answer (entity or timestamp) &Knowledge base &Retrieve-Rewrite-Retrieve-Rerank pipeline with contrastive learning and LLM finetuning & Temporal fact retrieval, question rewriting, time-aware reranking, and LLM finetuning \\

\textbf{MRAG \cite{siyue2024mrag}} & Temporal questions + Document

 & Ranked passages & Wikipedia & Symbolic temporal scoring + Semantic-temporal hybrid ranking

 & Question decomposition (MC + TC) + Symbolic temporal

\\

\hline

\end{tabularx}

\caption{Detailed comparison of TQA methods across key system dimensions. Each method is described in terms of its \textit{Input} (the structure and temporal components of the data provided to the model), \textit{Output} (the nature of predictions or generated responses), \textit{Knowledge} (the underlying corpora or knowledge sources used), \textit{Processing} (the mechanisms used to handle temporal information, such as conditioning, masking, or alignment), and \textit{Methodology} (the core modeling strategy or training paradigm employed). This table highlights the diverse ways in which temporal signals and knowledge are integrated to enhance reasoning over time-sensitive questions.}
\label{tab:temporal_qa_part2}
\end{table*}
\endgroup

\begin{table*}[t]
\centering
\small
\begin{tabularx}{\textwidth}{l|X}
\hline
\textbf{Evaluation Goal} & \textbf{Recommended Datasets} \\
\hline
Temporal Robustness / Answer Drift & StreamingQA \cite{liska22a-streaminqa}, RealTimeQA \cite{realtimeqa}, PAT-Questions \cite{meem-etal-2024-pat}, FreshQA \cite{vu-etal-2024-freshllms} \\

Temporal Retrieval (Diachronic Evidence) & ArchivalQA \cite{wang2022archivalqa}, ChroniclingAmericaQA \cite{piryani2024chroniclingamericaqa}, NewsQA \cite{trischler-etal-2017-newsqa} \\
Reasoning-focused Temporal Retrieval & TEMPO \cite{abdallah2026tempo}\\

Implicit Temporal Intent & TiQ \cite{jia2024tiq}, TemporalQuestions \cite{wang2021improving}, ArchivalQA \cite{wang2022archivalqa} \\

Event Ordering & TORQUE \cite{ning-etal-2020-torque}, TRACIE \cite{zhou-etal-2021-temporal}, TimeBench \cite{chu-etal-2024-timebench} \\

Duration Inference & MenatQA \cite{wei-etal-2023-menatqa}, TEMPREASON \cite{tan-etal-2023-towards}, TimeBench \cite{chu-etal-2024-timebench} \\

Multi-hop Temporal Reasoning & ComplexTempQA \cite{gruber2024complextempqa}, Complex-TR \cite{tan-etal-2024-towards} , TempQuestions \cite{jia2018tempquestions} \\

Knowledge Updating & TempLAMA \cite{dhingra-etal-2022-time}, StreamingQA \cite{liska22a-streaminqa}, PAT-Questions \cite{meem-etal-2024-pat} \\

End-to-End Temporal QA & ComplexTempQA \cite{gruber2024complextempqa}, ChroniclingAmericaQA \cite{piryani2024chroniclingamericaqa} TimeQA \cite{chen2dataset}, ArchivalQA \cite{wang2022archivalqa} \\

Temporal Ambiguity Resolution &
SituatedQA \cite{zhang-choi-2021-situatedqa}, TEMPAMBIQA \cite{piryani-etal-2024-detecting} \\

\hline
\end{tabularx}
\caption{Dataset selection guide for temporal question answering. Each row summarizes which benchmarks are most suitable for evaluating specific temporal capabilities.}
\label{tab:dataset_selection_guide}
\end{table*}

\end{document}